\documentclass[sn-apa, iicol]{sn-jnl}% APA Reference Style 
\usepackage{graphicx}%
\usepackage{multirow}%
\usepackage{amsmath,amssymb,amsfonts}%
\usepackage{amsthm}%
\usepackage{mathrsfs}%
\usepackage[title]{appendix}%
\usepackage{xcolor}%
\usepackage{textcomp}%
\usepackage{manyfoot}%
\usepackage{booktabs}%
\usepackage{algorithm}%
\usepackage{algorithmicx}%
\usepackage{algpseudocode}%
\usepackage{listings}%
\usepackage{bm}%
\usepackage{array}%
\usepackage{color}
\usepackage[accsupp]{axessibility}
\usepackage{diagbox}
\usepackage{adjustbox}

\theoremstyle{thmstyleone}%
\theoremstyle{thmstyletwo}%

\theoremstyle{thmstylethree}%

\begin{document}
% Task-agnostic Unified Face Alignment via Face Structure Prompts and Semantic Alignment
\title[Article Title]{Mitigating Knowledge Discrepancies among Multiple Datasets for Task-agnostic Unified Face Alignment}


\author[1]{\fnm{Jiahao} \sur{Xia}}\email{Jiahao.Xia-1@uts.edu.au}

\author*[1]{\fnm{Min} \sur{Xu}}\email{Min.Xu@uts.edu.au}

\author[2]{\fnm{Wenjian} \sur{Huang}}\email{huangwj@sustech.edu.cn}

\author[2]{\fnm{Jianguo} \sur{Zhang}}\email{zhangjg@sustech.edu.cn}

\author[1]{\fnm{Haimin} \sur{Zhang}}\email{Haimin.Zhang@uts.edu.au}

\author[3]{\fnm{Chunxia} \sur{Xiao}}\email{cxxiao@whu.edu.cn}

\affil*[1]{\orgdiv{Faculty of Engineering and IT}, \orgname{University of Technology Sydney}, \orgaddress{\city{Ultimo}, \postcode{2007}, \state{NSW}, \country{Australia}}}

\affil[2]{\orgdiv{Department of Computer Science and Engineering}, \orgname{Southern University of Science and Technology}, \orgaddress{\city{Shenzhen}, \postcode{518055}, \state{Guangdong}, \country{China}}}

\affil[3]{\orgdiv{School of Computer Science}, \orgname{Wuhan University}, \orgaddress{\city{Wuhan}, \postcode{430072}, \state{Hubei}, \country{China}}}

\abstract{Despite the similar structures of human faces, existing face alignment methods cannot learn unified knowledge from multiple datasets with different landmark annotations. The limited training samples in a single dataset commonly result in fragile robustness in this field. To mitigate knowledge discrepancies among different datasets and train a task-agnostic unified face alignment (TUFA) framework, this paper presents a strategy to unify knowledge from multiple datasets. Specifically, we calculate a mean face shape for each dataset. To explicitly align these mean shapes on an interpretable plane based on their semantics, each shape is then incorporated with a group of semantic alignment embeddings. The 2D coordinates of these aligned shapes can be viewed as the anchors of the plane. By encoding them into structure prompts and further regressing the corresponding facial landmarks using image features, a mapping from the plane to the target faces is finally established, which unifies the learning target of different datasets. Consequently, multiple datasets can be utilized to boost the generalization ability of the model. The successful mitigation of discrepancies also enhances the efficiency of knowledge transferring to a novel dataset, significantly boosts the performance of few-shot face alignment. Additionally, the interpretable plane endows TUFA with a task-agnostic characteristic, enabling it to locate landmarks unseen during training in a zero-shot manner. Extensive experiments are carried on seven benchmarks and the results demonstrate an impressive improvement in face alignment brought by knowledge discrepancies mitigation. The code is available at \url{https://github.com/Jiahao-UTS/TUFA}.}


\maketitle

\section{Introduction}\label{sec1}

Facial landmark serves as an important intermediate feature in many downstream tasks, such as face recognition \citep{FaceAAAI}, facial emotion recognition \citep{Dynamic_Facial_Expression, Sparse_Code_Facial_Expressing} and facial image synthesis \citep{IJCVGAN}. Despite significant progress in recent years, few existing face alignment methods focus on mitigating the knowledge discrepancies among datasets, resulting in very limited usage of dataset knowledge. Consequently, despite the similar structure of human faces, most existing methods can be only trained on a single dataset. The limited training samples lead to unsatisfied generalization ability. When applying a model to a group of newly defined landmarks, the entire model still needs to be re-trained. As a result, these knowledge discrepancies lead to performance degradation and high costs for downstream tasks.

As shown in Fig. 1, similar to other vision tasks \citep{ZeroshotReview, FewshotSurvey}, there are three types of methods that can train a face alignment model: many-shot, few-shot, and zero-shot learning. Many-shot learning is a widely used method in face alignment. It does not consider sharing knowledge among multiple datasets, directly training a model on a dataset with consistent landmark annotations. Hence, the training can be very costly and time-consuming. To relief the reliance on the number of training samples, few-shot face alignment has been explored in recent works \citep{3FabRec, GeoFewShot}. Similar to few-shot learning in other tasks\citep{FewshotSurvey, FewShotRec}, few-shot face alignment transfers the knowledge learned from a large-scale base dataset to the target landmarks using very few samples. Explicitly mitigating the knowledge discrepancies between the base and target datasets can significantly increase the efficiency of knowledge transferring, while this problem is ignored by existing methods. Further mitigating these discrepancies can extend the learned knowledge from predefined landmarks to the overall face structure, making the model task-agnostic. Consequently, using semantic information as a prompt can locate landmarks that were unseen during training in a zero-shot manner. Unfortunately, no work has been done on this topic yet.

\begin{figure}[t!]
	\centering
	\includegraphics[width=\linewidth]{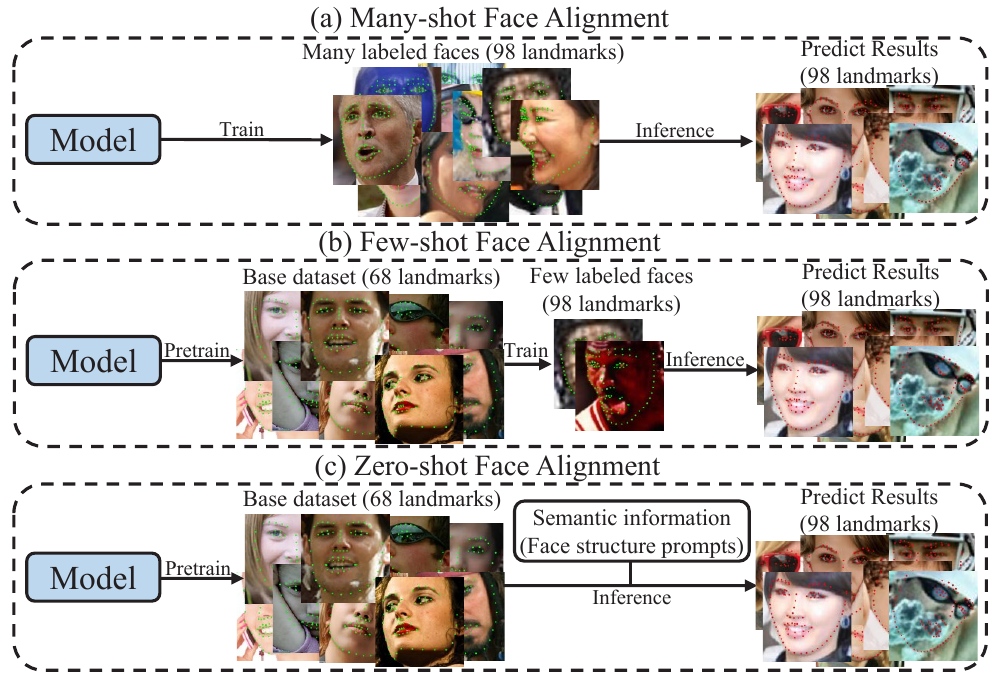}
	\caption{A schematic diagram of many-shot, few-shot, and zero-shot learning in face alignment. (a) many-shot face alignment directly trains the model on a large-scale dataset labeled with target landmarks. (b) few-shot face alignment pre-trains the model on a base dataset, and then further transfer the model using a few samples labeled with target landmarks. (c) zero-shot face alignment also pre-trains the model on a base dataset, while the model directly locates the target landmarks via semantic information without training.}  
	\label{fig1}
\end{figure}

In this paper, we propose a novel strategy to mitigate the knowledge discrepancies among multiple datasets for task-agnostic unified face alignment (TUFA). Instead of treating each facial landmark as an independent regression target, our strategy uses labeled facial landmarks to learn a mapping from an interpretable plane to the target faces. This approach unifies the learning objective and provides a foundation for mitigating knowledge discrepancies. Specifically, we calculate a mean face shape for each dataset. To align the mean shapes from multiple datasets on the same plane according to their semantics and eliminates their semantic ambiguity, we incorporate each mean shape with a group of semantic alignment embeddings. They transform these mean shapes during training and finally find optimal positions on the plane to represents their semantics. Consequently, the semantic ambiguity can be eliminated. Each point from these aligned shapes from different datasets can be view a anchor on the plane. As a result, the learning targets of datasets are unified and their knowledge discrepancy are largely mitigated. By encoding them into structure prompts and regressing the positions of their corresponding landmarks, the mapping to target face can be established using multiple datasets.

\begin{figure}[t!]
	\centering
	\includegraphics[width=\linewidth]{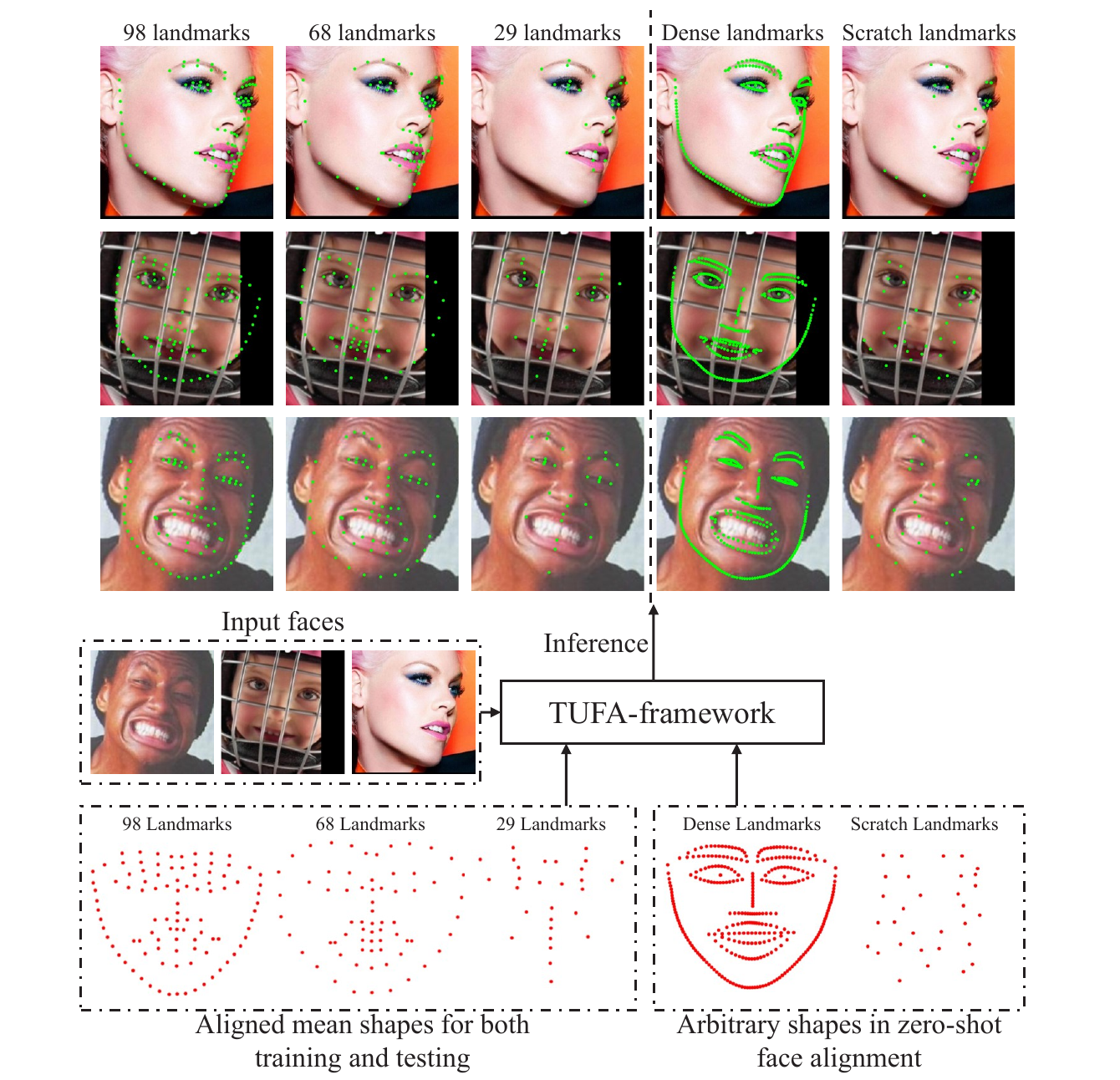}
	\caption{Teaser of TUFA. TUFA employs the aligned face shapes as anchors to mapping an interpretable plane to target faces. Consequently, it maximizes the knowledge use from multiple datasets to improve the generalization ability. By explicitly representing face structure using this plane, TUFA is able to locate both pre-defined landmarks and those landmarks unseen during training in inference phase.}  
	\label{fig2}
\end{figure}

Moreover, previous face-alignment methods based on heatmap regression and coordinate regression treat each landmark regression as a separate task, resulting in task-specific knowledge. Although human faces share a similar structure, it remains challenging to extend this knowledge to unseen facial landmarks. Nevertheless, by mitigating the knowledge discrepancies using the proposed strategy, the learned knowledge is not only limited to the labeled landmarks in training but also includes \textbf{task-agnostic} knowledge: the entire face structure. The learned task-agnostic knowledge can be efficiently transferred to a group of newly defined landmarks using only a few labeled samples, which significantly boosts the performance of few-shot face alignment. Moreover, by explicitly representing the face on an interpretable plane and encoding its coordinates into structure prompts, this knowledge can also be used to localize unseen facial landmarks without fine-tuning. Therefore, TUFA also achieves zero-shot face alignment for the first time, predicting not only the pre-defined landmarks, but also those landmarks unseen during training, as shown in Fig. 2.

We train TUFA using four widely used datasets and carry out four \textbf{within-dataset} validations and three \textbf{cross-dataset} validations in a many-shot manner. TUFA achieves state-of-the-art performance on these seven validations using \textbf{only one unified model}. This pre-trained model demonstrates strong generalization ability, making it particularly useful for further few-shot face alignment. Thanks to the mitigated discrepancies, the TUFA trained with only 10 labeled samples can even outperform the state-of-the-art few-shot face alignment methods trained on the entire training set. Moreover, due to its task-agnostic property, TUFA achieves zero-shot face alignment for the first time. The quantitative results show that TUFA even outperforms some fully supervised methods in a \textbf{zero-shot} manner. 

Overall, the main contributions of this paper can be summarized as follows:

\begin{itemize}
	\item This paper presents a novel strategy to mitigate knowledge discrepancies among multiple face alignment datasets by unifying their learning objectives and eliminating their semantic ambiguity. Based on this strategy, a task-agnostic unified face alignment model with stronger generalization ability can be trained across multiple datasets.

	\item To efficiently transfer the knowledge learned across multiple datasets to newly defined landmarks, we propose a few-shot and zero-shot face alignment strategy using the structural information contained in the proposed structure prompts and the learned interpretable plane.

	\item Extensive experiments and ablation studies are carried out on widely used benchmarks. TUFA achieves state-of-the-art performance in the setting of many-shot, few-shot and zero-shot face alignment, which demonstrates the strong generalization ability and transferable capability of TUFA.
\end{itemize}

\section{Related Works}\label{sec2}

It has been noticed that the performance of \textbf{many-shot face alignment} has improved significantly in recent years. However, the high cost of dataset annotation has shifted the focus of recent works towards \textbf{few-shot \& zero-shot face alignment}. Moreover, the existing models still generalizes poorly in a cross-dataset validation setting, especially for those landmarks have a big gap compared to the seen landmarks. \textbf{Multi-dataset learning in face alignment}, which is a potential strategy to address this problem, has also gained a lot of attention in recent years. Therefore, we present a concise review focusing on these three topics, given that they represent the primary challenges tackled in this paper.

\subsection{Many-shot Face Alignment}

As a very pivotal technology in computer vision, many-shot face alignment has undergone over two decades of intensive research and exploration. At the very early stage, constrained local model (CLM) \citep{CLM}, active shape model (ASM) \citep{ASM}, active appearance model (AAM) \citep{AAM, AAM_E1, AAM_E2} dominated this task. Their core idea is to align a statistical mean shape to a new face and that is why this task is named as face alignment rather than facial landmark detection. To overcome the fragile robustness and generalization, cascaded shape regression (CSR) methods \citep{COFW, SDM, DAC, CFSS, DAC-CSR} are proposed to achieve face alignment with multi-steps. Lately, the development of CNN and vision transformer (ViT) \citep{VIT} significantly boost the performance of face alignment. Recent many-shot face alignment methods can be roughly divided into two categories: heatmap regression methods and coordinate regression methods. They set each landmark as an independent regression target and their learned models are task-specific.

\textbf{Coordinate regression methods} directly project image features into 2D landmark coordinates with fully connected (FC) layers. To extract more fine-grained image feature, cascaded networks \citep{Deep_Reg, DAN, LDR}, recurrent networks \citep{MDM, RAR} and face local patches \citep{LDR, GlomFace} are widely used. The regular shape of human faces serves as an important prompt in face alignment, and recent coordinate regression methods further improve the performance by retaining the prompt in networks. Lin et al. \citep{SCDF} represent face structure with an adjacent matrix and Li et al. \citep{SDL} make the matrix learnable. Xia et al. \citep{SLPT}, Xia et al. \citep{DSLPT} and Prados-Torreblanca et al. \citep{SPIGA} further encourage the model to learn a case-dependent inherent relation based on the attention mechanism. However, the learned face structure is represented in high-dimensional latent space, making it hard to be understood and edited by human. In this paper, TUFA represents the face structure on an interpretable plane, making the prompt can be edited by human for the first time.

\textbf{Heatmap regression methods} differ significantly from coordinate regression methods, as they generate a high-resolution likelihood heatmap for each landmark using a deep network \citep{HRNet, Hourglass, SimBa, Dunet}. Although the semantic supervision enables heatmap regression methods to perform better with limited training samples, they still have some drawbacks. For instance, to transfer the predicted heatmaps into 2D coordinates, they consider the pixel with highest intensity as the optimal output. Therefore, the predicted coordinates can only be integers, resulting in a quantitative error due to lower resolution of the output heatmap than the input image. To mitigate the error, Zhang et al. \citep{SNH}, Jin et al. \citep{PIP} and Lan et al. \citep{HIH} predict the decimal part of landmark coordinate on heatmaps with an additional head; Tai et al. \citep{THA}, Chen et al. \citep{KDN}, Kumar et al. \citep{LUVLI} and Zhou el al. \citep{StarLoss} achieve subpixel landmark localization on heatmap by estimating landmark distribution. Furthermore, it is more challenging for heatmap regression methods to model the regular structure of human faces because they do not explicitly learn a representation for each individual landmark. Despite the utilization of facial boundaries as a key indicator in heatmap regression models \citep{LAB, Awing, ADNet} to maintain face structure, these models still overlook the long-term relation that exists among landmarks. Their network structure also makes it impossible for them to predict an arbitrary number of landmarks. Therefore, for any newly defined landmark, these methods must be re-trained.

\subsection{Few-shot \& Zero-shot Face Alignment}
Given that existing methods require both model re-training and dataset re-annotation for predicting a group of newly defined landmarks, recent research \citep{AVS, 3FabRec, GeoFewShot} has increasingly focused few-shot face alignment. It enables the swift adaptation to new face alignment tasks. Qian et al. \citep{AVS} expand the training set by style transfer so that the model learns a more generalized result with a limited number of training samples. However, the synthetic images have a significant domain gap to the real images, which can lead to performance degradation. And, the style transfer network still requires hundreds of samples for training. Browatzki et al. \citep{3FabRec} and He et al. \citep{GeoFewShot} set face reconstruction as a pre-text task. With the feature learned from reconstruction, the models can be transferred into face alignment with few-shot examples. Nevertheless, as the experiment results shown in \citep{GeoFewShot}, the large discrepancies between face alignment and image reconstruction can result in performance degradation when the samples for fine-tuning are relatively sufficient. Therefore, mitigating knowledge descrepancies in few-shot face alignment is quite crucial, but no study has focused on this problem yet. Self-supervised \citep{SelfLandmark} and unsupervised landmark detection \citep{Unsuper_Thewlis, Unsuper_Zhang, Unsuper_Lorenz, Unsuper_Jakab, Autolink} have also gained more attention as they eliminate the need for manual annotation. Nevertheless, the unpredictability of semantic definitions for the landmarks discovered during training, which cannot be manually assigned, presents a challenge when applying them to downstream tasks. Zhang et al. \citep{OpenAnimal} align textual prompts with visual features to extend landmark detection model to unseen categories in zero-shot manner. However, textual prompts are difficult to describe the structure relations among landmarks, making it hard to extend the model to unseen landmarks. This work implements zero-shot learning in landmark-level for the first time, significantly broadening the application range of face alignment.

\begin{figure*}[t!]
	\centering
	\includegraphics[width=\linewidth]{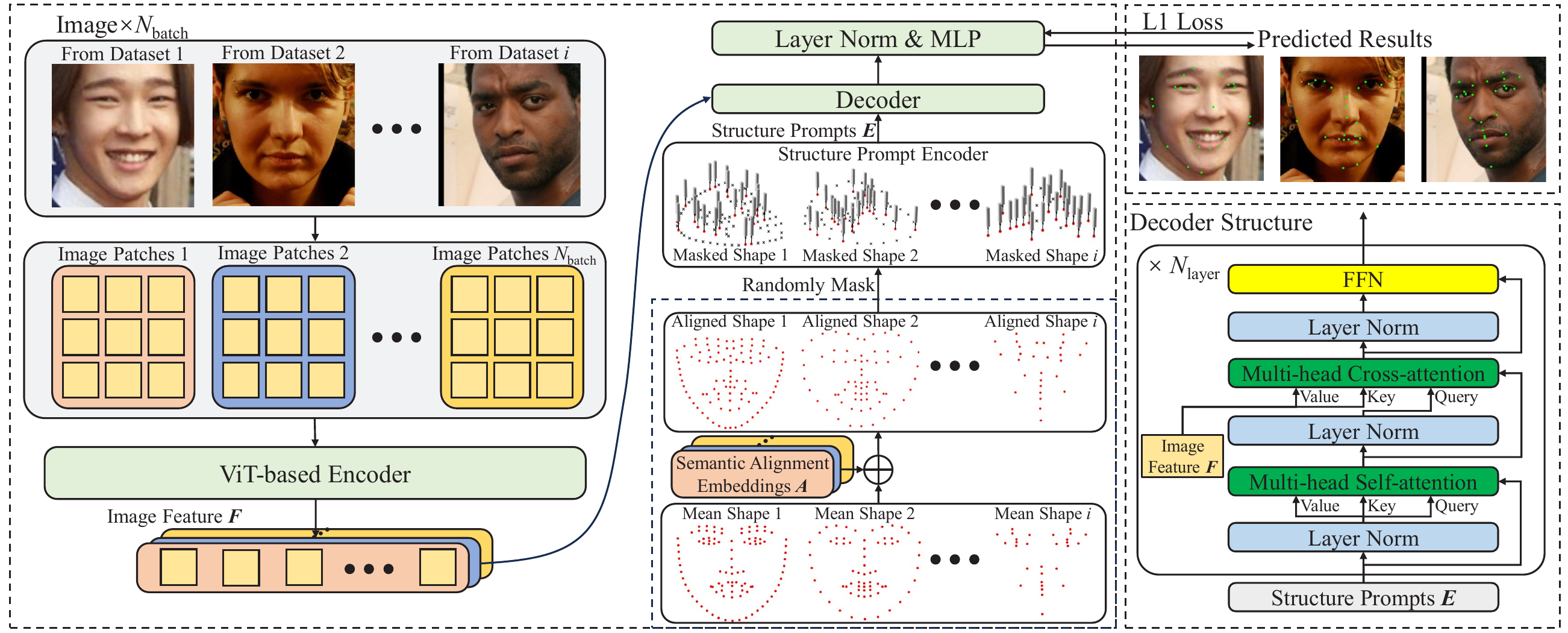}
	\caption{Overall training pipeline of TUFA. It mainly consists of three parts: a ViT-based encoder, a structure prompt encoder and a decoder. TUFA first splits input images into patches and extracts image features $\bm{F}$ with a ViT-based encoder. Each image has a corresponding statistical mean shape determined by its dataset. For multi-dataset learning, the learnable semantic alignment embeddings are added to the mean shapes to align their semantics on a shared plane. During each iteration, TUFA randomly masks these input shapes, keeping only a certain number of landmarks as the input anchors to form a batch. Then, the structure prompt encoder encodes 2D positions of the input anchors into structure prompts $\bm{E}$ using sine and cosine functions. The structure prompts finally serve as queries, mapping the mean shapes to target landmarks based on the image features. The detailed structure of the decoder is shown at the bottom right.}
	\label{fig3}
\end{figure*}

\section{Multi-datasets Learning in Face Alignment}\label{sec3}

In this section, we first introduce the overall training pipeline for TUFA and explain its approach to achieving task-agnostic face alignment. Next, we describe the proposed method for mitigating the knowledge discrepancies among different datasets, as well as the loss function of TUFA. Finally, we discuss the strategy for achieving few-shot and zero-shot face alignment using TUFA.  

\subsection{Task-agnostic Unified Face Alignment}
The overall training pipeline of TUFA is demonstrated in Fig. 3. It mainly consists of three parts: a ViT-based encoder for image feature extraction, a structure prompt encoder that generates structure prompts from the 2D coordinates of the input anchors, and a decoder aimed at mapping the input shapes to target faces.

\subsubsection{ViT-based Encoder}
We employ a ViT-based encoder to extract an image features $\bm{F}$ that later serve as essential clues in constructing the mapping between the pre-defined plane and various target faces. Given an input image $\textbf{\emph{I}} \in \mathbb{R}^{H_{\rm I} \times W_{\rm I} \times 3}$, the ViT-based encoder first splits it into regular patches with size $(H_{\rm P},  W_{\rm P})$. Then, each patch is projected into a vector with $C$ dimensions by a CNN layer and is further added with learnable positional embeddings $\bm{P} \in \mathbb{R}^{L \times C}$ to retain spatial information, where $L = \frac{H_{\rm I}}{H_{\rm P}} \times \frac{W_{\rm I}}{W_{\rm P}}$. Finally, the encoder learns the image feature $\textbf{\emph{F}} \in \mathbb{R}^{L \times C}$ via the attention mechanism. $\bm{F}$ is further fed into the cross-attention block of the decoder.

\subsubsection{Structure Prompt Encoder}
A human face can be represented on a 2D plane because it is the surface of the human head and has a regular shape. During single dataset learning, to constrain the representation of the human face to the 2D plane, we calculate a statistical mean shape from the training samples, and the coordinates of the mean shape are normalized in the range of $[-1, 1]$. The mean shape, which explicitly represents the regular structure of faces, later serves as the anchors of this plane. By regressing target landmarks based on the structure prompts of these anchors, we can establish a mapping between the 2D plane and various faces. Moreover, the mean shape also ensures that the face structure representation on the plane is easily understood by humans, allowing the structure prompts to be readily edited. During multi-dataset learning, the additional semantic alignment embeddings $\bm{A}$ should be incorporated into the mean shapes to mitigate the knowledge discrepancies among multiple datasets. The details of the semantic alignment embeddings $\bm{A}$ will be discussed in Section 3.2.

Before feeding the mean shapes into the structure prompt encoder, TUFA randomly masks part of these shapes during each iteration, retaining only $N_{\rm a}$ landmarks as the input anchors ($N_{\rm a}$ is set to 24, 25\% of 98 landmarks). The use of the mean shape masking is threefold: 1) the masked mean shape is much sparser, encouraging TUFA to learn the mapping based on long-term landmark relation rather than the relation between neighbouring landmarks. It can effectively prevent overfitting and facilitate a coherent mapping for improved performance. 2) The computational complexity of the decoder during the training phase is significantly reduced with fewer structure prompts, which enhances training speed and reduces GPU memory consumption. 3) For multi-dataset learning, it ensures the consistency in the input anchors numbers across different datasets, allowing them to form a training batch.

To encode the 2D coordinates of masked shapes into high-dimensional vectors while retaining their geometric relations, we utilize cosine and sine functions. The high-dimensional vector can be written as:
\begin{equation}
	\begin{aligned}
		E^{\rm X}_{(x, 2c)} &= \sin(x /\tau^{(2c / 0.5C)}),\\
		E^{\rm X}_{(x, 2c+1)} &= \cos(x /\tau^{(2c / 0.5C)}),
	\end{aligned}
\end{equation}
\begin{equation}
	\begin{aligned}
		E^{\rm Y}_{(y, 2c)} &= \sin(y /\tau^{(2c / 0.5C)}),\\
		E^{\rm Y}_{(y, 2c+1)} &= \cos(y /\tau^{(2c / 0.5C)}),
	\end{aligned}
\end{equation}
where $\bm{E}^{\rm X}_{x}$ and $\bm{E}^{\rm Y}_{y}$ are two vectors with 0.5$C$ dimensions, representing the landmark coordinate $(x, y)$ on X-axis and Y-axis respectively. $c \in [0, C/4)$ is the index of dimension and $\tau$ is a hyperparameter that determines the wavelengths ($\tau$ is set to 10000 in this paper). The final structure prompt $\bm{E}$ can be formulated as:
\begin{equation}
	\bm{E}_{(x, y)} = {\rm Concat}(\bm{E}^{\rm X}_{x}; \bm{E}^{\rm Y}_{y}),
\end{equation}
where ${\rm Concat}$ means concatenation process. For any fixed offset $(_{\Delta}x, _{\Delta}y)$, the vectors $\bm{E}^{\rm X}_{x + _{\Delta}x}$ and $\bm{E}^{\rm Y}_{y + _{\Delta}y}$ can be written as:
\begin{footnotesize}
	\begin{equation}
		\begin{bmatrix}\!
			E^{\rm X}_{(x+ _{\Delta}x\!, 2c)}\! \\
			\!E^{\rm X}_{(x+ _{\Delta}x\!, 2c+1)}\!
		\end{bmatrix}\!=\!
		\begin{bmatrix}
			\cos(\!\frac{_{\Delta}x}{\tau^{2c/0.5C}})&\!\sin(\frac{_{\Delta}x}{\tau^{2c/0.5C}}\!)\!\\
			\!-\!\sin(\!\frac{_{\Delta}x}{\tau^{2c/0.5C}})&\!\cos(\frac{_{\Delta}x}{\tau^{2c/0.5C}}\!)\!
		\end{bmatrix}
		\begin{bmatrix}\!
			E^{\rm X}_{(\!x\!, 2c\!)}\! \\
			\!E^{\rm X}_{(\!x\!, 2c+1\!)}\!
		\end{bmatrix},
	\end{equation}
\end{footnotesize}
\begin{footnotesize}
	\begin{equation}
		\begin{bmatrix}
			\!E^{\rm Y}_{(y+ _{\Delta}y\!, 2c)}\! \\
			\!E^{\rm Y}_{(y+ _{\Delta}y\!, 2c+1)\!}
		\end{bmatrix}\!=\!
		\begin{bmatrix}\!
			\cos(\!\frac{_{\Delta}y}{\tau^{2c/0.5C}}\!)&\!\sin(\frac{_{\Delta}y}{\tau^{2c/0.5C}})\! \\
			\!-\!\sin(\!\frac{_{\Delta}y}{\tau^{2c/0.5C}}\!)&\!\cos(\frac{_{\Delta}y}{\tau^{2c/0.5C}}\!)
		\end{bmatrix}
		\begin{bmatrix}
			\!E^{\rm X}_{(y, 2c)}\! \\
			\!E^{\rm X}_{(y, 2c\!+\!1)}\!
		\end{bmatrix}.
	\end{equation}
\end{footnotesize}Thus, $\bm{E}^{\rm X}_{x + _{\Delta}x}$ and $\bm{E}^{\rm Y}_{y + _{\Delta}y}$ can be viewed as a linear function of $\bm{E}^{\rm X}_{x}$ and $\bm{E}^{\rm Y}_{y}$ respectively, and the transformation matrix is determined by the corresponding offset. This property enables the 2D geometric relationships of the mean shape landmarks to be well retained in the high-dimensional vector for the learning of the mapping.

\subsubsection{Decoder}

The decoder is the key component for learning the mapping from the 2D plane to target faces. As shown at the bottom left in Fig. 3, the decoder mainly consists of three blocks: multi-head self-attention (MSA) block, multi-head cross-attention (MCA) block, and feed-forward network (FFN). Besides, there is an extra LayerNorm \citep{LN} before each block.

The mapping between the 2D plane and target faces should be determined by multiple anchors, not just a single one. Therefore, the MSA block is crucial because it enables the structure prompts to share their geometric information for the mapping determination. The key ($\bm{K}_z$), query ($\bm{Q}_z$), and value ($\bm{V}_z$) of the $z$-th head in MSA block can be calculated as:
\begin{small}
\begin{equation}
	\bm{K}_z\!=\!(\bm{T}_z+\bm{E}_z)\bm{W}^{\rm k}_z, \bm{Q}_z\!=\!(\bm{T}_z+\bm{E}_z)\bm{W}^{\rm q}_z, \bm{V}_z\!=\!\bm{T}_z\bm{W}^{\rm v}_z,
\end{equation}
\end{small}where $\bm{E} \in \mathbb{R}^{N_{\rm a} \times C}$ and $\bm{T} \in \mathbb{R}^{N_{\rm a} \times C}$ are the face structure prompts and the input to the MSA block respectively. $\bm{E}$ and $\bm{T}$ are further divided into $N_{\rm h}$ ($N_{\rm h}$ is the number of heads) sequences equally with $C_{\rm h} = C/N_{\rm h}$ dimensions. $\bm{W}^{\rm k}_z \in \mathbb{R}^{C_{\rm h} \times C_{\rm h}}$, $\bm{W}^{\rm q}_z \in \mathbb{R}^{C_{\rm h} \times C_{\rm h}}$, and $\bm{W}^{\rm v}_z \in \mathbb{R}^{C_{\rm h} \times C_{\rm h}}$ are three learnable matrices. The output of $z$-th head $\bm{H}_z$ can be calculated as:
\begin{equation}
	\bm{H}_z = {\rm softmax}\left(\frac{\bm{Q}_{z}\bm{K}_{z}^{T}}{\sqrt{C_{\rm h}}}\right) \bm{V}_{z}.
\end{equation}
The final output of MSA block can be written as:
\begin{equation}
	\begin{aligned}
		\mathcal{F}_{\rm MSA}(\bm{T}) = {\rm Concat}(\bm{H}_1;...;\bm{H}_{N_{\rm h}})\bm{W}_{\rm MSA},
	\end{aligned}
\end{equation}
where $\bm{W}_{\rm MSA} \in \mathbb{R}^{C \times C}$ is also a learnable matrix for linear projection.

The mapping also depends on the input face. Therefore, the MCA block is essential since it aggregates the image feature $\bm{F}$ learned by the ViT-based encoder based on the structure prompts, enabling the learned mapping to be case-dependent. The key ($\bm{K}^\prime_z$), query ($\bm{V}^\prime_z$) and value ($\bm{V}^\prime_z$) of the $z$-th head in MCA block can be formulated as:
\begin{small}
\begin{equation}
	\bm{K}_z^\prime\!=\!(\bm{F}_z\!+\!\bm{P}_z)\bm{W}^{\rm k \prime}_z\!,\! \bm{Q}_z^\prime\!=\!(\bm{T}_z^\prime\!+\!\bm{E}_z)\bm{W}^{\rm q \prime}_z\!,\! \bm{V}_z^\prime\!=\!\bm{F}_z\bm{W}^{\rm v \prime}_z\!,\!
\end{equation}
\end{small}where $\bm{T}^\prime \in \mathbb{R}^{N_{\rm a} \times C}$ and $\bm{F} \in \mathbb{R}^{L \times C}$ are the input to MCA block and image feature respectively. We reuse the learnable positional embeddings $\bm{P} \in \mathbb{R}^{L \times C}$ in the ViT-based encoder to retain the spatial information of image. Similar to the MSA block, $\bm{T}^\prime$, $\bm{F}$ and $\bm{P}$ are further divided into $N_{\rm h}$ sequences equally with $C_{\rm h} = C/N_{\rm h}$ dimensions. $\bm{W}^{\rm k \prime}_z \in \mathbb{R}^{C_{\rm h} \times C_{\rm h}}$, $\bm{W}^{\rm q  \prime}_z \in \mathbb{R}^{C_{\rm h} \times C_{\rm h}}$, and $\bm{W}^{\rm v  \prime}_z \in \mathbb{R}^{C_{\rm h} \times C_{\rm h}}$ are three learnable matrices. The output of $z$-th head $\bm{H}_z^{\prime}$ can be written as:
\begin{equation}
	\bm{H}_z^{\prime} = {\rm softmax}\left(\frac{\bm{Q}_{z}^{\prime}\bm{K}_{z}^{\prime T}}{\sqrt{C_{\rm h}}}\right) \bm{V}_{z}^{\prime}.
\end{equation}
The final output of MCA block can be formulated as:
\begin{equation}
	\begin{aligned}
		\mathcal{F}_{\rm MCA}(\bm{T}) = {\rm Concat}(\bm{H}_1^{\prime};...;\bm{H}_{N_{\rm h}}^{\prime})\bm{W}_{\rm MCA},
	\end{aligned}
\end{equation}
where $\bm{W}_{\rm MCA} \in \mathbb{R}^{C \times C}$ is also a learnable matrix.

Moreover, a FFN block fuses the features aggregated from $\bm{F}$ in the channel-wise, further enhancing the expressive ability of the network. It allows the decoder to model a more fine-grained mapping between the target face and 2D plane.

Finally, the output features of the decoder are fed into an MLP for regressing the coordinates of the mapped anchors.

\begin{figure}[t!]
	\centering
	\includegraphics[width=\linewidth]{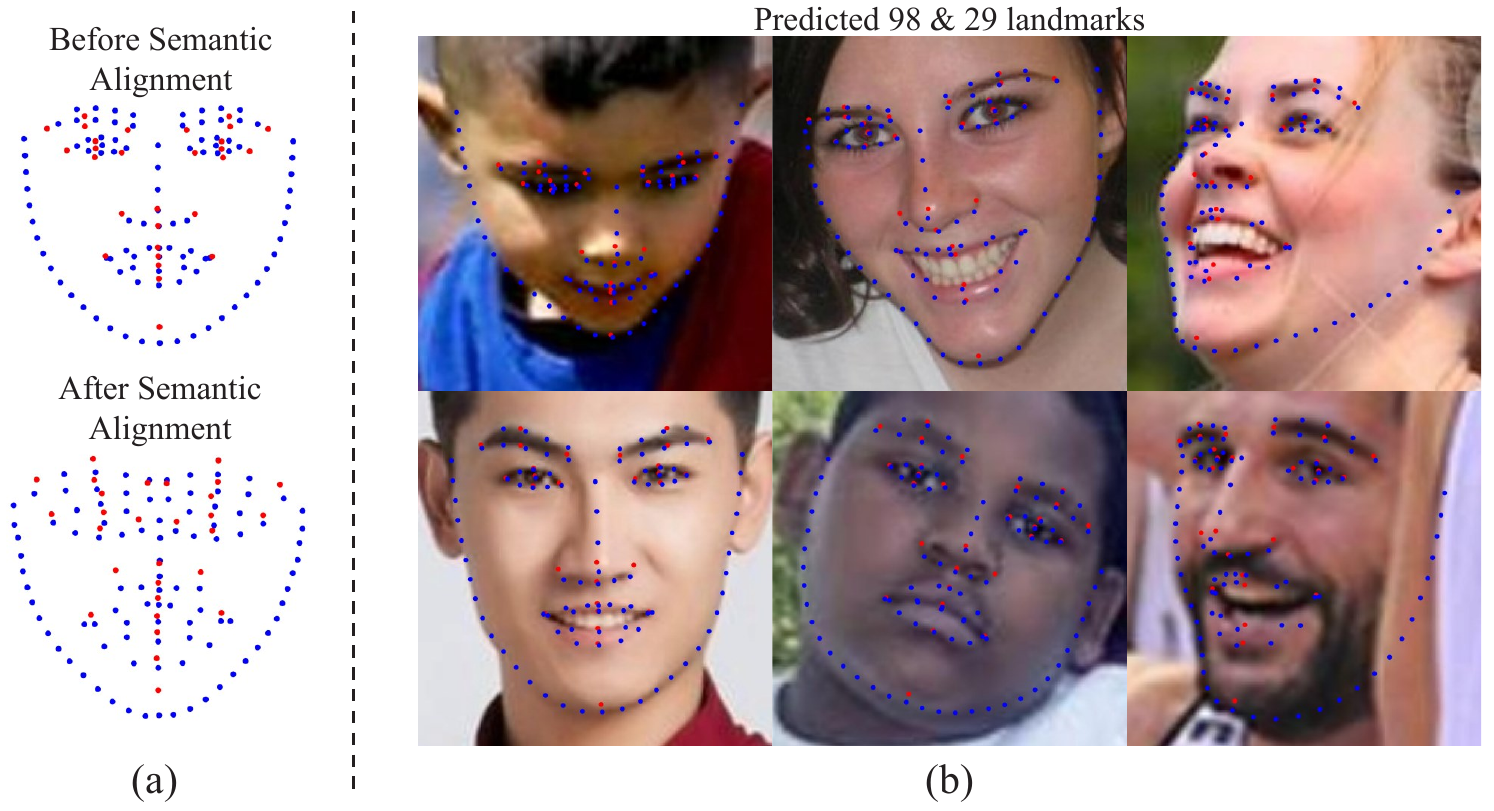}
	\caption{(a) the mean shapes of 29 \& 98 landmarks before/after adding semantic alignment embeddings $\bm{A}$. (b) the predicted 29 \& 98 landmarks on the test image. The \textcolor[rgb]{1, 0, 0}{\textbf{red}} and \textcolor[rgb]{0, 0, 1}{\textbf{blue}} points indicate the 29 landmarks and 98 landmarks respectively.}  
	\label{fig4}
\end{figure}

\subsection{Semantic Alignment Embedding}
The calculated mean shape reflects the semantic definitions and statistically geometric relationships between the pre-defined landmarks. Based on the fact that human faces have a regular structure, the semantics of pre-defined landmarks across different datasets can be aligned on the same 2D plane, regardless of the variations in landmark definition and number. This alignment effectively mitigates the knowledge discrepancies among multiple datasets and unifies their learning targets.

Therefore, we introduce extra learnable semantic embeddings $\bm{A}_i \in \mathbb{R}^{N_{\rm D}^i \times 2}$ to the mean shape of each dataset, where $i$ is the index of dataset and ${N_{\rm D}^i}$ is the number of the pre-defined landmarks in the $i$-th dataset. $\bm{A}$ ensures that each landmark of the mean shapes has a corresponding learnable 2D offset. As shown in Eq. 4 and Eq. 5, adding $\bm{A}$ to the meanshape equals to applying a linear transformation to the initial structure prompts. Because the coordinates of mean shapes are normalized in the range of $[-1, 1]$, the transformation matrices are continuous and unique for any offset. It ensures that the structure prompts after linear transformation are also continuous and unique. As a result, it is possible for TUFA to find an optimal offset for the semantic alignment. 

As shown in Fig. 4 (a), after adding the semantic alignment embeddings, the mean shapes of both the 98 and 29 landmarks still retain a clear face structure, which illustrates that these semantic alignment embeddings specifically incorporate the face structure information. The predicted results shown in Fig. 4 (b) demonstrates that even the landmarks from different datasets, defined similarly, still exhibit semantic variance due to different annotation methods. For instance, the eye corners in the 29 landmarks are always located above the eye corners in the 98 landmarks. This can be also viewed as a kind of knowledge discrepancies. Therefore, directly using unaligned mean shapes for training may lead to semantic ambiguity and degrade performance further. After semantic alignment, the geometric relationship between the predicted 29 \& 98 landmarks is consistent with that of the mean shapes of 29 \& 98 landmarks, and the knowledge discrepancies are largely mitigated. Therefore, with the semantic alignment embeddings, the semantics of the landmarks from different datasets have been aligned on a plane successfully.

\subsection{Loss Function}

When the input anchors are dense enough, TUFA can learn a mapping from a 2D plane to a very complex surface by minimizing the distance between the mapped anchors and the corresponding labeled landmarks. We measure the distance using L1 loss as follows:
\begin{small}
\begin{equation}
	\mathcal{L}=\frac{1}{N_{\rm batch}N_{\rm a}}\sum_{j=1}^{N_{\rm batch}}\sum_{k=1}^{N_{\rm a}}\left| (x_{\rm map}^{jk}, y_{\rm map}^{jk}) -  (x_{\rm gt}^{jk}, y_{\rm gt}^{jk}) \right|,
\end{equation}
\end{small}where $N_{\rm batch}$ is the batch size, $(x_{\rm map}^{jk}, y_{\rm map}^{jk})$ is the position of the $k$-th mapped anchor in $j$-th sample, and $(x_{\rm gt}^{jk}, y_{\rm gt}^{jk})$ is the corresponding labeled position.

\begin{figure}[t!]
	\centering
	\includegraphics[width=\linewidth]{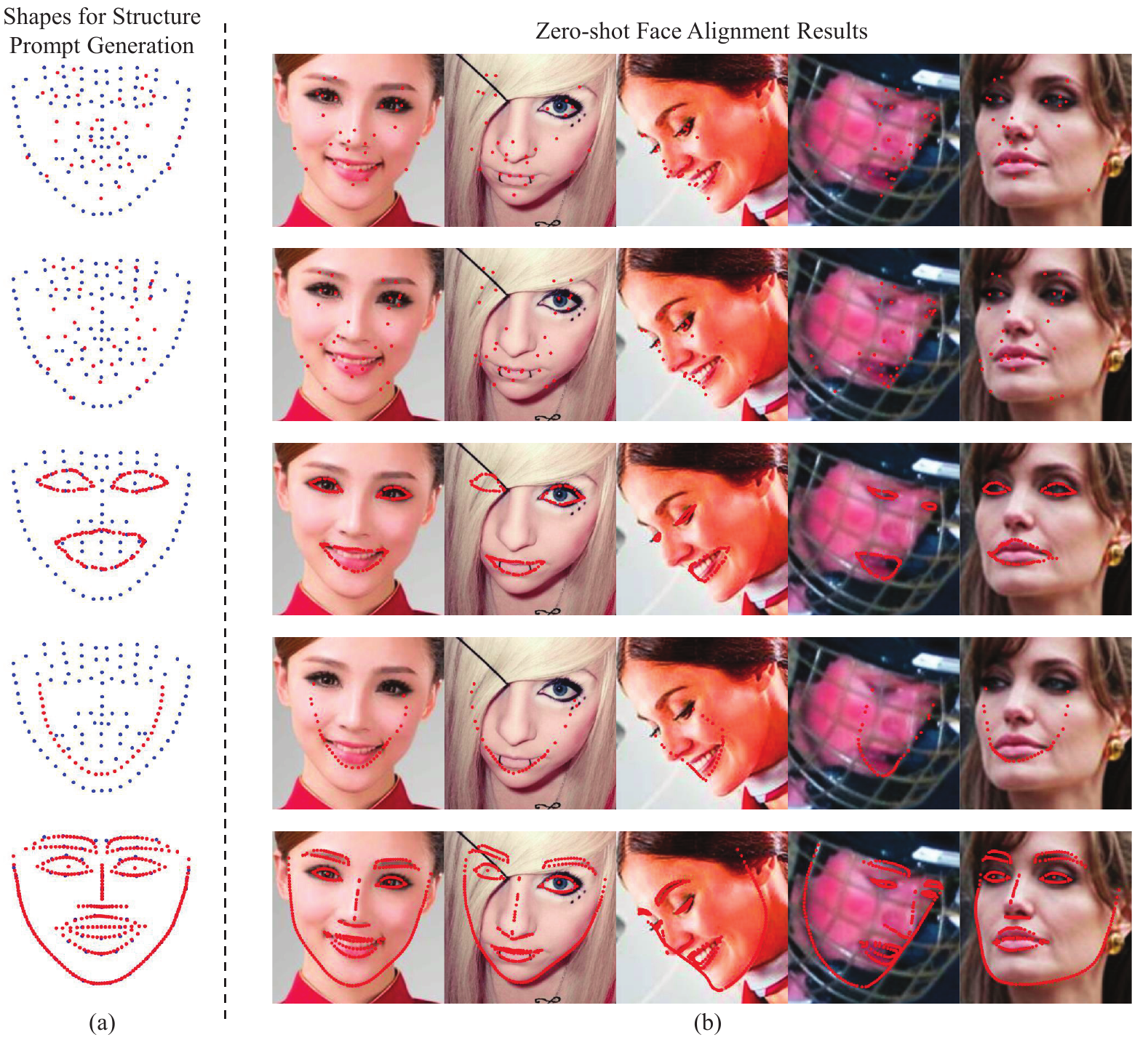}
	\caption{(a) the \textcolor[rgb]{1, 0, 0}{\textbf{red}} points are used to generate face structure prompts for zero-shot face alignment. The \textcolor[rgb]{0, 0, 1}{\textbf{blue}} points represent the mean shape of 98 landmarks, which serves as anchors during training. (b) the \textcolor[rgb]{1, 0, 0}{\textbf{red}} points represent the zero-shot face alignment results under various conditions.}  
	\label{fig5}
\end{figure}

\subsection{Zero-shot \& Few-shot Face Alignment}
The transformer-based decoder enables TUFA to accept an arbitrary number of structure prompts as the input, and the interpretable plane, learned based on the statistical mean shape, allows for easy editing of these structure prompts. Leveraging the structure prompts, the gap between the seen and unseen landmarks can be bridged. To locate an unseen landmark, we can determine its corresponding position on the plane, based on its geometric relationship with the anchors used during training. By encoding this 2D coordinate into a structure prompt, TUFA can locate the coordinate of the unseen landmark across different faces. As shown in Fig. 5, we randomly generate three scratch shapes and locate the corresponding landmarks on the target faces. We can clearly observe that the geometric relationship between the scratch shapes and the training anchors on the plane remains consistent with that geometric relationship on the target faces. Moreover, the semantics of the located landmarks also keep consistent across different faces. This demonstrates the editability of the predicted target in TUFA, as well as the successful implementation of zero-shot face alignment.

The unified learning target also enables TUFA to be easily transferred to a group of newly defined facial landmarks in a few-shot manner. We first calculate a mean shape from the very few training samples. This mean shape is then added with a group of newly defined learnable semantic alignment embeddings. Finally, the pretrained TUFA is fine-tuned end-to-end on the new dataset without any changes of the structure. Despite the semantic difference in the newly defined landmarks, TUFA can still be efficiently transferred to the dataset. This is because TUFA explicitly learns the face structure, unifying the learning target across different datasets. Therefore, TUFA can easily inherit the knowledge learned from pre-trained datasets in few-shot learning, regardless of the differences in landmark semantics. Even with fewer training samples, TUFA significantly outperforms other state-of-the-art few-shot face alignment methods.

\section{Experiments}\label{sec4}

\subsection{Datasets}
\begin{itemize}
	\item \textbf{WFLW} \citep{LAB}: WFLW consists of 10,000 faces (7,500 for training and 2,500 for testing) from WIDER Face \citep{WIDER}. Each face is fully manual annotated with 98 landmarks and attribute labels. Compared to other datasets, the WFLW is more challenging because it contains a large number of samples under extreme conditions, such as heavy occlusion and profile view.
	\item \textbf{300W} \citep{300W}: 300W consists of 3,837 faces (3,148 for training and 689 for testing) from AFW \citep{AFW}, HELEN \citep{HELEN}, LFPW \citep{LFPW} and IBUG \citep{300W}. Each face is annotated with 68 landmarks using a semi-automatic methodology \citep{300W-A}. The testing set can be further divided into a challenging subset (135 faces) and a common subset (554 faces). Moreover, 300W also provides additional 600 faces named as \textbf{300W-private} to test the generalization ability of the trained model.
	\item \textbf{Masked 300W} \citep{SAAT}: Masked 300W is a variant of 300W \citep{300W} used exclusively for testing the performance of face alignment under heavy occlusion. It synthesizes 689 masked faces from the testing set of 300W, and the landmark annotation remains consistent with the original 300W.
	\item \textbf{MERL-RAV} \citep{LUVLI}: MERL-RAV manually re-annotates 19,314 faces from AFLW \citep{AFLW} with 68 landmarks, providing 15,449 samples for training and 3,865 samples for testing. MERL-RAV further categorizes facial landmarks into unoccluded, externally occluded and self-occluded landmarks. Only unoccluded and externally occluded landmarks are provided with location information.
	\item \textbf{COFW} \citep{COFW}: COFW collects 1,007 challenging faces with an average occlusion of over 23\% from a variety of sources. It uses 500 of these faces and 845 samples from LFPW \citep{LFPW} training set for training, and tests on the remaining 507 faces. Each face is annotated with 29 landmarks.
	\item \textbf{COFW68} \citep{COFW68}: COFW68, a variant of COFW \citep{COFW}, is used for cross-dataset validation. The testing set of COFW is manually re-annotated with 68 landmarks in this variant.
	\item \textbf{CelebA-aligned} \citep{CelebA}: CelebA-aligned contains 10,000 identities with 200,000 face images. Each face is labeled with 5 landmarks and aligned to center. As suggested by \citep{Unsuper_Thewlis}, CelebA can be subdivided into three subsets: CelebA training set without MAFL (160,000 images), MAFL training set (19,000 images) and MAFL testing set (1,000 images).
\end{itemize}

\begin{table*}[t!]
	\centering
	\caption{Performance comparison against state-of-the-art heatmap regression and coordinate regression methods on WFLW and its subsets. Corresponding parameter size, flops are also reported. Key: [{\color{red} \textbf{Best}}, {\color{blue} \textbf{Second Best}}, $^\star$=\textbf{trained with multiple datasets}]}
	\resizebox{\textwidth}{!}{
		\begin{tabular}{m{4.0cm}<{\centering}|m{2.2cm}<{\centering}|m{1.5cm}<{\centering}|m{1.8cm}<{\centering}|m{1.0cm}<{\centering}|m{1.4cm}<{\centering}|m{0.7cm}<{\centering}m{0.7cm}<{\centering}m{0.7cm}<{\centering}m{0.7cm}<{\centering}m{0.7cm}<{\centering}m{0.7cm}<{\centering}m{0.7cm}<{\centering}}
			\hline
			\multirow{2}{*}{Method} & \multirow{2}{*}{Backbone} & \multirow{2}{*}{type} & Extra face & \multirow{2}{*}{Flops $\downarrow$} & \multirow{2}{*}{Params $\downarrow$} & \multicolumn{7}{c}{Inter-Ocular NME (\%) $\downarrow$} \\
			& & &images & & & Full & Pose  & Exp. & Ill. & Mu. & Occ. & Blur  \\ \hline
			3FabRec \citep{3FabRec}$^\star$ & ResNet18 & heatmap & 2.1 millions & - & - & 5.62 & 10.23 & 6.09 & 5.55 & 5.68 & 6.92 & 6.38 \\ 
			ATF \citep{ATF}$^\star$ & HRNetW18C & heatmap & 20,000 & 4.75G & 9.66M & 4.50 & 7.54 & 4.63 & 4.45 & 4.20 & 5.30 & 5.19 \\
			PIPNet \citep{PIP} & ResNet101 & heatmap & N/A & 10.5G & 45.7M & 4.31 & 7.51 & 4.44 & 4.19 & 4.02 & 5.36 & 5.02 \\
			AWing \citep{Awing} & 4 Hourglass& heatmap & N/A & 26.8G & 24.15M & 4.21 & 7.21 & 4.46 & 4.23 & 4.02 & 4.99 & 4.82 \\
			HIH \citep{HIH} & 2 Hourglass& heatmap & N/A & 10.38G & 14.47M & 4.18 & 7.20 & {\color{blue} $\bm{4.19}$} & 4.45 & 3.97 &  5.00 & 4.81 \\ 
			ADNet \citep{ADNet} & 4 Hourglass& heatmap & N/A & 17.04G & 13.37M & 4.14 & 6.96 & 4.38& 4.09 & 4.05 & 5.06 & 4.79 \\ 
			FaRL \citep{FaRL}$^\star$ & ViT-B/16 & heatmap & 20 millions & - & - & 4.03 & 6.81 & 4.32 & 3.92 & 3.87 & {\color{blue} $\bm{4.70}$} & {\color{blue} $\bm{4.54}$} \\ 
			STAR \citep{StarLoss} & 4 Hourglass & heatmap & N/A & - & 13.37M & 4.02 & 6.79 & 4.27 & 3.97 & {\color{blue} $\bm{3.83}$} & 4.80 & 4.58 \\ \hline
			AV w. SAN \citep{AVS}$^\star$ & ResNet152& coordinate & 120,000 & 33.87G & 35.02M & 4.39 & 8.42 & 4.68 & 4.24 & 4.37 & 5.60 & 4.86 \\ 
			SDFL \citep{SCDF} & HRNetW18C& coordinate & N/A & 5.17G & - & 4.35 & 7.42 & 4.63 & 4.29 & 4.22 & 5.19 & 5.08 \\ 
			SDL \citep{SDL} & HRNetW18C& coordinate & N/A & - & - & 4.21 & 7.36 & 4.49 & 4.12 & 4.05 & 4.98 & 4.82 \\ 
			SLPT \citep{SLPT} & HRNetW18C-lite& coordinate & N/A & 6.12G & 13.19M & 4.14 & 6.96 & 4.45 & 4.05 & 4.00 & 5.06 & 4.79  \\
			SPIGA \citep{SPIGA} & 4 Hourglass & coordinate & N/A & - & - & 4.06 & 7.14 & 4.46 & 4.00 & {\color{red} $\bm{3.81}$} & 4.95 & 4.65 \\
			DSLPT \citep{DSLPT} & HRNetW18C & coordinate & N/A & 7.83G & 19.35M & 4.01 & 6.87 & 4.29 & 3.99 & 3.86 & 4.79 & 4.66 \\ \hline
			TUFA & ViT-S/16 & coordinate & N/A & 8.043G & 36.00M & 4.38 & 7.51 & 4.72 & 4.32 & 4.32 & 5.22 & 4.91 \\
			TUFA$^\star$ & ViT-S/16 & coordinate & 19,942 & 8.043G & 36.00M & {\color{blue} $\bm{4.00}$} & {\color{blue} $\bm{6.57}$} & 4.20 & {\color{blue} $\bm{3.90}$} & 3.86 & 4.73 & 4.55 \\
			TUFA & ViT-S/8 & coordinate & N/A & 35.17G & 36.03M & 4.23 & 7.23 & 4.53 & 4.17 & 4.12 & 5.05 & 4.78 \\
			TUFA$^\star$ & ViT-S/8 & coordinate & 19,942 & 35.17G & 36.03M & {\color{red}$\bm{3.93}$} & {\color{red}$\bm{6.48}$} & {\color{red}$\bm{4.11}$} & {\color{red}$\bm{3.82}$} & {\color{red}$\bm{3.81}$} & {\color{red}$\bm{4.68}$} & {\color{red}$\bm{4.53}$} \\
			\hline
		\end{tabular}
	}
	\label{Table1}
\end{table*}

\begin{table*}[t!]
	\centering
	\caption{Evaluation results in FR$_{0.1}$ and AUC$_{0.1}$ on WFLW. Key: [{\color{red} \textbf{Best}}, {\color{blue} \textbf{Second Best}}, $^\star$=\textbf{trained with multiple datasets}]}
	\resizebox{\textwidth}{!}{
		\begin{tabular}{m{3.0cm}<{\centering}|m{0.8cm}<{\centering}m{0.8cm}<{\centering}m{0.8cm}<{\centering}m{0.8cm}<{\centering}m{0.8cm}<{\centering}m{0.8cm}<{\centering}m{0.8cm}<{\centering}|m{0.8cm}<{\centering}m{0.8cm}<{\centering}m{0.8cm}<{\centering}m{0.8cm}<{\centering}m{0.8cm}<{\centering}m{0.8cm}<{\centering}m{0.8cm}<{\centering}}
			\hline
			\multirow{2}{*}{Method} & \multicolumn{7}{c|}{FR$_{0.1}$(\%)$\downarrow$} & \multicolumn{7}{c}{AUC$_{0.1}$$\uparrow$} \\
			& Full & Pose  & Exp. & Ill. & Mu. & Occ. & Blur & Full & Pose  & Exp. & Ill. & Mu. & Occ. & Blur \\ \hline
			3FabRec$^\star$ & 8.28 & 34.35 & 8.28 & 6.73 & 10.19 & 15.08 & 9.44 & 0.484 & 0.192 & 0.448 & 0.496 & 0.473 & 0.398 & 0.434 \\
			ATF$^\star$ & 2.52 & 13.19 & 2.23 & 2.44 & {\color{red} $\bm{0.49}$} & 5.03 & 3.88 & 0.560 & 0.301 & 0.546 & 0.566 & 0.581 & 0.487 & 0.489 \\
			PIPNet & - & - & - & - & - & - & - & - & - & - & - & - & - & - \\
			AWing & 2.04 & 9.20 & {\color{red} $\bm{1.27}$} & 2.01 & {\color{blue} $\bm{0.98}$} & 4.21 & 2.72 & 0.590 & 0.334 & 0.572 & 0.596 & 0.602 & 0.528 & 0.539\\
			HIH & 2.96 & 15.03 & {\color{blue} $\bm{1.59}$} & 2.58 & 1.46 & 6.11 & 3.49 & 0.597 & 0.342 & {\color{blue}$\bm{0.590}$} & 0.606 & 0.604 & 0.527 & 0.549 \\
			ADNet & 2.72 & 12.72 & 2.15 & 2.44 & 1.94 & 5.79 & 3.54 & 0.602 & 0.344 & 0.523 & 0.580 & 0.601 & 0.530 & 0.548 \\
			FaRL$^\star$ & 1.76 & - & - & - & - & - & - & 0.602 & - & - & - & - & - & - \\ 
			STAR & 2.32 & 11.69 & 2.24 & 1.58 & {\color{blue} $\bm{0.98}$} & 4.76 & 3.24 & 0.605 & {\color{blue} $\bm{0.362}$} & 0.584 & 0.609 & {\color{blue} $\bm{0.622}$} & {\color{blue} $\bm{0.538}$} & {\color{blue} $\bm{0.551}$} \\ \hline
			AV w. SAN$^\star$ & 4.08 & 18.10 & 4.46 & 2.72 & 4.37 &7.74 & 4.40 & 0.591 & 0.311 & 0.549 & 0.609 & 0.581 & 0.516 & {\color{blue}$\bm{0.551}$} \\
			SDFL & 2.72 & 12.88 & {\color{blue} $\bm{1.59}$} & 2.58 & 2.43 & 5.71 & 3.62 & 0.576 & 0.315 & 0.550 & 0.585 & 0.583 & 0.504 & 0.515 \\ 
			SDL & 3.04 & 15.95 & 2.86 & 2.72 & 1.46 & 5.29 & 4.01 & 0.589 & 0.315 & 0.566 & 0.595 & 0.604 & 0.524 & 0.533 \\ 
			SLPT & 2.76 & 12.27 & 2.23 & 1.86 & 3.40 & 5.98 & 3.88 & 0.595 & 0.348 & 0.574 & 0.601 & 0.605 & 0.515 & 0.535 \\  
			SPIGA & 2.08 & 11.66 & 2.23 & 1.58 & 1.46 & 4.48 & {\color{red} $\bm{2.20}$} & 0.606 & 0.353 & 0.580 & 0.613 & {\color{blue} $\bm{0.622}$} & 0.533 & {\color{red} $\bm{0.553}$} \\
			DSLPT & 2.52 & 13.19& 2.23 & 2.44 & {\color{blue} $\bm{0.98}$} & 4.89 & 3.49 & {\color{blue} $\bm{0.607}$} & 0.353 & 0.586 & {\color{blue} $\bm{0.614}$} & {\color{red} $\bm{0.623}$} & 0.535 & 0.549 \\ \hline
			TUFA (ViT-S/16) & 2.92 & 14.42 & 3.50 & 2.29 & 2.43 & 5.84 & 4.53 & 0.570 & 0.296 & 0.540 & 0.579 & 0.571 & 0.497 & 0.524 \\
			TUFA (ViT-S/16)$^\star$ & {\color{blue} $\bm{1.72}$} & {\color{blue} $\bm{8.28}$} & {\color{blue} $\bm{1.59}$} & {\color{blue} $\bm{1.29}$} & 1.46 & {\color{blue} $\bm{3.53}$} & 2.59 & 0.604 & 0.359 & 0.584 & 0.613 & 0.616 & 0.534 & {\color{blue}$\bm{0.551}$} \\
			TUFA (ViT-S/8) & 2.44 & 12.27 & 1.91 & 2.15 & 1.46 & 5.30 & 3.49 & 0.585 & 0.321 & 0.553 & 0.593 & 0.591 & 0.513 & 0.535 \\
			TUFA (ViT-S/8)$^\star$ & {\color{red} $\bm{1.52}$} & {\color{red} $\bm{7.36}$} & {\color{red} $\bm{1.27}$} & {\color{red} $\bm{0.86}$} & 1.46 & {\color{red} $\bm{2.99}$} & {\color{blue} $\bm{2.33}$} & {\color{red} $\bm{0.610}$} & {\color{red} $\bm{0.371}$} & {\color{red}$\bm{0.592}$} & {\color{red} $\bm{0.620}$} & {\color{blue} $\bm{0.622}$} & {\color{red} $\bm{0.540}$} & {\color{red} $\bm{0.553}$} \\ \hline 
			
		\end{tabular}
	}
	\label{Table2}
\end{table*}

\subsection{Evaluation Metrics}
Following previous works \citep{LAB}, \citep{ADNet}, \citep{SLPT}, \citep{DSLPT}, we use Normalized Mean Error (NME), Failure Rate (FR) and Area Under Curve (AUC) to quantitatively measure the performance of TUFA. NME is the mean of L2 distance between the predicted landmarks and the annotated landmarks. The mean distance is then normalized by a factor, denoted as $d_{\rm norm}$. For \textit{inter-pupil} NME, $d_{\rm norm}$ represents the distance between pupil centers, while for \textit{inter-ocular} NME, it stands for the distance between outer eye corners. In the case of NME$_{\rm box}$, $d_{\rm norm}$ is defined as the geometric mean of the labeled box, calculated as $d_{\rm norm} = \sqrt{W_{\rm box} \times H_{\rm box}}$. FR$_{\alpha}$ represents the percentage of the testing cases in which the NME exceeds a certain threshold, denoted as $\alpha$. AUC indicates the area beneath the Cumulative Errors Distribution (CED) curve from 0 to the threshold of FR $\alpha$, which can be formulated as $\int_{0}^{\alpha}f \left( \epsilon \right) d \epsilon$.

\subsection{Implementation Details}

The input image is the face region cropped from the initial image and is then resized to a fixed size ($256\times256$). For training data, we apply augmentation techniques, which include random translation ($\pm10$ pixels), random rotation ($\pm30^\circ$), random scaling ($\pm5\%$), random horizontal flipping ($50\%$), random gray ($20\%$), random brightness adjustment ($50\%, \pm0.3$), random occlusion ($50\%$), random shearing ($33\%$). In TUFA, We employ two types of encoders : ViT-S/16 and ViT-S/8 \citep{VIT}, both of which are pretrained on ImageNet \citep{ImageNet} with DINO \citep{DINO}. The parameters of the decoder and semantic alignment embeddings are initialized from scratch. By default, the batch size is set to 16 and the layer number of the decoder is set to 6. TUFA is trained with AdamW \citep{AdamW}, setting the initial learning rate to $1\times10^{-4}$ for ViT-S/8 and $5\times10^{-5}$ for ViT-S/16. For multi-dataset learning, we employ the training sets of WLFW, 300W, MERL-RAV and COFW for training. The model is trained for 100 epochs, with the learning rate decaying by a factor of 0.1 at the 80th and 90th epochs respectively. The learned result is then tested on the WFLW, 300W, 300W-private, masked 300W, MERL-RAV, COFW and COFW68 \textbf{simultaneously}. For single-dataset learning, due to the smaller number of training samples, the model is trained for 140 epochs, with the learning rate decaying by a factor of 0.1 at the 120th and 130th epochs respectively.

\subsection{Comparisons in Within-dataset Validation}

\textbf{WFLW}: we report the inter-ocular NME, parameter size and flops of TUFA and other state-of-the-art methods in Table 1. FR$_{0.1}$ and AUC$_{0.1}$ are tabulated in Table 2. Without increasing any computational complexity, the ability of TUFA o mitigate knowledge discrepancies across multiple datasets maximizes the utilization of extra training samples. We observe a significant improvement of 7.09\% in NME on the full set compared to TUFA (ViT-S/8) trained on a single dataset, even though TUFA uses fewer extra samples than other methods trained on multiple datasets.

Considering efficiency, we also implemented a TUFA with a lighter backbone, ViT-S/16. Similarly, training on multiple datasets achieves an impressive improvement of 8.68\% in NME on the full set compared to single dataset training. Despite having much lower computational complexity (flops) than most heatmap regression methods, it still yields the second best performance in NME and FR$_{0.1}$. It also outperforms DSLPT in NME on the full set and all subsets with comparable flops. Moreover, the entire training process for this TUFA can be completed within five hours on a single A40 GPU, demonstrating the high efficiency of our method.

\begin{figure}[t!]
	\centering
	\includegraphics[width=\linewidth]{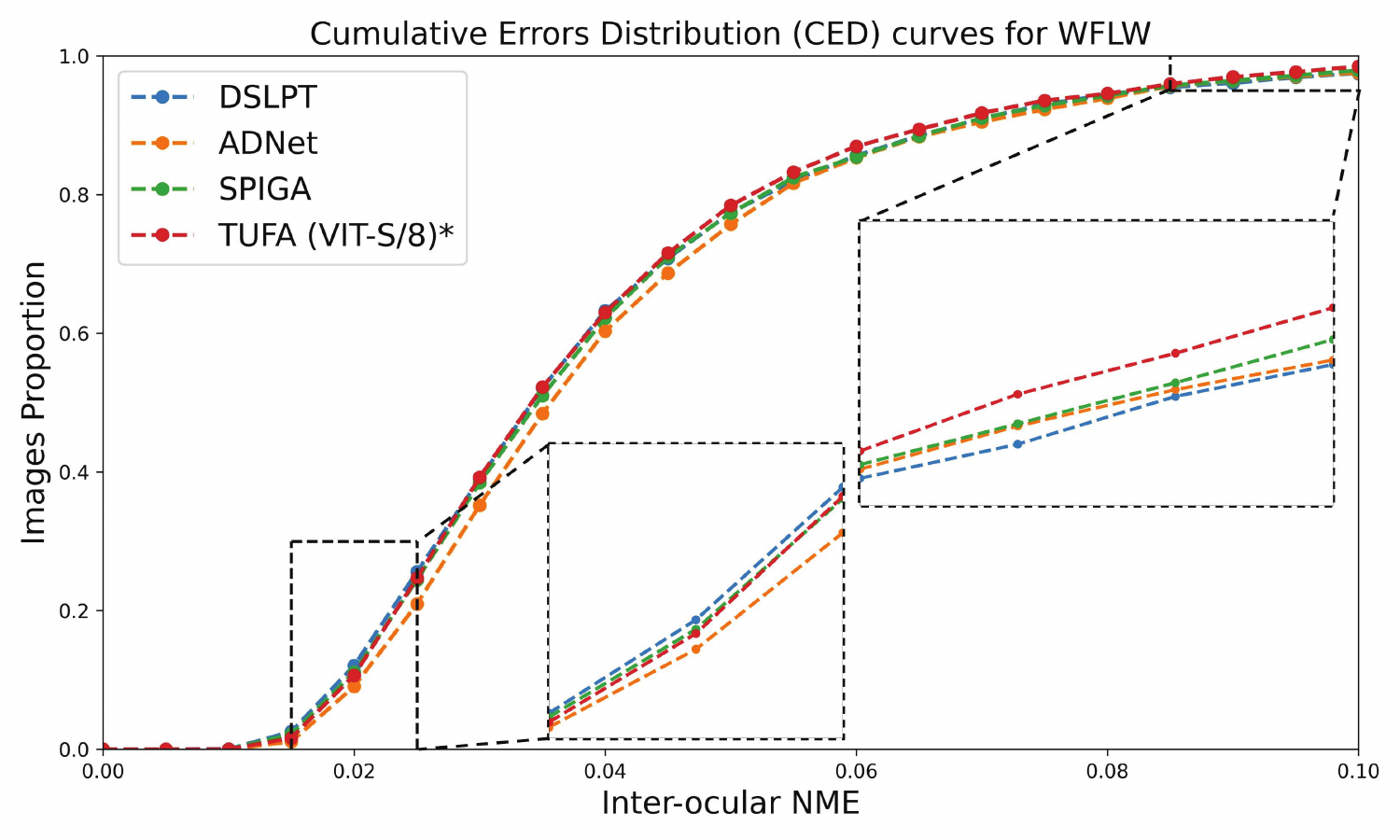}
	\caption{The comparisons of CED curves between TUFA and other state-of-the-art methods on WFLW full set.}  
	\label{fig6}
\end{figure}

Most existing methods, such as DSLPT, SPIGA, ADNet, and STAR, achieve competitive performance based on a multi-stage approach, while TUFA locates facial landmarks using only a single stage. For a comprehensive analysis, we plot the CED curves of several multi-stage methods and TUFA (using multi-dataset learning) in Fig.6. Note that the official implementation of ADNet performs slightly better than the results reported in their original paper. The CED curves of DSLPT and SPIGA are higher than that of TUFA in the range of $[0, 0.025]$. It suggests the proportion of the samples with very small NME predicted by DSLPT and SPIGA is larger than that of TUFA, despite that TUFA (using multi-dataset learning) performs better in NME. This indicates that these multi-stage face alignment methods primarily reduce the NME on easy samples for better numerical results. However, their improvements under challenging conditions are relatively insignificant. Despite the fact that these conditions are long-tailed, they are critical as they ultimately determine the quality of face alignment in real-world scenarios. By employing extra data, TUFA significantly improves the performance on these challenging conditions. As a result, TUFA achieves improvements of 26.9\%, 36.9\% and 33.3\% in FR$_{0.1}$ on full set, largepose subset and occlusion subset respectively compared to SPIGA.

\begin{table}[t!]
	\caption{Performance comparison against state-of-the-art methods on 300W. Key: [{\color{red} \textbf{Best}}, {\color{blue} \textbf{Second Best}}, $^\star$=\textbf{trained with multiple datasets}]}
	\begin{tabular}{m{3.0cm}<{\centering}|m{0.8cm}<{\centering}m{1.2cm}<{\centering}m{0.8cm}<{\centering}}
		\hline
		\multirow{2}{*}{Method} &  \multicolumn{3}{c}{Inter-Ocular NME (\%) $\downarrow$} \\
		&  Common & Challenging & Fullset \\ \hline
		3FabRec \citep{3FabRec}$^\star$ & 3.36 & 5.74 & 3.82 \\
		HIH \citep{HIH} & 2.93 & 5.00 & 3.33 \\
		PIPNet \citep{PIP} & 2.78 & 4.89 & 3.19 \\ 
		ATF \citep{ATF}$^\star$ & 2.75 & 4.86 & 3.17 \\
		AWing \citep{Awing} & 2.72 & 4.53 & 3.07 \\
		ADNet \citep{ADNet} & {\color{blue} \textbf{2.53}} & 4.58 & {\color{red} \textbf{2.93}} \\ 
		SLPT \citep{SLPT} & 2.75 & 4.90 & 3.17 \\ 
		FaRL \citep{FaRL}$^\star$ & 2.70 & 4.64 & 3.08 \\
		SPIGA \citep{SPIGA} & 2.59 & 4.66 & 2.99 \\ 
		DSLPT \citep{DSLPT} & 2.57 & 4.69 & 2.98 \\
		STAR \citep{StarLoss} & {\color{red} \textbf{2.52}} & {\color{red} \textbf{4.32}} & 2.87 \\
		\hline
		TUFA (ViT-S/16) & 2.92 & 5.05 & 3.34 \\
		TUFA (ViT-S/16)$^\star$ & 2.68 & 4.58 & 3.05 \\
		TUFA (ViT-S/8) & 2.82 & 4.90 & 3.23 \\
		TUFA (ViT-S/8)$^\star$ & 2.59 & {\color{blue} \textbf{4.45}} & {\color{blue} \textbf{2.95}} \\ \hline
	\end{tabular}
	\label{Table3}
\end{table}

\textbf{300W}: as tabulated in Table 3, multi-dataset learning also improves the performance of TUFA on common set, challenging set and full set by 8.16\%, 9.18\% and 8.67\% respectively in the metric of NME. Heatmap regression methods, such as ADNet, Awing and STAR, commonly demonstrate superior performance on 300W, especially the challenging subset, compared to coordinate regression methods. This is because they provide semantic supervision to the network by encoding the annotated landmarks into heatmaps, which delivers better performance with limited training samples. However, the unique properties of TUFA successfully address this limitation of coordinate regression methods, achieving the best performance in NME on the challenging set, as low as 4.45\%, even though TUFA is a single-stage face alignment method. This demonstrates that the multi-dataset learning of TUFA promises a very competitive generalization ability. Compared to FaRL and ATF, which are also trained with multiple face datasets, TUFA outperforms them by 4.22\% and 6.94\% respectively in NME despite fewer extra faces used in training.

\begin{table}[t!]
	\centering
	\caption{Performance comparison against state-of-the-art methods on COFW. The FR$_{0.1}$ in Inter-Ocular and Inter-Pupil normalization are reported. Key: [{\color{red} \textbf{Best}}, {\color{blue} \textbf{Second Best}}, $^\star$=\textbf{trained with multiple datasets}]}
	\begin{tabular}{m{2.6cm}<{\centering}|m{0.7cm}<{\centering}m{0.7cm}<{\centering}|m{0.7cm}<{\centering}m{0.7cm}<{\centering}}
		\hline
		\multirow{2}{*}{Method} &  \multicolumn{2}{c|}{Inter-Ocular} & \multicolumn{2}{c}{Inter-Pupil} \\
		& NME$\downarrow$ & FR $\downarrow$ & NME$\downarrow$ & FR$\downarrow$ \\  \hline
		SDFL \citep{SCDF} & 3.63\% & {\color{red} $\bm{0.00\%}$} & - & - \\
		ATF \citep{ATF}$^\star$ & 3.32\% & - & - & -\\
		AWing \citep{Awing} &-&-& 4.94\% & 0.99\% \\
		ADNet \citep{ADNet} &-&-& 4.68\% & {\color{blue} $\bm{0.59\%}$} \\ 
		SLPT \citep{SLPT}  & 3.32\% & {\color{red} $\bm{0.00\%}$} & 4.79\% & 1.18\% \\
		DSLPT \citep{DSLPT} & 3.33\% & {\color{blue} $\bm{0.20\%}$} & 4.79\% & 1.36\% \\ 
		STAR \citep{StarLoss} & 3.21\% & {\color{red} $\bm{0.00\%}$} & 4.62\% & 0.79\% \\
		PIP \citep{PIP} & {\color{blue} $\bm{3.08\%}$} & - & - & - \\ \hline
		TUFA (ViT-S/16) & 3.49\% & {\color{blue} $\bm{0.20\%}$} & 5.03\% & 0.79\% \\
		TUFA (ViT-S/16)$^\star$ & 3.18\% & {\color{blue} $\bm{0.20\%}$} & {\color{blue} $\bm{4.58\%}$} & {\color{red} $\bm{0.39\%}$} \\
		TUFA (ViT-S/8) & 3.40\% & {\color{blue} $\bm{0.20\%}$} & 4.91\% & 1.18\% \\
		TUFA (ViT-S/8)$^\star$ & {\color{red} $\bm{3.07\%}$} & {\color{blue} $\bm{0.20\%}$} & {\color{red} $\bm{4.43\%}$} & {\color{red} $\bm{0.39\%}$} \\ 
		\hline		
	\end{tabular}
	\label{Table4}
\end{table}

\begin{table*}[t!]
	\centering
	\caption{NME$_{\rm box}$ and AUC$_{\rm box}^{0.07}$ on the MERL-RAV full set and its subsets. The NME$_{\rm box}$ and AUC$_{\rm box}^{0.07}$ of unoccluded and externally occluded landmarks are also reported. Key: [{\color{red} \textbf{Best}}, {\color{blue} \textbf{Second Best}}, $^\star$=\textbf{trained with multiple datasets}]}
	\resizebox{\textwidth}{!}{
		\begin{tabular}{m{3.1cm}<{\centering}|m{1.3cm}<{\centering}m{1.4cm}<{\centering}|m{1.3cm}<{\centering}m{1.4cm}<{\centering}|m{1.3cm}<{\centering}m{1.4cm}<{\centering}|m{1.3cm}<{\centering}m{1.4cm}<{\centering}|m{1.3cm}<{\centering}m{1.4cm}<{\centering}|m{1.3cm}<{\centering}m{1.4cm}<{\centering}}
			\hline
			\multirow{2}{*}{Method} & \multicolumn{2}{c|}{Full Set} & \multicolumn{2}{c|}{Frontal Subset} & \multicolumn{2}{c|}{Half-Profile Subset} & \multicolumn{2}{c|}{Profile SubSet} & \multicolumn{2}{c|}{Unoccluded} & \multicolumn{2}{c}{Externally Occluded} \\
			& NME$_{\rm box}\downarrow$ & AUC$_{\rm box}^{0.07}\uparrow$ & NME$_{\rm box}\downarrow$ & AUC$_{\rm box}^{0.07}\uparrow$ & NME$_{\rm box}\downarrow$ & AUC$_{\rm box}^{0.07}\uparrow$ & NME$_{\rm box}\downarrow$ & AUC$_{\rm box}^{0.07}\uparrow$ & NME$_{\rm box}\downarrow$ & AUC$_{\rm box}^{0.07}\uparrow$ & NME$_{\rm box}\downarrow$ & AUC$_{\rm box}^{0.07}\uparrow$ \\  \hline
			DU-Net \citep{Dunet} & 1.99\% & 71.80\% & 1.89\% & 73.25\% & 2.50\% & 64.78\% & 1.92\% & 72.79\% & - & - & - & - \\
			LUVLi \citep{LUVLI} & 1.61\% & 77.08\% & 1.74\% & 75.33\% & 1.79\% & 74.69\% & 1.25\% & 82.10\% & 1.60\% & - & 3.53\% & - \\
			SLPT \citep{SLPT} & {\color{blue} $\bm{1.51\%}$} & 78.33\% & {\color{blue} $\bm{1.62\%}$} & 76.82\% & {\color{blue} $\bm{1.68\%}$} & {\color{blue} $\bm{76.01\%}$} & 1.21\% & 82.74\% & {\color{blue} $\bm{1.50\%}$} & {\color{blue} $\bm{79.22\%}$} & 3.33\% & {\color{blue} $\bm{56.24\%}$} \\
			SIPGA \citep{SPIGA} & {\color{blue} $\bm{1.51\%}$} & {\color{blue} $\bm{78.47\%}$} & {\color{blue} $\bm{1.62\%}$} & {\color{blue} $\bm{76.96\%}$} & {\color{blue} $\bm{1.68\%}$} & 75.64\% & {\color{blue} $\bm{1.19\%}$} & {\color{blue} $\bm{83.00\%}$} & - & - & - & - \\
			DSLPT \citep{DSLPT} & {\color{red} $\bm{1.48\%}$} & {\color{red} $\bm{78.87\%}$} & {\color{red} $\bm{1.60\%}$} & {\color{red} $\bm{77.24\%}$} & {\color{red} $\bm{1.64\%}$} & {\color{red} $\bm{76.58\%}$} & {\color{red} $\bm{1.16\%}$} & {\color{red} $\bm{83.46\%}$} & {\color{red} $\bm{1.48\%}$} & {\color{red} $\bm{79.37\%}$} & {\color{blue} $\bm{3.26\%}$} & {\color{red} $\bm{57.01\%}$} \\
			\hline
			TUFA (ViT-S/16) & 1.63\% & 76.74\% & 1.75\% & 75.07\% & 1.80\% & 74.31\% & 1.29\% & 81.56\% & 1.65\% & 76.81\% & 3.37\% & 55.30\% \\
			TUFA (ViT-S/16)$^\star$ & 1.61\% & 77.09\% & 1.72\% & 75.52\% & 1.78\% & 74.57\% & 1.28\% & 81.73\% & 1.63\% & 77.11\% & 3.31\% & 56.12\% \\
			TUFA (ViT-S/8) & 1.55\% & 77.93\% & 1.65\% & 76.42\% & 1.72\% & 75.38\% & 1.23\% & 82.46\% & 1.56\% & 78.29\% & 3.30\% & 56.10\% \\
			TUFA (ViT-S/8)$^\star$ & 1.54\% & 78.01\% & 1.65\% & 76.48\% & 1.71\% & 75.50\% & 1.22\% & 82.55\% & 1.56\% & 78.17\% & {\color{red} $\bm{3.21\%}$} & {\color{red} $\bm{57.01\%}$} \\
			\hline
		\end{tabular}
	}
	\label{Table5}
\end{table*}

\textbf{COFW}: with the very limited number of training samples and an average occlusion of over 23\% on the testing faces, COFW presents a significant challenge for all face alignment methods. As shown in Table 4, despite the competitive performance on other datasets, existing state-of-the-art methods, such as DSLPT, ADNet and STAR, often suffer from overfitting on this dataset. Similarly, TUFA, when trained with a single dataset, also tends to overfit the training set, achieving only 3.40\% in the metric of NME. Nevertheless, TUFA successfully mitigates the knowledge discrepancies among multiple datasets and utilizes the knowledge learned from other face datasets, thereby significantly improving the NME metric by 9.71\% and yielding the best performance. Although ATF also leverages extra datasets, it fails to consider the ambiguity between the landmarks with similar definitions, which inadvertently introduces noise into the training and leads to a less significant improvement.

\textbf{MERL-RAV}: the results on MERL-RAV are shown in Table 5. The proportions of unoccluded, externally occluded and self-occluded landmarks in MERL-RAV full set are 76.56\%, 10.81\% and 12.63\% respectively. Because the position annotations of the self-occluded landmarks are not provided, the numerical results on the full set and its subsets exclude the results of these landmarks. Therefore, these numerical results are largely determined by the performance on the easy samples. As mentioned before, multi-stage methods tend to perform better on the easy samples compared to one-stage methods. Therefore, TUFA only demonstrates comparable performance to these multi-stage methods. Nevertheless, when we investigate the externally occluded landmarks separately, we can find that the multi-dataset learning still improves the performance on these challenging landmarks effectively, despite the relatively fewer extra samples provided by WFLW, 300W and COFW compared to the training set of MEAL-RAV. Moreover, the qualitative results shown in Fig. 7 illustrate TUFA performs well on the self-occluded landmarks based on the knowledge of other datasets, even though MERL-RAV does not provide their position labels for training.

\begin{figure}[t!]
	\centering
	\includegraphics[width=\linewidth]{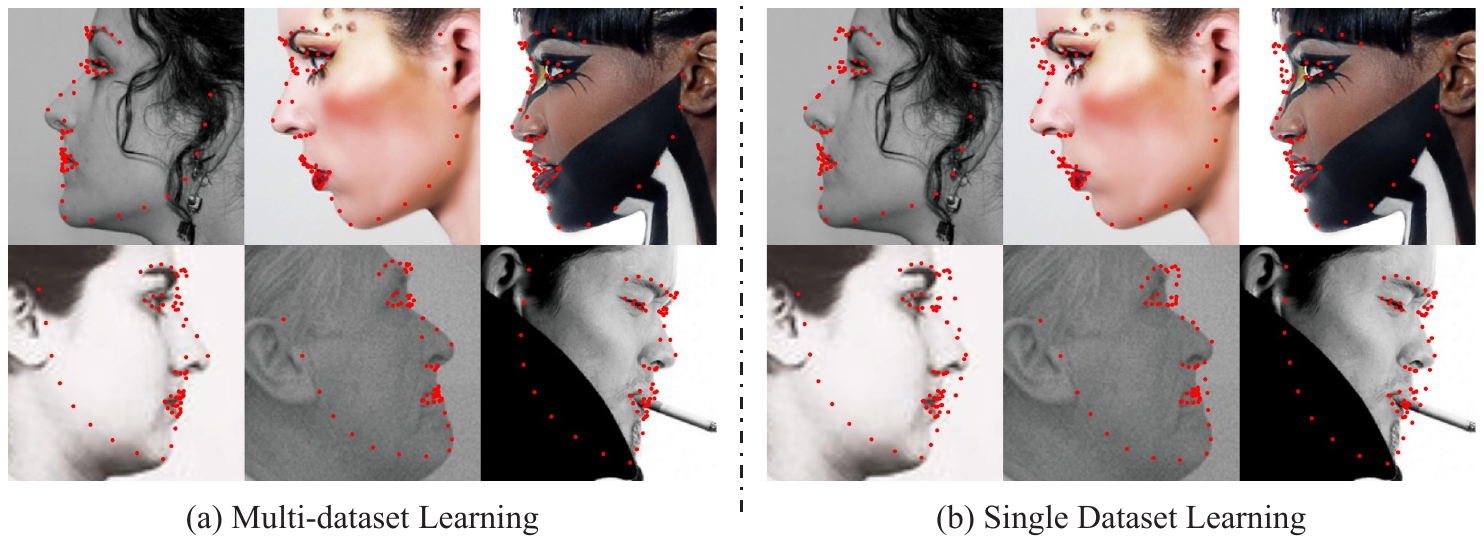}
	\caption{(a) Results in profile view predicted by the TUFA with multi-dataset learning. (b) Results in profile view predicted by the TUFA with single dataset learning. The \textcolor[rgb]{1, 0, 0}{\textbf{red}} points represent the predicted landmarks.}  
	\label{fig7}
\end{figure}

\subsection{Comparisons in Cross-dataset Validation}
\textbf{COFW68}: the cross-dataset validation aims at evaluating the generalization ability of face alignment methods. In the case of TUFA with single dataset learning, the model is trained on 300W training set. For TUFA trained with multiple datasets, the model used in four within-dataset validations is directly evaluated on COFW68. The evaluation results are show in Table 6. Despite not re-training the TUFA for COFW68, it still outperforms existing state-of-the-art methods by a significant margin. Compared to single dataset learning, the multi-dataset learning approach of TUFA demonstrates much stronger generalization ability, achieving an impressive improvement of 9.98\% in NME.

\begin{table}[t!]
	\centering
	\caption{Performance comparison of \textit{cross}-dataset validation on COFW68. The NME and FR$_{0.1}$ under inter-ocular normalization are reported. Key: [{\color{red} \textbf{Best}}, {\color{blue} \textbf{Second Best}}, $^\star$=\textbf{trained with multiple datasets}]}
	\begin{tabular}{m{3.6cm}<{\centering}|m{1.2cm}<{\centering}m{1.2cm}<{\centering}}
		\hline
		Method & NME(\%)$\downarrow$ & FR$_{0.1}$(\%)$\downarrow$ \\ \hline
		AV w. SAN \citep{AVS}$^\star$ & 4.43 & 2.82 \\
		LAB \citep{LAB} & 4.62 & 2.17 \\
		GlomFace \citep{GlomFace} & 4.21 & 0.79 \\
		SDL \citep{SDL} & 4.22 & 0.39 \\
		SDFL \citep{SCDF} & 4.18 & {\color{red} $\bm{0.00}$} \\
		SLPT \citep{SLPT} & 4.10 & 0.59 \\ 
		DSLPT \citep{DSLPT} & 4.03 &  {\color{blue} $\bm{0.20}$} \\ 
		SPIGA \citep{SPIGA} & 3.93 & - \\ \hline
		TUFA (ViT-S/16) & 4.22 &  {\color{blue} $\bm{0.20}$}  \\
		TUFA (ViT-S/16)$^\star$ & {\color{blue} $\bm{3.72}$} &  {\color{red} $\bm{0.00}$} \\
		TUFA (ViT-S/8) & 4.11 &  {\color{blue} $\bm{0.20}$} \\ 
		TUFA (ViT-S/8)$^\star$ & {\color{red} $\bm{3.70}$} &  {\color{red} $\bm{0.00}$} \\ \hline
	\end{tabular}
	\label{Tabel6}
\end{table}

\textbf{Masked 300W}: the setting of cross-dataset validation on Masked-300W is kept consistent with the setting on COFW68, and the evaluation results are displayed in Table 7. The testing faces are with an average occlusion of over 50\%, leading to a significant domain gap between the training and testing faces. To minimize this gap, existing state-of-the-art methods have optimized the data augmentation, randomly masking each training sample with several blocks of varying sizes. Although this approach can improve the quantitative results in cases with heavy occlusion, it also results in performance degradation under other conditions. Without any modification in data augmentation techniques, the TUFA learned from multiple datasets still achieves an impressive result, as low as 5.58\% in NME. This outcome also demonstrates the remarkable generalization ability and robustness of TUFA.

\begin{table}[t!]
	\caption{Performance comparison of \textit{cross}-dataset validation on Masked 300W. Key: [{\color{red} \textbf{Best}}, {\color{blue} \textbf{Second Best}}, $^\star$=\textbf{trained with multiple datasets}, $^\dag$=\textbf{data augmentation adjustment}]}
	\begin{tabular}{m{2.8cm}<{\centering}|m{0.9cm}<{\centering}m{1.2cm}<{\centering}m{0.9cm}<{\centering}}
		\hline
		\multirow{2}{*}{Method} &  \multicolumn{3}{c}{Inter-Ocular NME (\%) $\downarrow$} \\
		&  Common & Challenging & Fullset \\ \hline
		Hourglass \citep{Hourglass} & 8.17 & 13.52 & 9.22 \\
		FAN \citep{3DFAN} & 7.36 & 10.81 & 8.02 \\ 
		LAB \citep{LAB} & 6.07 & 9.59 & 6.76 \\ 
		SAAT \citep{SAAT}$^\dag$ & 5.42 & 11.36 & 6.58 \\
		GlomFace \citep{GlomFace}$^\dag$ & 5.29 & 8.81 & 5.98 \\ 
		DSLPT \citep{DSLPT} & 6.01 & 10.19 & 6.83 \\
		DSLPT \citep{DSLPT}$^\dag$ & {\color{red} $\bm{4.78}$} & {\color{blue} $\bm{8.10}$} & {\color{red} $\bm{5.42}$} \\ \hline
		TUFA (ViT-S/16) & 6.08 & 9.52 & 6.76 \\
		TUFA (ViT-S/16)$^\star$ & 5.21 &  8.25 & 5.80 \\
		TUFA (ViT-S/8) & 7.12 &  9.96 & 7.67 \\
		TUFA (ViT-S/8)$^\star$ & {\color{blue} $\bm{4.98}$} &  {\color{red} $\bm{8.08}$} & {\color{blue} $\bm{5.58}$} \\ \hline
	\end{tabular}
	\label{Table7}
\end{table}

\begin{table*}[t!]
	\centering
	\caption{Performance comparison of \textit{cross}-dataset validation on 300W-private. The NME, FR$_{0.08}$ and AUC$_{0.08}$ under inter-ocular normalization are reported. Key: [{\color{red} \textbf{Best}}, {\color{blue} \textbf{Second Best}}, $^\star$=\textbf{trained with multiple datasets}]}
	\resizebox{\textwidth}{!}{
		\begin{tabular}{m{4.5cm}<{\centering}|m{1.4cm}<{\centering}m{1.4cm}<{\centering}m{1.4cm}<{\centering}|m{1.4cm}<{\centering}m{1.4cm}<{\centering}m{1.4cm}<{\centering}|m{1.4cm}<{\centering}m{1.4cm}<{\centering}m{1.4cm}<{\centering}}
			\hline
			\multirow{2}{*}{Method} &  \multicolumn{3}{c|}{Indoor subset} & \multicolumn{3}{c|}{Outdoor subset} & \multicolumn{3}{c}{Full set} \\
			& NME$\downarrow$ & AUC$_{0.08}\uparrow$ &  FR$_{0.08}\downarrow$ & NME$\downarrow$ & AUC$_{0.08}\uparrow$ &  FR$_{0.08}\downarrow$ & NME$\downarrow$ & AUC$_{0.08}\uparrow$ &  FR$_{0.08}\downarrow$  \\ 
			\hline
			DAN \citep{DAN} & - & - & - & - & - & - & 4.30\% & 47.00\% & 2.67\% \\
			SHN \citep{SHN} & 4.10\% & - & - & 4.00\% & - & - & 4.05\% & - & - \\
			DCFE \citep{DCFE} & 3.96\% & 52.28\% & 2.33\% & 3.81\% & 52.56\% & 1.33\% & 3.88\% & 52.42\% & 1.83\% \\
			SPIGA \citep{SPIGA} & {\color{blue} $\bm{3.43\%}$} & {\color{blue} $\bm{57.35\%}$} & 1.00\% & {\color{blue} $\bm{3.43\%}$} & {\color{blue} $\bm{57.17\%}$} & {\color{blue} $\bm{0.33\%}$} & {\color{blue} $\bm{3.43\%}$} & {\color{blue} $\bm{57.27\%}$} & 0.67\% \\
			DSLPT \citep{DSLPT} & 3.47\% & 56.60\% & {\color{blue} $\bm{0.33\%}$} & 3.47\% & 56.68\% & {\color{red} $\bm{0.00\%}$} & 3.47\% & 56.64\% & {\color{red} $\bm{0.17\%}$}  \\ \hline
			TUFA (ViT-S/16) & 3.89\% & 51.39\% & 0.67\% & 3.92\% & 51.11\% & 1.00\% & 3.91\% & 51.25\%  & 0.83\% \\
			TUFA (ViT-S/16)$^\star$ & 3.51\% & 56.09\% & {\color{blue} $\bm{0.33\%}$} & 3.52\% & 55.96\% & 0.67\% & 3.52\% & 56.03\%  & {\color{blue} $\bm{0.50\%}$} \\
			TUFA (ViT-S/8) & 3.86\% & 51.98\% & 1.33\% & 3.86\% & 51.90\% & 1.00\% & 3.86\% & 51.94\%  & 1.17\% \\
			TUFA (ViT-S/8)$^\star$ & {\color{red} $\bm{3.40\%}$} & {\color{red} $\bm{57.48\%}$} & {\color{red} $\bm{0.00\%}$} & {\color{red} $\bm{3.42\%}$} & {\color{red} $\bm{57.31\%}$} & {\color{blue} $\bm{0.33\%}$} & {\color{red} $\bm{3.41\%}$} & {\color{red} $\bm{57.40\%}$}  & {\color{red} $\bm{0.17\%}$} \\ \hline
		\end{tabular}
	}
	\label{Table8}
\end{table*}

\begin{table*}[t!]
	\centering
	\caption{Performance comparison in inter-ocular NME (\%)$\downarrow$ with reduced training set on WFLW full set. Key: [{\color{red} \textbf{Best}}, {\color{blue} \textbf{Second Best}}]}
	\resizebox{0.8\textwidth}{!}{
		\begin{tabular}{m{4.8cm}<{\centering}|m{1.0cm}<{\centering}m{1.0cm}<{\centering}m{1.0cm}<{\centering}m{1.0cm}<{\centering}m{1.0cm}<{\centering}m{1.0cm}<{\centering}m{1.0cm}<{\centering}}
			\hline
			\multirow{2}{*}{Method} &  \multicolumn{7}{c}{Training set size} \\
			& 1 & 10 & 20 & 50 & 375 & 1500 & 7500 \\ \hline
			AV w. SAN \citep{AVS} & - & - & - & - & - & 6.00 & {\color{blue} \textbf{4.39}} \\
			Xiao et al. \citep{SimBa} & 43.0 & 21.9 & 19.3 & 17.6 & 10.6 & 7.08 & 5.62 \\
			Autolink \citep{Autolink} & 14.9 & 13.5 & 13.3 & 11.2 & 7.68 & 7.31 & 6.35 \\
			3FabRec \citep{3FabRec} & 15.8 & 9.66 & - & 8.39 & 7.68 & 6.51 & 5.62 \\
			He et al. \citep{GeoFewShot} & {\color{blue} \textbf{12.4}} & {\color{blue} \textbf{9.19}} & {\color{blue} \textbf{8.62}} & {\color{blue} \textbf{7.90}} & {\color{blue} \textbf{6.22}} & {\color{blue} \textbf{5.61}} & 5.38 \\ \hline
			TUFA & {\color{red} \textbf{7.62}} & {\color{red} \textbf{5.20}} & {\color{red} \textbf{4.90}} & {\color{red} \textbf{4.77}} & {\color{red} \textbf{4.43}} & {\color{red} \textbf{4.28}} & {\color{red} \textbf{3.99}}  \\   \hline
		\end{tabular}
	}
	\label{Table9}
\end{table*}

\textbf{300W-private}: we carry out cross-dataset validation on 300W-private using the same settings as those for COFW and Masked 300W. Similarly, the improvement brought to TUFA by multi-dataset learning is also significant. Even though most samples from 300W-private are under common conditions, which are easily handled by other multi-stage methods, TUFA still yields the best performance across all metrics on the full set. The FR$_{0.08}$ of TUFA is as low as 0.17\%, which means only one sample failed to be aligned by TUFA in the entire dataset. This demonstrates that TUFA is highly robust compared to all existing methods.

\subsection{Few-shot Face Alignment}
We pretrain TUFA (ViT-S/8) on COFW, 300W and MERL-RAV (\textbf{29 \& 68 landmarks}) and implement few-shot face alignment on WFLW (\textbf{98 landmarks}) as described in Section 3.4. Note that there are large differences between the semantic definitions of 29 \& 68 landmarks and 98 landmarks. We randomly select a certain number of training samples in each experiment. We repeat each experiment five times and report the mean NME with each training set size in Table 9. With only \textit{10} training samples, TUFA still demonstrates superior performance compared to other methods, outperforming 3FabRec and He et al. trained with \textit{7500} annotated faces. When we fine-tune TUFA on WFLW using 100\% of the original training set (7500 samples), the NME on the testing set improves to 3.99\%. This performance is comparable to the NME of the TUFA when trained with all four datasets together. This result indicates that TUFA can effectively leverage the knowledge learned from pretraining. This is achieved because TUFA successfully unifies the learning objectives of different datasets and mitigates their knowledge discrepancies, rather than treating landmark regression as an independent target. Because of this unique property, TUFA can be transferred to a group of newly defined landmarks efficiently, even with a very limited number of annotated samples. 3FabRec, Autolink and He et al. set image reconstruction as a pre-text task to encourage model to perform better with very limited training images. However, He et al. also found that this pre-text task may also lead to performance degradation when the number of training samples is sufficient. The main reason is that the learning objectives of image reconstruction and face alignment are not aligned. This large knowledge discrepancy significantly degrades performance.

\subsection{Zero-shot Face Alignment}
\textbf{CelebA-aligned}: to evaluate the semantic consistency of the landmarks predicted by TUFA in a zero-shot manner, we follow the same evaluation protocol as other unsupervised landmark detection methods \citep{Unsuper_Jakab}, \citep{Unsuper_Lorenz}, \citep{Autolink} used on CelebA-aligned \citep{CelebA}. We randomly generate a scratch shape with a certain number of points $N_{\rm pre}$. Then, the TUFA (ViT-S/8), trained with the four datasets, directly predicts the corresponding landmarks on both MAFL training and testing subset. The predicted coordinates on MAFL training subset are used to calculate a matrix that projects the coordinates to the positions of the labeled landmarks. We use this matrix to transfer the predicted landmark coordinates on MAFL testing subset to the coordinates of the labeled landmarks. Finally, we quantitatively measure the performance with inter-ocular NME.

\begin{table}[t!]
	\centering
	\caption{Performance of TUFA in inter-ocular NME ($N_{\rm pre}$=10) on CelebA under the setting of zero-shot face alignment. The evaluation results of state-of-the-art self-supervised and unsupervised methods also reported. Key: [{\color{red} \textbf{Best}}, {\color{blue} \textbf{Second Best}}]}
	\begin{tabular}{m{3.0cm}<{\centering}|m{1.2cm}<{\centering}|m{1.8cm}<{\centering}}
		\hline
		Method  & type & NME $\downarrow$   \\  \hline
		Thewlis et al. \citep{Unsuper_Thewlis} & unsuper-vised & 7.95\% \\ 
		Zhang et al. \citep{Unsuper_Zhang} & unsuper-vised & 3.46\% \\ 
		Lorenz et al. \citep{Unsuper_Lorenz} & unsuper-vised & 3.24\% \\ 
		IMM \citep{Unsuper_Jakab} & unsuper-vised & {\color{blue} \textbf{3.19\%}} \\ 
		AutoLink \citep{Autolink} & unsuper-vised & 3.92$\pm$0.69\% \\
		Mallis et al. \citep{SelfLandmark} & self-supervised & 3.83\% \\ \hline
		TUFA & zero-shot & {\color{red} \textbf{2.65$\pm$0.33\%}} \\ \hline
	\end{tabular}
	\label{Table10}
\end{table}

We conduct this experiment 10 times with 10 randomly generated scratch shapes. The mean and variance of the NMEs are reported in Table 10. Although TUFA is not trained with the 160,000 in-domain faces from CelebA training set, unlike other self-supervised or unsupervised methods, it still yields the best performance, significantly outperforming Mallis et al. and IMM, by 30.81\% and 16.93\% respectively, in terms of NME. This result demonstrates the landmarks predicted by TUFA in the zero-shot manner display better semantic consistency in various cases, as compared to other methods. Moreover, the semantic definitions of the landmarks predicted by TUFA can be assigned by humans via a manually generated shape, whereas those in self-supervised or unsupervised methods are learned randomly. Consequently, TUFA has a much broader application range.

\begin{table}[t!]
	\centering
	\caption{Performance of TUFA in inter-ocular NME (\%)$\downarrow$ on WFLW under the setting of zero-shot face alignment. The evaluation results of some fully supervised methods before 2018 are also reported. Key: [{\color{red} \textbf{Best}}, {\color{blue} \textbf{Second Best}}]}
	\begin{tabular}{m{1.7cm}<{\centering}|m{0.4cm}<{\centering}m{0.4cm}<{\centering}m{0.4cm}<{\centering}m{0.3cm}<{\centering}m{0.3cm}<{\centering}m{0.3cm}<{\centering}m{0.45cm}<{\centering}}
		\hline
		Method  & Full & Pose  & Exp. & Ill. & Mu. & Occ. & Blur  \\ \hline
		\multicolumn{8}{c}{Fully-supervised Face Alignment} \\ \hline
		SDM \citep{SDM} & 10.29 & 24.10 & 11.45 & 9.32 & 9.38 & 13.03 & 11.28 \\
		CFSS \citep{CFSS} & 9.07 & 21.36 & 10.09 & 8.30 & 8.74 & 11.76 & 9.96 \\
		DVLN \citep{DVLN} & {\color{blue} \textbf{6.08}} & {\color{blue} \textbf{11.54}} & {\color{blue} \textbf{6.78}} & {\color{red} \textbf{5.73}} & {\color{blue} \textbf{5.98}} & {\color{blue} \textbf{7.33}} & {\color{blue} \textbf{6.88}} \\ \hline
		\multicolumn{8}{c}{Zero-shot Face Alignment} \\ \hline
		TUFA & {\color{red} \textbf{5.90}} & {\color{red} \textbf{10.03}} & {\color{red} \textbf{6.16}} & {\color{blue} \textbf{5.78}} & {\color{red} \textbf{5.50}} & {\color{red} \textbf{6.68}} & {\color{red} \textbf{6.69}} \\ \hline
	\end{tabular}
	\label{Table11}
\end{table}

\begin{table*}[t!]
	\caption{The influence of different numbers of datasets used in training. The inter-ocular NME on 300W, COFW and WFLW, and the NME$_{\rm box}$ on MERL-RAV are reported. Key: [{\color{red} \textbf{Best}}, {\color{blue} \textbf{Second Best}}]}
	\resizebox{\textwidth}{!}{
		\begin{tabular}{m{3.0cm}<{\centering}|m{1.5cm}<{\centering}m{1.5cm}<{\centering}m{1.5cm}<{\centering}m{1.5cm}<{\centering}|m{1.5cm}<{\centering}m{1.5cm}<{\centering}m{1.5cm}<{\centering}m{1.5cm}<{\centering}}
			\hline
			\multirow{2}{*}{Method} & \multicolumn{4}{c|}{Training Datasets} & \multicolumn{4}{c}{NME (\%)$\downarrow$} \\
			& 300W & COFW & WFLW & MERL & 300W & COFW & WFLW & MERL \\ \hline
			TUFA 1 (ViT-S/8) & \checkmark & - & - & - & 3.23 & - & - & - \\
			TUFA 2 (ViT-S/8) & \checkmark & \checkmark & - & - & 3.14 & 3.27 & - & - \\
			TUFA 3 (ViT-S/8) & \checkmark & \checkmark & \checkmark & - & {\color{blue} \textbf{3.08}} & {\color{blue} \textbf{3.18}} & {\color{blue} \textbf{4.17}} & - \\
			TUFA 4 (ViT-S/8) & \checkmark & \checkmark & \checkmark & \checkmark & {\color{red} \textbf{2.95}} & {\color{red} \textbf{3.07}} & {\color{red} \textbf{3.93}} & {\color{red} \textbf{1.54}} \\ \hline
		\end{tabular}
	}
	\label{Table12}
\end{table*}

\begin{table*}[t!]
	\caption{The influence of semantic alignment embeddings. The inter-ocular NME on 300W, COFW, WFLW, COFW68, Masked 300W and 300W-private, and the NME$_{\rm box}$ on MERL-RAV are reported. Key: [{\color{red} \textbf{Best}}, 300W-M=masked 300W, 300W-P=300W-private]}
	\resizebox{\textwidth}{!}{
		\begin{tabular}{m{3.0cm}<{\centering}|m{3.2cm}<{\centering}|m{1.5cm}<{\centering}m{1.5cm}<{\centering}m{1.5cm}<{\centering}m{1.5cm}<{\centering}m{1.5cm}<{\centering}m{1.5cm}<{\centering}m{1.5cm}<{\centering}}
			\hline
			\multirow{2}{*}{Method} & Semantic Alignment & \multicolumn{7}{c}{NME (\%)$\downarrow$} \\
			&  Embeddings & 300W & COFW & WFLW & MERL & COFW68 & 300W-M & 300W-P \\ \hline
			TUFA1 (ViT-S/8) & w/o & 3.42 & 3.08 & 3.97 & 1.57 & 3.71 & 5.87 & 4.08  \\
			TUFA2 (ViT-S/8) & w & {\color{red} \textbf{2.95}} & {\color{red} \textbf{3.07}} & {\color{red} \textbf{3.93}} & {\color{red} \textbf{1.54}} & {\color{red} \textbf{3.70}} & {\color{red} \textbf{5.58}} & {\color{red} \textbf{3.40}}  \\ \hline
			
		\end{tabular}
	}
	\label{Table13}
\end{table*}

\begin{table*}[t!]
	\caption{Performance comparison to multi-task learning. The inter-ocular NME on 300W, COFW, WFLW, COFW68, Masked 300W and 300W-private, and the NME$_{\rm box}$ on MERL-RAV are reported. Key: [{\color{red} \textbf{Best}}, 300W-M=masked 300W, 300W-P=300W-private]}
	\resizebox{\textwidth}{!}{
		\begin{tabular}{m{5.0cm}<{\centering}|m{1.8cm}<{\centering}m{1.8cm}<{\centering}m{1.8cm}<{\centering}m{1.8cm}<{\centering}m{1.8cm}<{\centering}m{1.8cm}<{\centering}m{1.8cm}<{\centering}}
			\hline
			\multirow{2}{*}{Method} & \multicolumn{7}{c}{NME (\%)$\downarrow$} \\
			 & 300W & COFW & WFLW & MERL & COFW68 & 300W-M & 300W-P \\ \hline
			Multi-task (ViT-S/8) & 3.01 & 3.26 & 4.01 & 1.94 & 3.73 & 6.06 & 3.49 \\
			TUFA (ViT-S/8) & {\color{red} \textbf{2.95}} & {\color{red} \textbf{3.07}} & {\color{red} \textbf{3.93}} & {\color{red} \textbf{1.54}} & {\color{red} \textbf{3.70}} & {\color{red} \textbf{5.58}} & {\color{red} \textbf{3.40}}  \\ \hline
		\end{tabular}
	}
	\label{Table14}
\end{table*}

\begin{table*}[t!]
	\caption{The influence of masking ratio on WFLW. The inter-ocular NME, FR$_{0.1}$, AUC$_{0.1}$ and the computational complexity of Decoder (Flops) are reported. Additionally, the memory usage and training speed on a single A40 are also provided (with the batch size set to 16). Key: [{\color{red} \textbf{Best}}, {\color{blue} \textbf{Second Best}}]}
	\centering
	\resizebox{0.8\textwidth}{!}{
		\begin{tabular}{m{4.0cm}<{\centering}|m{1.2cm}<{\centering}m{1.2cm}<{\centering}m{1.2cm}<{\centering}m{1.2cm}<{\centering}m{1.2cm}<{\centering}m{1.2cm}<{\centering}m{1.2cm}<{\centering}}
			\hline
			\multirow{2}{*}{Metric} & \multicolumn{7}{c}{masking ration} \\
			&  0\% & 25\% & 50\% & 75\% & 80\% & 85\% & 90\% \\ \hline
			NME(\%)$\downarrow$ & 4.265 & 4.256 & {\color{blue} \textbf{4.233}} & {\color{red} \textbf{4.229}} & {\color{red} \textbf{4.229}} & 4.241 & 4.682  \\ 
			FR$_{0.1}$(\%)$\downarrow$ & 2.800 & {\color{blue} \textbf{2.360}} & {\color{blue} \textbf{2.360}} & 2.440 & {\color{red} \textbf{2.320}} & 2.440 & 3.920  \\ 
			AUC$_{0.1}$$\uparrow$ & 0.582 & 0.582 & {\color{blue} \textbf{0.584}} & {\color{red} \textbf{0.585}} & {\color{blue} \textbf{0.584}} & 0.582 & 0.546  \\ 
			Decoder Flops (G) $\downarrow$ & 3.537 & 3.089 & 2.663 & 2.226 & 2.139 & {\color{blue} \textbf{2.053}} & {\color{red} \textbf{1.967}} \\ 
			Memory usage (MB)$\downarrow$ & 14,004 & 13,792 & 13,608 & 13,502 & 13,490 & {\color{blue} \textbf{13,436}} & {\color{red} \textbf{13,368}} \\
			Training Speed (samples/sec) $\uparrow$ & 45.8 & 46.4 & 46.8 &  47.2 & {\color{blue} \textbf{47.3}} & {\color{blue} \textbf{47.3}} & {\color{red} \textbf{47.5}} \\
			\hline
			
		\end{tabular}
	}
	\label{Table15}
\end{table*}

\begin{figure}[t!]
	\centering
	\includegraphics[width=\linewidth]{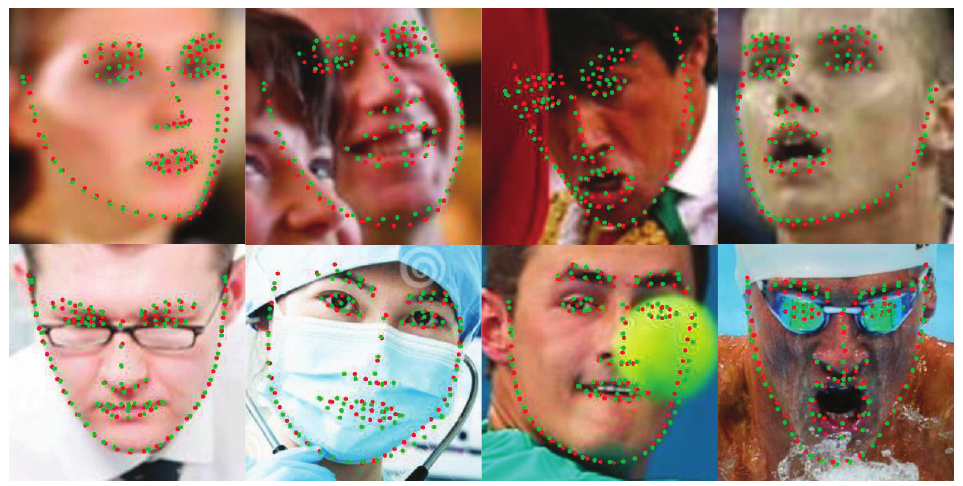}
	\caption{Zero-shot alignment results predicted by TUFA on WFLW. The \textcolor[rgb]{0, 1, 0}{\textbf{green}} points and \textcolor[rgb]{1, 0, 0}{\textbf{red}} points represent the predicted landmarks and ground truth respectively.}  
	\label{fig8}
\end{figure}

\textbf{WFLW}: to better evaluate the performance of zero-shot face alignment in practical application, we carry out an additional experiment. We first train TUFA (ViT-S/8) on MERL-RAV (\textbf{68 landmarks}). Then, we align the mean shape on WFLW (\textbf{98 landmarks}) to the mean shape on MERL-RAV using affine transformation. Finally, we evaluate TUFA on the WFLW full set directly without re-training. Since TUFA is the first framework for zero-shot face alignment, we can only compare TUFA in the zero-shot setting with some fully-supervised methods to quantitatively demonstrate its performance. The evaluation results of zero-shot face alignment with TUFA, and some fully supervised methods released before 2018, are listed in Table 11. Despite the large disparity in landmark semantics and numbers between WFLW and MERL-RAV, TUFA in the zero-shot setting still outperforms SDM, CFSS and DVLN by 42.66\%, 34.95\% and 2.96\% respectively in the metric of NME. The qualitative results shown in Fig. 8 demonstrate the semantics of the landmarks predicted by the zero-shot face alignment remain highly consistent, regardless of heavy occlusion or profile view. Therefore, TUFA can be easily applied to a group of newly defined landmarks without any annotated training samples and re-training processes, significantly broadening the application range of existing face alignment methods. Moreover, we can also conclude that TUFA learns a more universal knowledge for face alignment, which explicitly represents the regular structure of human face.

\subsection{Ablation Studies}
\textbf{Influence of the number of training datasets}: the performance of TUFA, when trained with different numbers of datasets, is tabulated in Table 12. The quantitative results on each dataset are significantly improved as the number of training datasets increases, despite the large difference between their annotation forms. This demonstrates TUFA effectively mitigates knowledge discrepancies among multiple datasets and maximizes the utilization of their knowledge. Therefore, even though COFW only provides 1,345 labeled faces, it still improves NME from 3.23\% to 3.14\% on 300W. This improvement is quite competitive in face alignment task. MERL-RAV provides the largest number of annotated faces, leading to the most significant improvement on 300W, COFW and WFLW. The unique property of TUFA also enables its performance to be further improved with the release of more high-quality face alignment datasets in the future.

\begin{figure*}[t!]
	\centering
	\includegraphics[width=\linewidth]{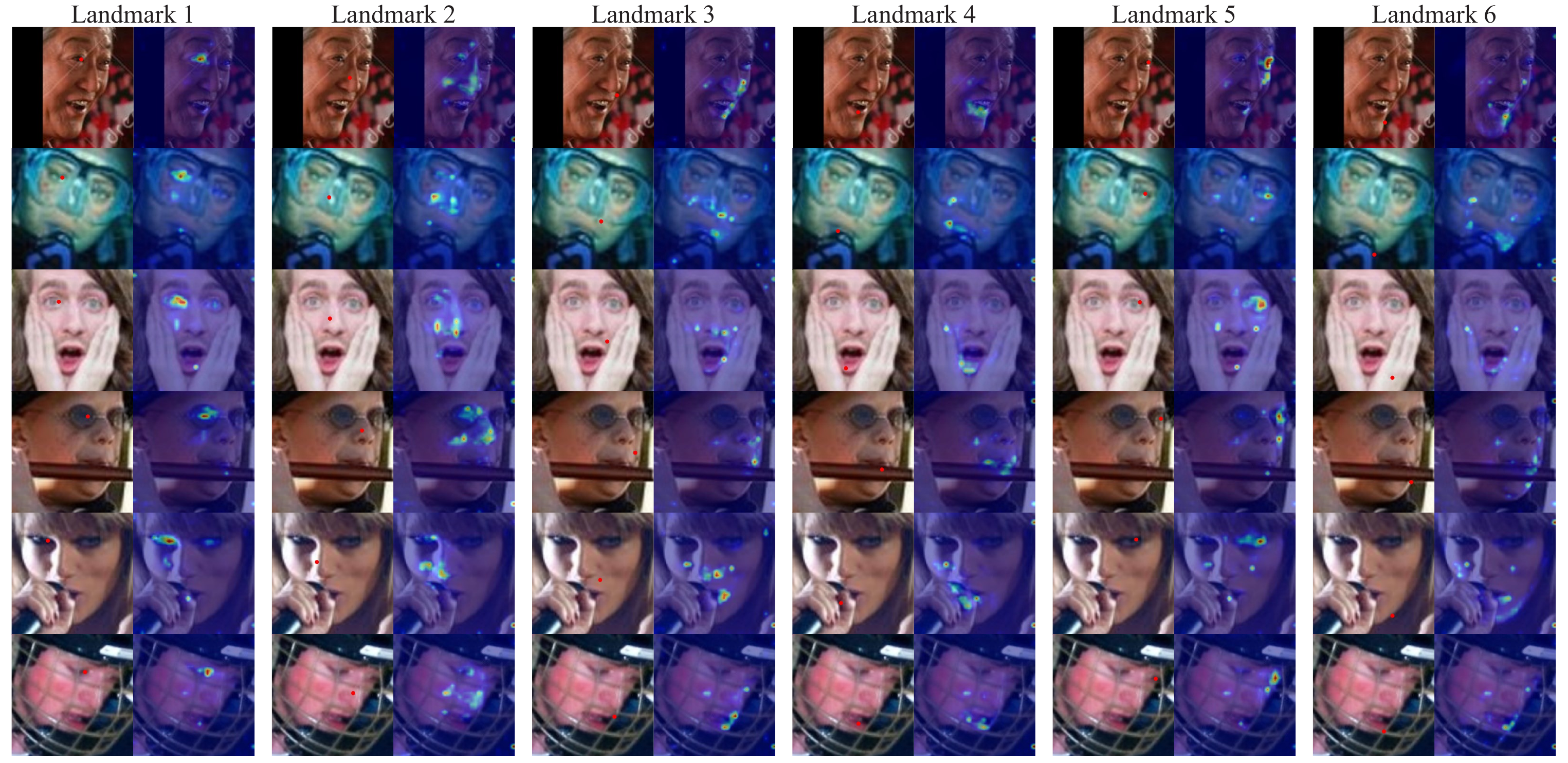}
	\caption{The zero-shot landmark predicted results of TUFA (column 1, 3, 5, 7, 9 and 11) and the corresponding mean cross attention maps (column 2, 4, 6, 8, 10 and 12). The \textcolor[rgb]{1, 0, 0}{\textbf{red}} points represent the predicted landmarks.}
	\label{fig9}
\end{figure*}

\textbf{Influence of the semantic alignment embeddings}: to further investigate the influence of semantic alignment embeddings, we also implement a version of TUFA that does not incorporate these embeddings. Instead, we align the mean shapes from different datasets using affine transformation and train TUFA using these aligned mean shapes, without adding the semantic alignment embeddings. The comparison results between the TUFA with/without these embeddings are shown in Table 13. The aligned mean shape based on affine transformation cannot explicitly mitigate the knowledge discrepancies among multiple datasets and may lead to semantic ambiguity. Consequently, it inevitably introduces a quantitative error into the predicted results. Because 300W, 300W-private and Masked 300W are labeled using semi-supervised methodology \citep{300W-A}, their annotations exhibit much smaller variance than manual annotations. Therefore, the quantitative errors result in significant performance degradation on these three datasets. With an average occlusion of over 23\%, the manually annotated landmarks on COFW and COFW68 exhibit much larger variance than other datasets. Contrary to the results on 300W, 300W-private and Masked 300W, the evaluation results on these datasets are less sensitive to the quantitative errors. Nevertheless, the absence of semantic alignment still leads to slight performance degradation. Overall, the subtle semantic differences between the landmarks with similar definitions cannot be ignored. The semantic alignment embeddings, while only slightly increasing the number of parameters in the training process, effectively eliminate the quantitative error caused by these semantic differences.

\textbf{Comparison to multi-task learning}: To further demonstrate the importance of mitigating knowledge discrepancies, we implement a model using multi-task learning under the same settings and compare it to TUFA. We define a learnable vector for each landmark to replace the structure prompt for regression. The results are presented in Table 14. Despite the similar semantics among the landmarks from 300W and MERL, they are treated as completely different regression targets in multi-task learning, leading to conflicts during training. Since MERL-RAV does not provide annotations for self-occluded landmarks, the model primarily fits 300W. Consequently, TUFA improves NME by 20.62\% on MERL-RAV, which is significantly greater compared to the improvement on other datasets. Additionally, the comparisons on other datasets illustrate that the proposed strategy can maximize the utilization of knowledge from multiple datasets and boost the generalization ability of model.

\textbf{Influence of the masking ratio}: to further explore the influence of masking ratio, we train TUFA (ViT-S/8) with different masking ratios on WFLW. The results are reported in Table 15. As the masking ratio increases, the performance of TUFA is improved. The primary reason is that randomly masking a certain ration of landmarks encourages TUFA to learn the mapping using the relationship between distant landmarks. This relationship ensures a more coherent mapping between the 2D plane and target faces, while the landmarks that are close together often introduce bias into the learned mapping. Moreover, increasing the masking ratio also significantly reduces the computational complexity of the TUFA decoder. When the ratio is set to 75\%, the computational complexity of TUFA decoder is only 2/3 of that at a 0\% masking ration. Although TUFA decoder only accounts for just about 10\% of the total computational complexity compared to TUFA encoder (31.572G flops), we can still observe a notable reduction in memory usage and an acceleration in training speed brought by the masking training strategy.

The results show TUFA works well within the range of $[50\%, 85\%]$. If the masking ratio exceeds 85\%, the number of remaining landmarks, which serves as anchors during training, is insufficient to construct a mapping from a 2D plane to a highly complex face surface. As a result, the performance of TUFA degrades significantly. After comprehensively considering both performance and computational complexity, we set the masking ratio to 0.75 in our model.

\textbf{Cross attention visualization}: we visualize the cross attention maps of a scratch shape predicted by TUFA in Fig. 9. The attention maps indicate the mean weights from the structure prompts to image patches in MCA blocks, and the definition of the scratch is shown in the second row of Fig. 5. Even though the landmarks of the scratch shape are unseen during training, they can still guide TUFA to focus on the corresponding parts of faces regardless of various poses and conditions. It reveals the reliability of the zero-shot face alignment achieved by TUFA. Moreover, if the corresponding parts on faces are occluded or lack significant features (row 3, landmark 3 and 6), TUFA can also look at nearby facial components and utilize their relative positions for landmark locating. That is why TUFA demonstrates much stronger robustness than other methods in very challenging cases.

\subsection{Limitations}

TUFA can only generalize to unseen landmarks within the face area. If TUFA uses a set of structural prompts generated from points outside the face area on the interpretable plane, it still localize a set of nearest landmarks to these points inside the face area. The main reason is that the training datasets only contain information about facial structure. By using a dataset for dense human pose estimation \citep{DensePose}, it becomes possible to locate unseen landmarks on human bodies in a zero-shot manner.

\section{Conclusion}\label{sec5}

In this paper, we introduce a novel strategy to mitigate knowledge discrepancies among multiple datasets and further achieve task-agnostic, unified face alignment. Compared to existing face alignment methods, the proposed method represents a significant breakthrough in three critical aspects: 1) TUFA successfully mitigates knowledge discrepancies among multiple datasets by aligning their landmark semantics on an interpretable plane and further unifying their learning target to map the plane to target faces. 2) the unified learning target also enables the learned knowledge to be easily transferred to a new set of defined landmarks. Consequently, TUFA significantly boosts the performance of few-shot face alignment. 3) By mitigating the discrepancies, the learned knowledge is extended from target landmarks to face structure. As a result, we can employ structure prompts to further bridge the gap between seen and unseen landmarks, achieving zero-shot face alignment for the first time. 

Although the gap between seen and unseen landmarks has been successfully addressed, the gap between different domains, such as varying lighting conditions, has not been explicitly considered in TUFA. In the future, we will explore the domain gap among facial landmarks to develop a more universal face alignment approach.

%Overall, we believe the unique properties of TUFA can propel face alignment to a new stage and provide a competitive baseline for future works.

\section*{Declarations}

\noindent \textbf{Data Availability Statement}: the datasets used in our paper are available in: 

WFLW: \url{https://wywu.github.io/projects/LAB/WFLW.html}, 

300W: \url{https://ibug.doc.ic.ac.uk/resources/300-W/#:~:text=Automatic\%20facial\%20landmark\%20detection\%20is\%20a}, 

Masked 300W: \url{https://github.com/zhuccly/saat},

MERL-RAV: \url{https://github.com/abhi1kumar/MERL-RAV_dataset},

COFW: \url{https://data.caltech.edu/records/bc0bf-nc666#:~:text=A\%20novel\%20face\%20dataset\%20focused\%20on}, 

COFW68: \url{https://github.com/golnazghiasi/cofw68-benchmark},

CelebA: \url{https://mmlab.ie.cuhk.edu.hk/projects/CelebA.html}.

\bibliography{egbib}% common bib file

\begin{thebibliography}{}
\renewcommand{\doi}[1]{\url{https://doi.org/#1}}
\bibcommenthead

\bibitem [\protect \citeauthoryear {%
Asthana%
, Marks%
, Jones%
, Tieu%
\BCBL {}\ \BBA {} Rohith%
}{%
Asthana%
\ \protect \BOthers {.}}{%
{\protect \APACyear {2011}}%
}]{%
AAM_E1}
\APACinsertmetastar {%
AAM_E1}%
\begin{APACrefauthors}%
Asthana, A.%
, Marks, T.K.%
, Jones, M.J.%
, Tieu, K.H.%
\BCBL {} Rohith, M.%
\end{APACrefauthors}%
\unskip\
\newblock
\APACrefYearMonthDay{2011}{}{}.
\newblock
{\BBOQ}\APACrefatitle {Fully automatic pose-invariant face recognition via 3D
  pose normalization} {Fully automatic pose-invariant face recognition via 3d
  pose normalization}.{\BBCQ}
\newblock
 \APACrefbtitle {Proc. IEEE Int. Conf. Comput. Vis.} {Proc. ieee int. conf.
  comput. vis.}\ (\BPG~937-944).
\PrintBackRefs{\CurrentBib}

\bibitem [\protect \citeauthoryear {%
Ba%
, Kiros%
\BCBL {}\ \BBA {} Hinton%
}{%
Ba%
\ \protect \BOthers {.}}{%
{\protect \APACyear {2016}}%
}]{%
LN}
\APACinsertmetastar {%
LN}%
\begin{APACrefauthors}%
Ba, J.L.%
, Kiros, J.R.%
\BCBL {} Hinton, G.E.%
\end{APACrefauthors}%
\unskip\
\newblock
\APACrefYearMonthDay{2016}{}{}.
\newblock
{\BBOQ}\APACrefatitle {Layer normalization} {Layer normalization}.{\BBCQ}
\newblock
\APACjournalVolNumPages{arXiv:1607.06450}{}{}{,}
\newblock

\newblock

\PrintBackRefs{\CurrentBib}

\bibitem [\protect \citeauthoryear {%
Belhumeur%
, Jacobs%
, Kriegman%
\BCBL {}\ \BBA {} Kumar%
}{%
Belhumeur%
\ \protect \BOthers {.}}{%
{\protect \APACyear {2011}}%
}]{%
LFPW}
\APACinsertmetastar {%
LFPW}%
\begin{APACrefauthors}%
Belhumeur, P.N.%
, Jacobs, D.W.%
, Kriegman, D.J.%
\BCBL {} Kumar, N.%
\end{APACrefauthors}%
\unskip\
\newblock
\APACrefYearMonthDay{2011}{}{}.
\newblock
{\BBOQ}\APACrefatitle {Localizing parts of faces using a consensus of
  exemplars} {Localizing parts of faces using a consensus of exemplars}.{\BBCQ}
\newblock
 \APACrefbtitle {Proc. IEEE Conf. Comput. Vis. Pattern Recognit.} {Proc. ieee
  conf. comput. vis. pattern recognit.}\ (\BPG~545-552).
\PrintBackRefs{\CurrentBib}

\bibitem [\protect \citeauthoryear {%
Browatzki%
\ \BBA {} Wallraven%
}{%
Browatzki%
\ \BBA {} Wallraven%
}{%
{\protect \APACyear {2020}}%
}]{%
3FabRec}
\APACinsertmetastar {%
3FabRec}%
\begin{APACrefauthors}%
Browatzki, B.%
\BCBT {}\ \BBA {} Wallraven, C.%
\end{APACrefauthors}%
\unskip\
\newblock
\APACrefYearMonthDay{2020}{}{}.
\newblock
{\BBOQ}\APACrefatitle {3FabRec: Fast Few-Shot Face Alignment by Reconstruction}
  {3fabrec: Fast few-shot face alignment by reconstruction}.{\BBCQ}
\newblock
 \APACrefbtitle {Proc. IEEE Conf. Comput. Vis. Pattern Recognit.} {Proc. ieee
  conf. comput. vis. pattern recognit.}\ (\BPG~6109-6119).
\PrintBackRefs{\CurrentBib}

\bibitem [\protect \citeauthoryear {%
Bulat%
\ \BBA {} Tzimiropoulos%
}{%
Bulat%
\ \BBA {} Tzimiropoulos%
}{%
{\protect \APACyear {2017}}%
}]{%
3DFAN}
\APACinsertmetastar {%
3DFAN}%
\begin{APACrefauthors}%
Bulat, A.%
\BCBT {}\ \BBA {} Tzimiropoulos, G.%
\end{APACrefauthors}%
\unskip\
\newblock
\APACrefYearMonthDay{2017}{}{}.
\newblock
{\BBOQ}\APACrefatitle {How Far are We from Solving the 2D \& 3D Face Alignment
  Problem? (and a Dataset of 230,000 3D Facial Landmarks)} {How far are we from
  solving the 2d \& 3d face alignment problem? (and a dataset of 230,000 3d
  facial landmarks)}.{\BBCQ}
\newblock
 \APACrefbtitle {Proc. IEEE Int. Conf. Comput. Vis.} {Proc. ieee int. conf.
  comput. vis.}\ (\BPG~1021-1030).
\PrintBackRefs{\CurrentBib}

\bibitem [\protect \citeauthoryear {%
Burgos-Artizzu%
, Perona%
\BCBL {}\ \BBA {} Dollár%
}{%
Burgos-Artizzu%
\ \protect \BOthers {.}}{%
{\protect \APACyear {2013}}%
}]{%
COFW}
\APACinsertmetastar {%
COFW}%
\begin{APACrefauthors}%
Burgos-Artizzu, X.P.%
, Perona, P.%
\BCBL {} Dollár, P.%
\end{APACrefauthors}%
\unskip\
\newblock
\APACrefYearMonthDay{2013}{}{}.
\newblock
{\BBOQ}\APACrefatitle {Robust Face Landmark Estimation under Occlusion} {Robust
  face landmark estimation under occlusion}.{\BBCQ}
\newblock
 \APACrefbtitle {Proc. IEEE Int. Conf. Comput. Vis.} {Proc. ieee int. conf.
  comput. vis.}\ (\BPG~1513-1520).
\PrintBackRefs{\CurrentBib}

\bibitem [\protect \citeauthoryear {%
Caron%
\ \protect \BOthers {.}}{%
Caron%
\ \protect \BOthers {.}}{%
{\protect \APACyear {2021}}%
}]{%
DINO}
\APACinsertmetastar {%
DINO}%
\begin{APACrefauthors}%
Caron, M.%
, Touvron, H.%
, Misra, I.%
, Jegou, H.%
, Mairal, J.%
, Bojanowski, P.%
\BCBL {} Joulin, A.%
\end{APACrefauthors}%
\unskip\
\newblock
\APACrefYearMonthDay{2021}{}{}.
\newblock
{\BBOQ}\APACrefatitle {Emerging Properties in Self-Supervised Vision
  Transformers} {Emerging properties in self-supervised vision
  transformers}.{\BBCQ}
\newblock
 \APACrefbtitle {Proc. IEEE Int. Conf. Comput. Vis.} {Proc. ieee int. conf.
  comput. vis.}\ (\BPG~9630-9640).
\PrintBackRefs{\CurrentBib}

\bibitem [\protect \citeauthoryear {%
Chen%
, Su%
\BCBL {}\ \BBA {} Ji%
}{%
Chen%
\ \protect \BOthers {.}}{%
{\protect \APACyear {2019}}%
}]{%
KDN}
\APACinsertmetastar {%
KDN}%
\begin{APACrefauthors}%
Chen, L.%
, Su, H.%
\BCBL {} Ji, Q.%
\end{APACrefauthors}%
\unskip\
\newblock
\APACrefYearMonthDay{2019}{}{}.
\newblock
{\BBOQ}\APACrefatitle {Face Alignment With Kernel Density Deep Neural Network}
  {Face alignment with kernel density deep neural network}.{\BBCQ}
\newblock
 \APACrefbtitle {Proc. IEEE Int. Conf. Comput. Vis.} {Proc. ieee int. conf.
  comput. vis.}\ (\BPG~6991-7001).
\PrintBackRefs{\CurrentBib}

\bibitem [\protect \citeauthoryear {%
Cootes%
, Edwards%
\BCBL {}\ \BBA {} Taylor%
}{%
Cootes%
\ \protect \BOthers {.}}{%
{\protect \APACyear {2001}}%
}]{%
AAM}
\APACinsertmetastar {%
AAM}%
\begin{APACrefauthors}%
Cootes, T.%
, Edwards, G.%
\BCBL {} Taylor, C.%
\end{APACrefauthors}%
\unskip\
\newblock
\APACrefYearMonthDay{2001}{}{}.
\newblock
{\BBOQ}\APACrefatitle {Active appearance models} {Active appearance
  models}.{\BBCQ}
\newblock
\APACjournalVolNumPages{IEEE Trans. Pattern Anal. Mach.
  Intell.}{23}{6}{681-685,}
\newblock
\begin{APACrefDOI} \doi{10.1109/34.927467} \end{APACrefDOI}
\newblock

\newblock

\PrintBackRefs{\CurrentBib}

\bibitem [\protect \citeauthoryear {%
Cootes%
, Taylor%
, Cooper%
\BCBL {}\ \BBA {} Graham%
}{%
Cootes%
\ \protect \BOthers {.}}{%
{\protect \APACyear {1995}}%
}]{%
ASM}
\APACinsertmetastar {%
ASM}%
\begin{APACrefauthors}%
Cootes, T.%
, Taylor, C.%
, Cooper, D.%
\BCBL {} Graham, J.%
\end{APACrefauthors}%
\unskip\
\newblock
\APACrefYearMonthDay{1995}{}{}.
\newblock
{\BBOQ}\APACrefatitle {Active Shape Models-Their Training and Application}
  {Active shape models-their training and application}.{\BBCQ}
\newblock
\APACjournalVolNumPages{Comput. Vis. Image Understand.}{61}{1}{38-59,}
\newblock
\begin{APACrefDOI} \doi{https://doi.org/10.1006/cviu.1995.1004}
  \end{APACrefDOI}
\newblock

\newblock

\PrintBackRefs{\CurrentBib}

\bibitem [\protect \citeauthoryear {%
Cristinacce%
\ \BBA {} Cootes%
}{%
Cristinacce%
\ \BBA {} Cootes%
}{%
{\protect \APACyear {2006}}%
}]{%
CLM}
\APACinsertmetastar {%
CLM}%
\begin{APACrefauthors}%
Cristinacce, D.%
\BCBT {}\ \BBA {} Cootes, T.F.%
\end{APACrefauthors}%
\unskip\
\newblock
\APACrefYearMonthDay{2006}{}{}.
\newblock
{\BBOQ}\APACrefatitle {Feature Detection and Tracking with Constrained Local
  Models} {Feature detection and tracking with constrained local
  models}.{\BBCQ}
\newblock
 \APACrefbtitle {Proc. British Mach. Vis. Conf.} {Proc. british mach. vis.
  conf.}\ (\BPG~95.1-95.10).
\newblock
\APACrefnote{doi:10.5244/C.20.95}
\PrintBackRefs{\CurrentBib}

\bibitem [\protect \citeauthoryear {%
Deng%
\ \protect \BOthers {.}}{%
Deng%
\ \protect \BOthers {.}}{%
{\protect \APACyear {2009}}%
}]{%
ImageNet}
\APACinsertmetastar {%
ImageNet}%
\begin{APACrefauthors}%
Deng, J.%
, Dong, W.%
, Socher, R.%
, Li, L\BHBI J.%
, Li, K.%
\BCBL {} Fei-Fei, L.%
\end{APACrefauthors}%
\unskip\
\newblock
\APACrefYearMonthDay{2009}{}{}.
\newblock
{\BBOQ}\APACrefatitle {ImageNet: A large-scale hierarchical image database}
  {Imagenet: A large-scale hierarchical image database}.{\BBCQ}
\newblock
 \APACrefbtitle {Proc. IEEE Conf. Comput. Vis. Pattern Recognit.} {Proc. ieee
  conf. comput. vis. pattern recognit.}\ (\BPG~248-255).
\PrintBackRefs{\CurrentBib}

\bibitem [\protect \citeauthoryear {%
Dosovitskiy%
\ \protect \BOthers {.}}{%
Dosovitskiy%
\ \protect \BOthers {.}}{%
{\protect \APACyear {2021}}%
}]{%
VIT}
\APACinsertmetastar {%
VIT}%
\begin{APACrefauthors}%
Dosovitskiy, A.%
, Beyer, L.%
, Kolesnikov, A.%
, Weissenborn, D.%
, Zhai, X.%
, Unterthiner, T.%
\BDBL {}Houlsby, N.%
\end{APACrefauthors}%
\unskip\
\newblock
\APACrefYearMonthDay{2021}{}{}.
\newblock
{\BBOQ}\APACrefatitle {An Image is Worth 16x16 Words: Transformers for Image
  Recognition at Scale} {An image is worth 16x16 words: Transformers for image
  recognition at scale}.{\BBCQ}
\newblock
 \APACrefbtitle {Proc. Int. Conf. Learn. Representations.} {Proc. int. conf.
  learn. representations.}
\PrintBackRefs{\CurrentBib}

\bibitem [\protect \citeauthoryear {%
Z.~Feng%
, Kittler%
, Christmas%
, Huber%
\BCBL {}\ \BBA {} Wu%
}{%
Z.~Feng%
\ \protect \BOthers {.}}{%
{\protect \APACyear {2017}}%
}]{%
DAC-CSR}
\APACinsertmetastar {%
DAC-CSR}%
\begin{APACrefauthors}%
Feng, Z.%
, Kittler, J.%
, Christmas, W.%
, Huber, P.%
\BCBL {} Wu, X.%
\end{APACrefauthors}%
\unskip\
\newblock
\APACrefYearMonthDay{2017}{}{}.
\newblock
{\BBOQ}\APACrefatitle {Dynamic Attention-Controlled Cascaded Shape Regression
  Exploiting Training Data Augmentation and Fuzzy-Set Sample Weighting}
  {Dynamic attention-controlled cascaded shape regression exploiting training
  data augmentation and fuzzy-set sample weighting}.{\BBCQ}
\newblock
 \APACrefbtitle {Proc. IEEE Conf. Comput. Vis. Pattern Recognit.} {Proc. ieee
  conf. comput. vis. pattern recognit.}\ (\BPG~3681-3690).
\PrintBackRefs{\CurrentBib}

\bibitem [\protect \citeauthoryear {%
Z\BHBI H.~Feng%
, Kittler%
, Christmas%
, Huber%
\BCBL {}\ \BBA {} Wu%
}{%
Z\BHBI H.~Feng%
\ \protect \BOthers {.}}{%
{\protect \APACyear {2017}}%
}]{%
DAC}
\APACinsertmetastar {%
DAC}%
\begin{APACrefauthors}%
Feng, Z\BHBI H.%
, Kittler, J.%
, Christmas, W.%
, Huber, P.%
\BCBL {} Wu, X\BHBI J.%
\end{APACrefauthors}%
\unskip\
\newblock
\APACrefYearMonthDay{2017}{}{}.
\newblock
{\BBOQ}\APACrefatitle {Dynamic Attention-Controlled Cascaded Shape Regression
  Exploiting Training Data Augmentation and Fuzzy-Set Sample Weighting}
  {Dynamic attention-controlled cascaded shape regression exploiting training
  data augmentation and fuzzy-set sample weighting}.{\BBCQ}
\newblock
 \APACrefbtitle {Proc. IEEE Conf. Comput. Vis. Pattern Recognit.} {Proc. ieee
  conf. comput. vis. pattern recognit.}\ (\BPG~3681-3690).
\PrintBackRefs{\CurrentBib}

\bibitem [\protect \citeauthoryear {%
Ghiasi%
\ \BBA {} Fowlkes%
}{%
Ghiasi%
\ \BBA {} Fowlkes%
}{%
{\protect \APACyear {2014}}%
}]{%
COFW68}
\APACinsertmetastar {%
COFW68}%
\begin{APACrefauthors}%
Ghiasi, G.%
\BCBT {}\ \BBA {} Fowlkes, C.C.%
\end{APACrefauthors}%
\unskip\
\newblock
\APACrefYearMonthDay{2014}{}{}.
\newblock
{\BBOQ}\APACrefatitle {Occlusion Coherence: Localizing Occluded Faces with a
  Hierarchical Deformable Part Model} {Occlusion coherence: Localizing occluded
  faces with a hierarchical deformable part model}.{\BBCQ}
\newblock
 \APACrefbtitle {Proc. IEEE Conf. Comput. Vis. Pattern Recognit.} {Proc. ieee
  conf. comput. vis. pattern recognit.}\ (\BPG~1899-1906).
\PrintBackRefs{\CurrentBib}

\bibitem [\protect \citeauthoryear {%
Guler%
, Neverova%
\BCBL {}\ \BBA {} Kokkinos%
}{%
Guler%
\ \protect \BOthers {.}}{%
{\protect \APACyear {2018}}%
}]{%
DensePose}
\APACinsertmetastar {%
DensePose}%
\begin{APACrefauthors}%
Guler, R.A.%
, Neverova, N.%
\BCBL {} Kokkinos, I.%
\end{APACrefauthors}%
\unskip\
\newblock
\APACrefYearMonthDay{2018}{}{}.
\newblock
{\BBOQ}\APACrefatitle {{ DensePose: Dense Human Pose Estimation in the Wild }}
  {{ DensePose: Dense Human Pose Estimation in the Wild }}.{\BBCQ}
\newblock
 \APACrefbtitle {Proc. IEEE Conf. Comput. Vis. Pattern Recognit.} {Proc. ieee
  conf. comput. vis. pattern recognit.}\ (\BPG~7297-7306).
\PrintBackRefs{\CurrentBib}

\bibitem [\protect \citeauthoryear {%
He%
, Bharaj%
, Ferman%
, Rhodin%
\BCBL {}\ \BBA {} Garrido%
}{%
He%
\ \protect \BOthers {.}}{%
{\protect \APACyear {2023}}%
}]{%
GeoFewShot}
\APACinsertmetastar {%
GeoFewShot}%
\begin{APACrefauthors}%
He, X.%
, Bharaj, G.%
, Ferman, D.%
, Rhodin, H.%
\BCBL {} Garrido, P.%
\end{APACrefauthors}%
\unskip\
\newblock
\APACrefYearMonthDay{2023}{June}{}.
\newblock
{\BBOQ}\APACrefatitle {Few-Shot Geometry-Aware Keypoint Localization} {Few-shot
  geometry-aware keypoint localization}.{\BBCQ}
\newblock
 \APACrefbtitle {Proc. IEEE Conf. Comput. Vis. Pattern Recognit.} {Proc. ieee
  conf. comput. vis. pattern recognit.}\ (\BPG~21337-21348).
\PrintBackRefs{\CurrentBib}

\bibitem [\protect \citeauthoryear {%
He%
, Wandt%
\BCBL {}\ \BBA {} Rhodin%
}{%
He%
\ \protect \BOthers {.}}{%
{\protect \APACyear {2022}}%
}]{%
Autolink}
\APACinsertmetastar {%
Autolink}%
\begin{APACrefauthors}%
He, X.%
, Wandt, B.%
\BCBL {} Rhodin, H.%
\end{APACrefauthors}%
\unskip\
\newblock
\APACrefYearMonthDay{2022}{}{}.
\newblock
{\BBOQ}\APACrefatitle {AutoLink: Self-supervised Learning of Human Skeletons
  and Object Outlines by Linking Keypoints} {Autolink: Self-supervised learning
  of human skeletons and object outlines by linking keypoints}.{\BBCQ}
\newblock
 S.~Koyejo, S.~Mohamed, A.~Agarwal, D.~Belgrave, K.~Cho\BCBL {}\ \BBA {} A.~Oh\
  (\BEDS), \APACrefbtitle {Proc. Advances Neural Inf. Process. Syst.} {Proc.
  advances neural inf. process. syst.}\ (\BVOL~35, \BPGS\ 36123--36141).
\newblock
\APACaddressPublisher{}{Curran Associates, Inc.}
\PrintBackRefs{\CurrentBib}

\bibitem [\protect \citeauthoryear {%
G.~Huang%
, Laradji%
, Vázquez%
, Lacoste-Julien%
\BCBL {}\ \BBA {} Rodríguez%
}{%
G.~Huang%
\ \protect \BOthers {.}}{%
{\protect \APACyear {2023}}%
}]{%
FewshotSurvey}
\APACinsertmetastar {%
FewshotSurvey}%
\begin{APACrefauthors}%
Huang, G.%
, Laradji, I.%
, Vázquez, D.%
, Lacoste-Julien, S.%
\BCBL {} Rodríguez, P.%
\end{APACrefauthors}%
\unskip\
\newblock
\APACrefYearMonthDay{2023}{}{}.
\newblock
{\BBOQ}\APACrefatitle {A Survey of Self-Supervised and Few-Shot Object
  Detection} {A survey of self-supervised and few-shot object
  detection}.{\BBCQ}
\newblock
\APACjournalVolNumPages{IEEE Trans. Pattern Anal. Mach.
  Intell.}{45}{4}{4071-4089,}
\newblock
\begin{APACrefDOI} \doi{10.1109/TPAMI.2022.3199617} \end{APACrefDOI}
\newblock

\newblock

\PrintBackRefs{\CurrentBib}

\bibitem [\protect \citeauthoryear {%
Y.~Huang%
, Yang%
, Li%
, Kim%
\BCBL {}\ \BBA {} Wei%
}{%
Y.~Huang%
\ \protect \BOthers {.}}{%
{\protect \APACyear {2021}}%
}]{%
ADNet}
\APACinsertmetastar {%
ADNet}%
\begin{APACrefauthors}%
Huang, Y.%
, Yang, H.%
, Li, C.%
, Kim, J.%
\BCBL {} Wei, F.%
\end{APACrefauthors}%
\unskip\
\newblock
\APACrefYearMonthDay{2021}{}{}.
\newblock
{\BBOQ}\APACrefatitle {ADNet: Leveraging Error-Bias Towards Normal Direction in
  Face Alignment} {Adnet: Leveraging error-bias towards normal direction in
  face alignment}.{\BBCQ}
\newblock
 \APACrefbtitle {Proc. IEEE Int. Conf. Comput. Vis.} {Proc. ieee int. conf.
  comput. vis.}\ (\BPG~3060-3070).
\PrintBackRefs{\CurrentBib}

\bibitem [\protect \citeauthoryear {%
Jakab%
, Gupta%
, Bilen%
\BCBL {}\ \BBA {} Vedaldi%
}{%
Jakab%
\ \protect \BOthers {.}}{%
{\protect \APACyear {2018}}%
}]{%
Unsuper_Jakab}
\APACinsertmetastar {%
Unsuper_Jakab}%
\begin{APACrefauthors}%
Jakab, T.%
, Gupta, A.%
, Bilen, H.%
\BCBL {} Vedaldi, A.%
\end{APACrefauthors}%
\unskip\
\newblock
\APACrefYearMonthDay{2018}{}{}.
\newblock
{\BBOQ}\APACrefatitle {Unsupervised Learning of Object Landmarks through
  Conditional Image Generation} {Unsupervised learning of object landmarks
  through conditional image generation}.{\BBCQ}
\newblock
 S.~Bengio, H.~Wallach, H.~Larochelle, K.~Grauman, N.~Cesa-Bianchi\BCBL {}\
  \BBA {} R.~Garnett\ (\BEDS), \APACrefbtitle {Proc. Advances Neural Inf.
  Process. Syst.} {Proc. advances neural inf. process. syst.}\ (\BVOL~31).
\newblock
\APACaddressPublisher{}{Curran Associates, Inc.}
\PrintBackRefs{\CurrentBib}

\bibitem [\protect \citeauthoryear {%
Jiang%
\ \protect \BOthers {.}}{%
Jiang%
\ \protect \BOthers {.}}{%
{\protect \APACyear {2023}}%
}]{%
FewShotRec}
\APACinsertmetastar {%
FewShotRec}%
\begin{APACrefauthors}%
Jiang, S.%
, Zhu, Y.%
, Liu, C.%
, Song, X.%
, Li, X.%
\BCBL {} Min, W.%
\end{APACrefauthors}%
\unskip\
\newblock
\APACrefYearMonthDay{2023}{}{}.
\newblock
{\BBOQ}\APACrefatitle {Dataset Bias in Few-Shot Image Recognition} {Dataset
  bias in few-shot image recognition}.{\BBCQ}
\newblock
\APACjournalVolNumPages{IEEE Trans. Pattern Anal. Mach.
  Intell.}{45}{1}{229-246,}
\newblock
\begin{APACrefDOI} \doi{10.1109/TPAMI.2022.3153611} \end{APACrefDOI}
\newblock

\newblock

\PrintBackRefs{\CurrentBib}

\bibitem [\protect \citeauthoryear {%
Jin%
, Liao%
\BCBL {}\ \BBA {} Shao%
}{%
Jin%
\ \protect \BOthers {.}}{%
{\protect \APACyear {2021}}%
}]{%
PIP}
\APACinsertmetastar {%
PIP}%
\begin{APACrefauthors}%
Jin, H.%
, Liao, S.%
\BCBL {} Shao, L.%
\end{APACrefauthors}%
\unskip\
\newblock
\APACrefYearMonthDay{2021}{Sep}{}.
\newblock
{\BBOQ}\APACrefatitle {Pixel-in-Pixel Net: Towards Efficient Facial Landmark
  Detection in the Wild} {Pixel-in-pixel net: Towards efficient facial landmark
  detection in the wild}.{\BBCQ}
\newblock
\APACjournalVolNumPages{Int. J. Comput. Vis.}{}{}{,}
\newblock
\begin{APACrefDOI} \doi{10.1007/s11263-021-01521-4} \end{APACrefDOI}
\newblock

\newblock

\PrintBackRefs{\CurrentBib}

\bibitem [\protect \citeauthoryear {%
Kowalski%
, Naruniec%
\BCBL {}\ \BBA {} Trzcinski%
}{%
Kowalski%
\ \protect \BOthers {.}}{%
{\protect \APACyear {2017}}%
}]{%
DAN}
\APACinsertmetastar {%
DAN}%
\begin{APACrefauthors}%
Kowalski, M.%
, Naruniec, J.%
\BCBL {} Trzcinski, T.%
\end{APACrefauthors}%
\unskip\
\newblock
\APACrefYearMonthDay{2017}{}{}.
\newblock
{\BBOQ}\APACrefatitle {Deep Alignment Network: A Convolutional Neural Network
  for Robust Face Alignment} {Deep alignment network: A convolutional neural
  network for robust face alignment}.{\BBCQ}
\newblock
 \APACrefbtitle {Proc. IEEE Conf. Comput. Vis. Pattern Recognit. Workshops}
  {Proc. ieee conf. comput. vis. pattern recognit. workshops}\
  (\BPG~2034-2043).
\PrintBackRefs{\CurrentBib}

\bibitem [\protect \citeauthoryear {%
Kumar%
\ \protect \BOthers {.}}{%
Kumar%
\ \protect \BOthers {.}}{%
{\protect \APACyear {2020}}%
}]{%
LUVLI}
\APACinsertmetastar {%
LUVLI}%
\begin{APACrefauthors}%
Kumar, A.%
, Marks, T.K.%
, Mou, W.%
, Wang, Y.%
, Jones, M.%
, Cherian, A.%
\BDBL {}Feng, C.%
\end{APACrefauthors}%
\unskip\
\newblock
\APACrefYearMonthDay{2020}{}{}.
\newblock
{\BBOQ}\APACrefatitle {LUVLi Face Alignment: Estimating Landmarks’ Location,
  Uncertainty, and Visibility Likelihood} {Luvli face alignment: Estimating
  landmarks’ location, uncertainty, and visibility likelihood}.{\BBCQ}
\newblock
 \APACrefbtitle {Proc. IEEE Conf. Comput. Vis. Pattern Recognit.} {Proc. ieee
  conf. comput. vis. pattern recognit.}\ (\BPG~8233-8243).
\PrintBackRefs{\CurrentBib}

\bibitem [\protect \citeauthoryear {%
Köstinger%
, Wohlhart%
, Roth%
\BCBL {}\ \BBA {} Bischof%
}{%
Köstinger%
\ \protect \BOthers {.}}{%
{\protect \APACyear {2011}}%
}]{%
AFLW}
\APACinsertmetastar {%
AFLW}%
\begin{APACrefauthors}%
Köstinger, M.%
, Wohlhart, P.%
, Roth, P.M.%
\BCBL {} Bischof, H.%
\end{APACrefauthors}%
\unskip\
\newblock
\APACrefYearMonthDay{2011}{}{}.
\newblock
{\BBOQ}\APACrefatitle {Annotated Facial Landmarks in the Wild: A large-scale,
  real-world database for facial landmark localization} {Annotated facial
  landmarks in the wild: A large-scale, real-world database for facial landmark
  localization}.{\BBCQ}
\newblock
 \APACrefbtitle {Proc. IEEE Int. Conf. Comput. Vis. Workshops} {Proc. ieee int.
  conf. comput. vis. workshops}\ (\BPG~2144-2151).
\PrintBackRefs{\CurrentBib}

\bibitem [\protect \citeauthoryear {%
Lan%
, Hu%
\BCBL {}\ \BBA {} Cheng%
}{%
Lan%
\ \protect \BOthers {.}}{%
{\protect \APACyear {2021}}%
}]{%
HIH}
\APACinsertmetastar {%
HIH}%
\begin{APACrefauthors}%
Lan, X.%
, Hu, Q.%
\BCBL {} Cheng, J.%
\end{APACrefauthors}%
\unskip\
\newblock
\APACrefYearMonthDay{2021}{}{}.
\newblock
{\BBOQ}\APACrefatitle {Revisting Quantization Error in Face Alignment}
  {Revisting quantization error in face alignment}.{\BBCQ}
\newblock
 \APACrefbtitle {Proc. IEEE Int. Conf. Comput. Vis. Workshops} {Proc. ieee int.
  conf. comput. vis. workshops}\ (\BPG~1521-1530).
\PrintBackRefs{\CurrentBib}

\bibitem [\protect \citeauthoryear {%
Lan%
, Hu%
\BCBL {}\ \BBA {} Cheng%
}{%
Lan%
\ \protect \BOthers {.}}{%
{\protect \APACyear {2022}}%
}]{%
ATF}
\APACinsertmetastar {%
ATF}%
\begin{APACrefauthors}%
Lan, X.%
, Hu, Q.%
\BCBL {} Cheng, J.%
\end{APACrefauthors}%
\unskip\
\newblock
\APACrefYearMonthDay{2022}{}{}.
\newblock
{\BBOQ}\APACrefatitle {ATF: An Alternating Training Framework for Weakly
  Supervised Face Alignment} {Atf: An alternating training framework for weakly
  supervised face alignment}.{\BBCQ}
\newblock
\APACjournalVolNumPages{IEEE Trans. Multim.}{Early Access}{}{1-1,}
\newblock
\begin{APACrefDOI} \doi{10.1109/TMM.2022.3164798} \end{APACrefDOI}
\newblock

\newblock

\PrintBackRefs{\CurrentBib}

\bibitem [\protect \citeauthoryear {%
Le%
, Brandt%
, Lin%
, Bourdev%
\BCBL {}\ \BBA {} Huang%
}{%
Le%
\ \protect \BOthers {.}}{%
{\protect \APACyear {2012}}%
}]{%
HELEN}
\APACinsertmetastar {%
HELEN}%
\begin{APACrefauthors}%
Le, V.%
, Brandt, J.%
, Lin, Z.%
, Bourdev, L.%
\BCBL {} Huang, T.S.%
\end{APACrefauthors}%
\unskip\
\newblock
\APACrefYearMonthDay{2012}{}{}.
\newblock
{\BBOQ}\APACrefatitle {Interactive Facial Feature Localization} {Interactive
  facial feature localization}.{\BBCQ}
\newblock
 \APACrefbtitle {Proc. Eur. Conf. Comput. Vis.} {Proc. eur. conf. comput.
  vis.}\ (\BPG~679-692).
\PrintBackRefs{\CurrentBib}

\bibitem [\protect \citeauthoryear {%
Li%
\ \protect \BOthers {.}}{%
Li%
\ \protect \BOthers {.}}{%
{\protect \APACyear {2020}}%
}]{%
SDL}
\APACinsertmetastar {%
SDL}%
\begin{APACrefauthors}%
Li, W.%
, Lu, Y.%
, Zheng, K.%
, Liao, H.%
, Lin, C.%
, Luo, J.%
\BDBL {}Miao, S.%
\end{APACrefauthors}%
\unskip\
\newblock
\APACrefYearMonthDay{2020}{}{}.
\newblock
{\BBOQ}\APACrefatitle {Structured Landmark Detection via Topology-Adapting Deep
  Graph Learning} {Structured landmark detection via topology-adapting deep
  graph learning}.{\BBCQ}
\newblock
 \APACrefbtitle {Proc. Eur. Conf. Comput. Vis.} {Proc. eur. conf. comput.
  vis.}\ (\BPGS\ 266--283).
\newblock
\APACaddressPublisher{Cham}{Springer International Publishing}.
\PrintBackRefs{\CurrentBib}

\bibitem [\protect \citeauthoryear {%
Lin%
\ \protect \BOthers {.}}{%
Lin%
\ \protect \BOthers {.}}{%
{\protect \APACyear {2021}}%
}]{%
SCDF}
\APACinsertmetastar {%
SCDF}%
\begin{APACrefauthors}%
Lin, C.%
, Zhu, B.%
, Wang, Q.%
, Liao, R.%
, Qian, C.%
, Lu, J.%
\BCBL {} Zhou, J.%
\end{APACrefauthors}%
\unskip\
\newblock
\APACrefYearMonthDay{2021}{}{}.
\newblock
{\BBOQ}\APACrefatitle {Structure-Coherent Deep Feature Learning for Robust Face
  Alignment} {Structure-coherent deep feature learning for robust face
  alignment}.{\BBCQ}
\newblock
\APACjournalVolNumPages{IEEE Trans. Image Process.}{30}{}{5313-5326,}
\newblock
\begin{APACrefDOI} \doi{10.1109/TIP.2021.3082319} \end{APACrefDOI}
\newblock

\newblock

\PrintBackRefs{\CurrentBib}

\bibitem [\protect \citeauthoryear {%
H.~Liu%
, Lu%
, Guo%
, Wu%
\BCBL {}\ \BBA {} Zhou%
}{%
H.~Liu%
\ \protect \BOthers {.}}{%
{\protect \APACyear {2020}}%
}]{%
LDR}
\APACinsertmetastar {%
LDR}%
\begin{APACrefauthors}%
Liu, H.%
, Lu, J.%
, Guo, M.%
, Wu, S.%
\BCBL {} Zhou, J.%
\end{APACrefauthors}%
\unskip\
\newblock
\APACrefYearMonthDay{2020}{}{}.
\newblock
{\BBOQ}\APACrefatitle {Learning Reasoning-Decision Networks for Robust Face
  Alignment} {Learning reasoning-decision networks for robust face
  alignment}.{\BBCQ}
\newblock
\APACjournalVolNumPages{IEEE Trans. Pattern Anal. Mach.
  Intell.}{42}{3}{679-693,}
\newblock
\begin{APACrefDOI} \doi{10.1109/TPAMI.2018.2885298} \end{APACrefDOI}
\newblock

\newblock

\PrintBackRefs{\CurrentBib}

\bibitem [\protect \citeauthoryear {%
X.~Liu%
}{%
X.~Liu%
}{%
{\protect \APACyear {2009}}%
}]{%
AAM_E2}
\APACinsertmetastar {%
AAM_E2}%
\begin{APACrefauthors}%
Liu, X.%
\end{APACrefauthors}%
\unskip\
\newblock
\APACrefYearMonthDay{2009}{}{}.
\newblock
{\BBOQ}\APACrefatitle {Discriminative Face Alignment} {Discriminative face
  alignment}.{\BBCQ}
\newblock
\APACjournalVolNumPages{IEEE Trans. Pattern Anal. Mach.
  Intell.}{31}{11}{1941-1954,}
\newblock
\begin{APACrefDOI} \doi{10.1109/TPAMI.2008.238} \end{APACrefDOI}
\newblock

\newblock

\PrintBackRefs{\CurrentBib}

\bibitem [\protect \citeauthoryear {%
Z.~Liu%
, Luo%
, Wang%
\BCBL {}\ \BBA {} Tang%
}{%
Z.~Liu%
\ \protect \BOthers {.}}{%
{\protect \APACyear {2015}}%
}]{%
CelebA}
\APACinsertmetastar {%
CelebA}%
\begin{APACrefauthors}%
Liu, Z.%
, Luo, P.%
, Wang, X.%
\BCBL {} Tang, X.%
\end{APACrefauthors}%
\unskip\
\newblock
\APACrefYearMonthDay{2015}{}{}.
\newblock
{\BBOQ}\APACrefatitle {Deep Learning Face Attributes in the Wild} {Deep
  learning face attributes in the wild}.{\BBCQ}
\newblock
 \APACrefbtitle {Proc. IEEE Int. Conf. Comput. Vis.} {Proc. ieee int. conf.
  comput. vis.}\ (\BPG~3730-3738).
\PrintBackRefs{\CurrentBib}

\bibitem [\protect \citeauthoryear {%
Lorenz%
, Bereska%
, Milbich%
\BCBL {}\ \BBA {} Ommer%
}{%
Lorenz%
\ \protect \BOthers {.}}{%
{\protect \APACyear {2019}}%
}]{%
Unsuper_Lorenz}
\APACinsertmetastar {%
Unsuper_Lorenz}%
\begin{APACrefauthors}%
Lorenz, D.%
, Bereska, L.%
, Milbich, T.%
\BCBL {} Ommer, B.%
\end{APACrefauthors}%
\unskip\
\newblock
\APACrefYearMonthDay{2019}{}{}.
\newblock
{\BBOQ}\APACrefatitle {Unsupervised Part-Based Disentangling of Object Shape
  and Appearance} {Unsupervised part-based disentangling of object shape and
  appearance}.{\BBCQ}
\newblock
 \APACrefbtitle {Proc. IEEE Conf. Comput. Vis. Pattern Recognit.} {Proc. ieee
  conf. comput. vis. pattern recognit.}\ (\BPG~10947-10956).
\PrintBackRefs{\CurrentBib}

\bibitem [\protect \citeauthoryear {%
Loshchilov%
\ \BBA {} Hutter%
}{%
Loshchilov%
\ \BBA {} Hutter%
}{%
{\protect \APACyear {2019}}%
}]{%
AdamW}
\APACinsertmetastar {%
AdamW}%
\begin{APACrefauthors}%
Loshchilov, I.%
\BCBT {}\ \BBA {} Hutter, F.%
\end{APACrefauthors}%
\unskip\
\newblock
\APACrefYearMonthDay{2019}{}{}.
\newblock
{\BBOQ}\APACrefatitle {Decoupled Weight Decay Regularization} {Decoupled weight
  decay regularization}.{\BBCQ}
\newblock
 \APACrefbtitle {Proc. Int. Conf. Learn. Representations.} {Proc. int. conf.
  learn. representations.}
\PrintBackRefs{\CurrentBib}

\bibitem [\protect \citeauthoryear {%
Lv%
, Shao%
, Xing%
, Cheng%
\BCBL {}\ \BBA {} Zhou%
}{%
Lv%
\ \protect \BOthers {.}}{%
{\protect \APACyear {2017}}%
}]{%
Deep_Reg}
\APACinsertmetastar {%
Deep_Reg}%
\begin{APACrefauthors}%
Lv, J.%
, Shao, X.%
, Xing, J.%
, Cheng, C.%
\BCBL {} Zhou, X.%
\end{APACrefauthors}%
\unskip\
\newblock
\APACrefYearMonthDay{2017}{}{}.
\newblock
{\BBOQ}\APACrefatitle {A Deep Regression Architecture with Two-Stage
  Re-initialization for High Performance Facial Landmark Detection} {A deep
  regression architecture with two-stage re-initialization for high performance
  facial landmark detection}.{\BBCQ}
\newblock
 \APACrefbtitle {Proc. IEEE Conf. Comput. Vis. Pattern Recognit.} {Proc. ieee
  conf. comput. vis. pattern recognit.}\ (\BPG~3691-3700).
\PrintBackRefs{\CurrentBib}

\bibitem [\protect \citeauthoryear {%
Mallis%
, Sanchez%
, Bell%
\BCBL {}\ \BBA {} Tzimiropoulos%
}{%
Mallis%
\ \protect \BOthers {.}}{%
{\protect \APACyear {2023}}%
}]{%
SelfLandmark}
\APACinsertmetastar {%
SelfLandmark}%
\begin{APACrefauthors}%
Mallis, D.%
, Sanchez, E.%
, Bell, M.%
\BCBL {} Tzimiropoulos, G.%
\end{APACrefauthors}%
\unskip\
\newblock
\APACrefYearMonthDay{2023}{}{}.
\newblock
{\BBOQ}\APACrefatitle {From Keypoints to Object Landmarks via Self-Training
  Correspondence: A Novel Approach to Unsupervised Landmark Discovery} {From
  keypoints to object landmarks via self-training correspondence: A novel
  approach to unsupervised landmark discovery}.{\BBCQ}
\newblock
\APACjournalVolNumPages{IEEE Trans. Pattern Anal. Mach.
  Intell.}{45}{7}{8390-8404,}
\newblock
\begin{APACrefDOI} \doi{10.1109/TPAMI.2023.3234212} \end{APACrefDOI}
\newblock

\newblock

\PrintBackRefs{\CurrentBib}

\bibitem [\protect \citeauthoryear {%
Newell%
, Yang%
\BCBL {}\ \BBA {} Deng%
}{%
Newell%
\ \protect \BOthers {.}}{%
{\protect \APACyear {2016}}%
}]{%
Hourglass}
\APACinsertmetastar {%
Hourglass}%
\begin{APACrefauthors}%
Newell, A.%
, Yang, K.%
\BCBL {} Deng, J.%
\end{APACrefauthors}%
\unskip\
\newblock
\APACrefYearMonthDay{2016}{}{}.
\newblock
{\BBOQ}\APACrefatitle {Stacked Hourglass Networks for Human Pose Estimation}
  {Stacked hourglass networks for human pose estimation}.{\BBCQ}
\newblock
 \APACrefbtitle {Proc. Eur. Conf. Comput. Vis.} {Proc. eur. conf. comput.
  vis.}\ (\BPGS\ 483--499).
\PrintBackRefs{\CurrentBib}

\bibitem [\protect \citeauthoryear {%
Otberdout%
, Daoudi%
, Kacem%
, Ballihi%
\BCBL {}\ \BBA {} Berretti%
}{%
Otberdout%
\ \protect \BOthers {.}}{%
{\protect \APACyear {2022}}%
}]{%
Dynamic_Facial_Expression}
\APACinsertmetastar {%
Dynamic_Facial_Expression}%
\begin{APACrefauthors}%
Otberdout, N.%
, Daoudi, M.%
, Kacem, A.%
, Ballihi, L.%
\BCBL {} Berretti, S.%
\end{APACrefauthors}%
\unskip\
\newblock
\APACrefYearMonthDay{2022}{}{}.
\newblock
{\BBOQ}\APACrefatitle {Dynamic Facial Expression Generation on Hilbert
  Hypersphere With Conditional Wasserstein Generative Adversarial Nets}
  {Dynamic facial expression generation on hilbert hypersphere with conditional
  wasserstein generative adversarial nets}.{\BBCQ}
\newblock
\APACjournalVolNumPages{IEEE Trans. Pattern Anal. Mach.
  Intell.}{44}{2}{848-863,}
\newblock
\begin{APACrefDOI} \doi{10.1109/TPAMI.2020.3002500} \end{APACrefDOI}
\newblock

\newblock

\PrintBackRefs{\CurrentBib}

\bibitem [\protect \citeauthoryear {%
Pourpanah%
\ \protect \BOthers {.}}{%
Pourpanah%
\ \protect \BOthers {.}}{%
{\protect \APACyear {2023}}%
}]{%
ZeroshotReview}
\APACinsertmetastar {%
ZeroshotReview}%
\begin{APACrefauthors}%
Pourpanah, F.%
, Abdar, M.%
, Luo, Y.%
, Zhou, X.%
, Wang, R.%
, Lim, C.P.%
\BDBL {}Wu, Q.M.J.%
\end{APACrefauthors}%
\unskip\
\newblock
\APACrefYearMonthDay{2023}{}{}.
\newblock
{\BBOQ}\APACrefatitle {A Review of Generalized Zero-Shot Learning Methods} {A
  review of generalized zero-shot learning methods}.{\BBCQ}
\newblock
\APACjournalVolNumPages{IEEE Trans. Pattern Anal. Mach.
  Intell.}{45}{4}{4051-4070,}
\newblock
\begin{APACrefDOI} \doi{10.1109/TPAMI.2022.3191696} \end{APACrefDOI}
\newblock

\newblock

\PrintBackRefs{\CurrentBib}

\bibitem [\protect \citeauthoryear {%
Prados-Torreblanca%
, Buenaposada%
\BCBL {}\ \BBA {} Baumela%
}{%
Prados-Torreblanca%
\ \protect \BOthers {.}}{%
{\protect \APACyear {2022}}%
}]{%
SPIGA}
\APACinsertmetastar {%
SPIGA}%
\begin{APACrefauthors}%
Prados-Torreblanca, A.%
, Buenaposada, J.M.%
\BCBL {} Baumela, L.%
\end{APACrefauthors}%
\unskip\
\newblock
\APACrefYearMonthDay{2022}{}{}.
\newblock
{\BBOQ}\APACrefatitle {Shape Preserving Facial Landmarks with Graph Attention
  Networks} {Shape preserving facial landmarks with graph attention
  networks}.{\BBCQ}
\newblock
 \APACrefbtitle {Proc. British Mach. Vis. Conf.} {Proc. british mach. vis.
  conf.}
\PrintBackRefs{\CurrentBib}

\bibitem [\protect \citeauthoryear {%
Qian%
, Sun%
, Wu%
, Qian%
\BCBL {}\ \BBA {} Jia%
}{%
Qian%
\ \protect \BOthers {.}}{%
{\protect \APACyear {2019}}%
}]{%
AVS}
\APACinsertmetastar {%
AVS}%
\begin{APACrefauthors}%
Qian, S.%
, Sun, K.%
, Wu, W.%
, Qian, C.%
\BCBL {} Jia, J.%
\end{APACrefauthors}%
\unskip\
\newblock
\APACrefYearMonthDay{2019}{}{}.
\newblock
{\BBOQ}\APACrefatitle {Aggregation via Separation: Boosting Facial Landmark
  Detector With Semi-Supervised Style Translation} {Aggregation via separation:
  Boosting facial landmark detector with semi-supervised style
  translation}.{\BBCQ}
\newblock
 \APACrefbtitle {Proc. IEEE Int. Conf. Comput. Vis.} {Proc. ieee int. conf.
  comput. vis.}\ (\BPG~10152-10162).
\PrintBackRefs{\CurrentBib}

\bibitem [\protect \citeauthoryear {%
Sagonas%
, Tzimiropoulos%
, Zafeiriou%
\BCBL {}\ \BBA {} Pantic%
}{%
Sagonas%
\ \protect \BOthers {.}}{%
{\protect \APACyear {2013}}%
{\protect \APACexlab {{\protect \BCnt {1}}}}}]{%
300W}
\APACinsertmetastar {%
300W}%
\begin{APACrefauthors}%
Sagonas, C.%
, Tzimiropoulos, G.%
, Zafeiriou, S.%
\BCBL {} Pantic, M.%
\end{APACrefauthors}%
\unskip\
\newblock
\APACrefYearMonthDay{2013{\protect \BCnt {1}}}{}{}.
\newblock
{\BBOQ}\APACrefatitle {300 Faces in-the-Wild Challenge: The First Facial
  Landmark Localization Challenge} {300 faces in-the-wild challenge: The first
  facial landmark localization challenge}.{\BBCQ}
\newblock
 \APACrefbtitle {Proc. IEEE Int. Conf. Comput. Vis. Workshops} {Proc. ieee int.
  conf. comput. vis. workshops}\ (\BPG~397-403).
\PrintBackRefs{\CurrentBib}

\bibitem [\protect \citeauthoryear {%
Sagonas%
, Tzimiropoulos%
, Zafeiriou%
\BCBL {}\ \BBA {} Pantic%
}{%
Sagonas%
\ \protect \BOthers {.}}{%
{\protect \APACyear {2013}}%
{\protect \APACexlab {{\protect \BCnt {2}}}}}]{%
300W-A}
\APACinsertmetastar {%
300W-A}%
\begin{APACrefauthors}%
Sagonas, C.%
, Tzimiropoulos, G.%
, Zafeiriou, S.%
\BCBL {} Pantic, M.%
\end{APACrefauthors}%
\unskip\
\newblock
\APACrefYearMonthDay{2013{\protect \BCnt {2}}}{}{}.
\newblock
{\BBOQ}\APACrefatitle {A Semi-automatic Methodology for Facial Landmark
  Annotation} {A semi-automatic methodology for facial landmark
  annotation}.{\BBCQ}
\newblock
 \APACrefbtitle {Proc. IEEE Conf. Comput. Vis. Pattern Recognit. Workshops}
  {Proc. ieee conf. comput. vis. pattern recognit. workshops}\ (\BPG~896-903).
\PrintBackRefs{\CurrentBib}

\bibitem [\protect \citeauthoryear {%
Tai%
\ \protect \BOthers {.}}{%
Tai%
\ \protect \BOthers {.}}{%
{\protect \APACyear {2019}}%
}]{%
THA}
\APACinsertmetastar {%
THA}%
\begin{APACrefauthors}%
Tai, Y.%
, Liang, Y.%
, Liu, X.%
, Duan, L.%
, Li, J.%
, Wang, C.%
\BDBL {}Chen, Y.%
\end{APACrefauthors}%
\unskip\
\newblock
\APACrefYearMonthDay{2019}{}{}.
\newblock
{\BBOQ}\APACrefatitle {Towards Highly Accurate and Stable Face Alignment for
  High-Resolution Videos} {Towards highly accurate and stable face alignment
  for high-resolution videos}.{\BBCQ}
\newblock
 \APACrefbtitle {Proc. AAAI Conf. Artif. Intell.} {Proc. aaai conf. artif.
  intell.}\ (\BVOL~33, \BPGS\ 8893--8900).
\PrintBackRefs{\CurrentBib}

\bibitem [\protect \citeauthoryear {%
Tanfous%
, Drira%
\BCBL {}\ \BBA {} Amor%
}{%
Tanfous%
\ \protect \BOthers {.}}{%
{\protect \APACyear {2020}}%
}]{%
Sparse_Code_Facial_Expressing}
\APACinsertmetastar {%
Sparse_Code_Facial_Expressing}%
\begin{APACrefauthors}%
Tanfous, A.B.%
, Drira, H.%
\BCBL {} Amor, B.B.%
\end{APACrefauthors}%
\unskip\
\newblock
\APACrefYearMonthDay{2020}{}{}.
\newblock
{\BBOQ}\APACrefatitle {Sparse Coding of Shape Trajectories for Facial
  Expression and Action Recognition} {Sparse coding of shape trajectories for
  facial expression and action recognition}.{\BBCQ}
\newblock
\APACjournalVolNumPages{IEEE Trans. Pattern Anal. Mach.
  Intell.}{42}{10}{2594-2607,}
\newblock
\begin{APACrefDOI} \doi{10.1109/TPAMI.2019.2932979} \end{APACrefDOI}
\newblock

\newblock

\PrintBackRefs{\CurrentBib}

\bibitem [\protect \citeauthoryear {%
H.~Tang%
, Shao%
, Torr%
\BCBL {}\ \BBA {} Sebe%
}{%
H.~Tang%
\ \protect \BOthers {.}}{%
{\protect \APACyear {2023}}%
}]{%
IJCVGAN}
\APACinsertmetastar {%
IJCVGAN}%
\begin{APACrefauthors}%
Tang, H.%
, Shao, L.%
, Torr, P.H.%
\BCBL {} Sebe, N.%
\end{APACrefauthors}%
\unskip\
\newblock
\APACrefYearMonthDay{2023}{}{}.
\newblock
{\BBOQ}\APACrefatitle {Bipartite graph reasoning gans for person pose and
  facial image synthesis} {Bipartite graph reasoning gans for person pose and
  facial image synthesis}.{\BBCQ}
\newblock
\APACjournalVolNumPages{Int. J. Comput. Vis.}{131}{3}{644--658,}
\newblock
\begin{APACrefDOI} \doi{10.1007/s11263-022-01722-5} \end{APACrefDOI}
\newblock

\newblock

\PrintBackRefs{\CurrentBib}

\bibitem [\protect \citeauthoryear {%
Z.~Tang%
, Peng%
, Li%
\BCBL {}\ \BBA {} Metaxas%
}{%
Z.~Tang%
\ \protect \BOthers {.}}{%
{\protect \APACyear {2020}}%
}]{%
Dunet}
\APACinsertmetastar {%
Dunet}%
\begin{APACrefauthors}%
Tang, Z.%
, Peng, X.%
, Li, K.%
\BCBL {} Metaxas, D.N.%
\end{APACrefauthors}%
\unskip\
\newblock
\APACrefYearMonthDay{2020}{}{}.
\newblock
{\BBOQ}\APACrefatitle {Towards Efficient U-Nets: A Coupled and Quantized
  Approach} {Towards efficient u-nets: A coupled and quantized
  approach}.{\BBCQ}
\newblock
\APACjournalVolNumPages{IEEE Trans. Pattern Anal. Mach.
  Intell.}{42}{8}{2038-2050,}
\newblock
\begin{APACrefDOI} \doi{10.1109/TPAMI.2019.2907634} \end{APACrefDOI}
\newblock

\newblock

\PrintBackRefs{\CurrentBib}

\bibitem [\protect \citeauthoryear {%
Thewlis%
, Bilen%
\BCBL {}\ \BBA {} Vedaldi%
}{%
Thewlis%
\ \protect \BOthers {.}}{%
{\protect \APACyear {2017}}%
}]{%
Unsuper_Thewlis}
\APACinsertmetastar {%
Unsuper_Thewlis}%
\begin{APACrefauthors}%
Thewlis, J.%
, Bilen, H.%
\BCBL {} Vedaldi, A.%
\end{APACrefauthors}%
\unskip\
\newblock
\APACrefYearMonthDay{2017}{}{}.
\newblock
{\BBOQ}\APACrefatitle {Unsupervised Learning of Object Landmarks by Factorized
  Spatial Embeddings} {Unsupervised learning of object landmarks by factorized
  spatial embeddings}.{\BBCQ}
\newblock
 \APACrefbtitle {Proc. IEEE Int. Conf. Comput. Vis.} {Proc. ieee int. conf.
  comput. vis.}\ (\BPG~3229-3238).
\PrintBackRefs{\CurrentBib}

\bibitem [\protect \citeauthoryear {%
Trigeorgis%
, Snape%
, Nicolaou%
, Antonakos%
\BCBL {}\ \BBA {} Zafeiriou%
}{%
Trigeorgis%
\ \protect \BOthers {.}}{%
{\protect \APACyear {2016}}%
}]{%
MDM}
\APACinsertmetastar {%
MDM}%
\begin{APACrefauthors}%
Trigeorgis, G.%
, Snape, P.%
, Nicolaou, M.A.%
, Antonakos, E.%
\BCBL {} Zafeiriou, S.%
\end{APACrefauthors}%
\unskip\
\newblock
\APACrefYearMonthDay{2016}{}{}.
\newblock
{\BBOQ}\APACrefatitle {Mnemonic Descent Method: A Recurrent Process Applied for
  End-to-End Face Alignment} {Mnemonic descent method: A recurrent process
  applied for end-to-end face alignment}.{\BBCQ}
\newblock
 \APACrefbtitle {Proc. IEEE Conf. Comput. Vis. Pattern Recognit.} {Proc. ieee
  conf. comput. vis. pattern recognit.}\ (\BPG~4177-4187).
\PrintBackRefs{\CurrentBib}

\bibitem [\protect \citeauthoryear {%
Valle%
, Buenaposada%
, Vald{\'e}s%
\BCBL {}\ \BBA {} Baumela%
}{%
Valle%
\ \protect \BOthers {.}}{%
{\protect \APACyear {2018}}%
}]{%
DCFE}
\APACinsertmetastar {%
DCFE}%
\begin{APACrefauthors}%
Valle, R.%
, Buenaposada, J.M.%
, Vald{\'e}s, A.%
\BCBL {} Baumela, L.%
\end{APACrefauthors}%
\unskip\
\newblock
\APACrefYearMonthDay{2018}{}{}.
\newblock
{\BBOQ}\APACrefatitle {A Deeply-Initialized Coarse-to-fine Ensemble of
  Regression Trees for Face Alignment} {A deeply-initialized coarse-to-fine
  ensemble of regression trees for face alignment}.{\BBCQ}
\newblock
 \APACrefbtitle {Proc. Eur. Conf. Comput. Vis.} {Proc. eur. conf. comput.
  vis.}\ (\BPGS\ 609--624).
\newblock
\APACaddressPublisher{Cham}{}.
\PrintBackRefs{\CurrentBib}

\bibitem [\protect \citeauthoryear {%
J.~Wang%
\ \protect \BOthers {.}}{%
J.~Wang%
\ \protect \BOthers {.}}{%
{\protect \APACyear {2021}}%
}]{%
HRNet}
\APACinsertmetastar {%
HRNet}%
\begin{APACrefauthors}%
Wang, J.%
, Sun, K.%
, Cheng, T.%
, Jiang, B.%
, Deng, C.%
, Zhao, Y.%
\BDBL {}Xiao, B.%
\end{APACrefauthors}%
\unskip\
\newblock
\APACrefYearMonthDay{2021}{}{}.
\newblock
{\BBOQ}\APACrefatitle {Deep High-Resolution Representation Learning for Visual
  Recognition} {Deep high-resolution representation learning for visual
  recognition}.{\BBCQ}
\newblock
\APACjournalVolNumPages{IEEE Trans. Pattern Anal. Mach.
  Intell.}{43}{10}{3349-3364,}
\newblock
\begin{APACrefDOI} \doi{10.1109/TPAMI.2020.2983686} \end{APACrefDOI}
\newblock

\newblock

\PrintBackRefs{\CurrentBib}

\bibitem [\protect \citeauthoryear {%
X.~Wang%
, Bo%
\BCBL {}\ \BBA {} Fuxin%
}{%
X.~Wang%
\ \protect \BOthers {.}}{%
{\protect \APACyear {2019}}%
}]{%
Awing}
\APACinsertmetastar {%
Awing}%
\begin{APACrefauthors}%
Wang, X.%
, Bo, L.%
\BCBL {} Fuxin, L.%
\end{APACrefauthors}%
\unskip\
\newblock
\APACrefYearMonthDay{2019}{}{}.
\newblock
{\BBOQ}\APACrefatitle {Adaptive Wing Loss for Robust Face Alignment via Heatmap
  Regression} {Adaptive wing loss for robust face alignment via heatmap
  regression}.{\BBCQ}
\newblock
 \APACrefbtitle {Proc. IEEE Int. Conf. Comput. Vis.} {Proc. ieee int. conf.
  comput. vis.}\ (\BPG~6970-6980).
\PrintBackRefs{\CurrentBib}

\bibitem [\protect \citeauthoryear {%
Wu%
\ \protect \BOthers {.}}{%
Wu%
\ \protect \BOthers {.}}{%
{\protect \APACyear {2018}}%
}]{%
LAB}
\APACinsertmetastar {%
LAB}%
\begin{APACrefauthors}%
Wu, W.%
, Qian, C.%
, Yang, S.%
, Wang, Q.%
, Cai, Y.%
\BCBL {} Zhou, Q.%
\end{APACrefauthors}%
\unskip\
\newblock
\APACrefYearMonthDay{2018}{}{}.
\newblock
{\BBOQ}\APACrefatitle {Look at Boundary: A Boundary-Aware Face Alignment
  Algorithm} {Look at boundary: A boundary-aware face alignment
  algorithm}.{\BBCQ}
\newblock
 \APACrefbtitle {Proc. IEEE Conf. Comput. Vis. Pattern Recognit.} {Proc. ieee
  conf. comput. vis. pattern recognit.}\ (\BPG~2129-2138).
\PrintBackRefs{\CurrentBib}

\bibitem [\protect \citeauthoryear {%
Wu%
\ \BBA {} Yang%
}{%
Wu%
\ \BBA {} Yang%
}{%
{\protect \APACyear {2017}}%
}]{%
DVLN}
\APACinsertmetastar {%
DVLN}%
\begin{APACrefauthors}%
Wu, W.%
\BCBT {}\ \BBA {} Yang, S.%
\end{APACrefauthors}%
\unskip\
\newblock
\APACrefYearMonthDay{2017}{}{}.
\newblock
{\BBOQ}\APACrefatitle {Leveraging Intra and Inter-Dataset Variations for Robust
  Face Alignment} {Leveraging intra and inter-dataset variations for robust
  face alignment}.{\BBCQ}
\newblock
 \APACrefbtitle {Proc. IEEE Conf. Comput. Vis. Pattern Recognit. Workshops}
  {Proc. ieee conf. comput. vis. pattern recognit. workshops}\
  (\BPG~2096-2105).
\PrintBackRefs{\CurrentBib}

\bibitem [\protect \citeauthoryear {%
Xia%
\ \protect \BOthers {.}}{%
Xia%
\ \protect \BOthers {.}}{%
{\protect \APACyear {2022}}%
}]{%
SLPT}
\APACinsertmetastar {%
SLPT}%
\begin{APACrefauthors}%
Xia, J.%
, Qu, W.%
, Huang, W.%
, Zhang, J.%
, Wang, X.%
\BCBL {} Xu, M.%
\end{APACrefauthors}%
\unskip\
\newblock
\APACrefYearMonthDay{2022}{}{}.
\newblock
{\BBOQ}\APACrefatitle {Sparse Local Patch Transformer for Robust Face Alignment
  and Landmarks Inherent Relation Learning} {Sparse local patch transformer for
  robust face alignment and landmarks inherent relation learning}.{\BBCQ}
\newblock
 \APACrefbtitle {Proc. IEEE Conf. Comput. Vis. Pattern Recognit.} {Proc. ieee
  conf. comput. vis. pattern recognit.}\ (\BPG~4052-4061).
\PrintBackRefs{\CurrentBib}

\bibitem [\protect \citeauthoryear {%
Xia%
\ \protect \BOthers {.}}{%
Xia%
\ \protect \BOthers {.}}{%
{\protect \APACyear {2023}}%
}]{%
DSLPT}
\APACinsertmetastar {%
DSLPT}%
\begin{APACrefauthors}%
Xia, J.%
, Xu, M.%
, Zhang, H.%
, Zhang, J.%
, Huang, W.%
, Cao, H.%
\BCBL {} Wen, S.%
\end{APACrefauthors}%
\unskip\
\newblock
\APACrefYearMonthDay{2023}{}{}.
\newblock
{\BBOQ}\APACrefatitle {Robust Face Alignment via Inherent Relation Learning and
  Uncertainty Estimation} {Robust face alignment via inherent relation learning
  and uncertainty estimation}.{\BBCQ}
\newblock
\APACjournalVolNumPages{IEEE Trans. Pattern Anal. Mach.
  Intell.}{45}{8}{10358-10375,}
\newblock
\begin{APACrefDOI} \doi{10.1109/TPAMI.2023.3260926} \end{APACrefDOI}
\newblock

\newblock

\PrintBackRefs{\CurrentBib}

\bibitem [\protect \citeauthoryear {%
B.~Xiao%
, Wu%
\BCBL {}\ \BBA {} Wei%
}{%
B.~Xiao%
\ \protect \BOthers {.}}{%
{\protect \APACyear {2018}}%
}]{%
SimBa}
\APACinsertmetastar {%
SimBa}%
\begin{APACrefauthors}%
Xiao, B.%
, Wu, H.%
\BCBL {} Wei, Y.%
\end{APACrefauthors}%
\unskip\
\newblock
\APACrefYearMonthDay{2018}{}{}.
\newblock
{\BBOQ}\APACrefatitle {Simple Baselines for Human Pose Estimation and Tracking}
  {Simple baselines for human pose estimation and tracking}.{\BBCQ}
\newblock
 \APACrefbtitle {Proc. Eur. Conf. Comput. Vis.} {Proc. eur. conf. comput.
  vis.}\ (\BPGS\ 472--487).
\PrintBackRefs{\CurrentBib}

\bibitem [\protect \citeauthoryear {%
S.~Xiao%
\ \protect \BOthers {.}}{%
S.~Xiao%
\ \protect \BOthers {.}}{%
{\protect \APACyear {2016}}%
}]{%
RAR}
\APACinsertmetastar {%
RAR}%
\begin{APACrefauthors}%
Xiao, S.%
, Feng, J.%
, Xing, J.%
, Lai, H.%
, Yan, S.%
\BCBL {} Kassim, A.%
\end{APACrefauthors}%
\unskip\
\newblock
\APACrefYearMonthDay{2016}{}{}.
\newblock
{\BBOQ}\APACrefatitle {Robust Facial Landmark Detection via Recurrent
  Attentive-Refinement Networks} {Robust facial landmark detection via
  recurrent attentive-refinement networks}.{\BBCQ}
\newblock
 \APACrefbtitle {Proc. Eur. Conf. Comput. Vis.} {Proc. eur. conf. comput.
  vis.}\ (\BPGS\ 57--72).
\newblock
\APACaddressPublisher{Cham}{}.
\PrintBackRefs{\CurrentBib}

\bibitem [\protect \citeauthoryear {%
Xiong%
\ \BBA {} De~la Torre%
}{%
Xiong%
\ \BBA {} De~la Torre%
}{%
{\protect \APACyear {2013}}%
}]{%
SDM}
\APACinsertmetastar {%
SDM}%
\begin{APACrefauthors}%
Xiong, X.%
\BCBT {}\ \BBA {} De~la Torre, F.%
\end{APACrefauthors}%
\unskip\
\newblock
\APACrefYearMonthDay{2013}{}{}.
\newblock
{\BBOQ}\APACrefatitle {Supervised Descent Method and Its Applications to Face
  Alignment} {Supervised descent method and its applications to face
  alignment}.{\BBCQ}
\newblock
 \APACrefbtitle {Proc. IEEE Conf. Comput. Vis. Pattern Recognit.} {Proc. ieee
  conf. comput. vis. pattern recognit.}\ (\BPG~532-539).
\PrintBackRefs{\CurrentBib}

\bibitem [\protect \citeauthoryear {%
Xu%
\ \protect \BOthers {.}}{%
Xu%
\ \protect \BOthers {.}}{%
{\protect \APACyear {2021}}%
}]{%
FaceAAAI}
\APACinsertmetastar {%
FaceAAAI}%
\begin{APACrefauthors}%
Xu, X.%
, Meng, Q.%
, Qin, Y.%
, Guo, J.%
, Zhao, C.%
, Zhou, F.%
\BCBL {} Lei, Z.%
\end{APACrefauthors}%
\unskip\
\newblock
\APACrefYearMonthDay{2021}{}{}.
\newblock
{\BBOQ}\APACrefatitle {Searching for alignment in face recognition} {Searching
  for alignment in face recognition}.{\BBCQ}
\newblock
 \APACrefbtitle {Proc. AAAI Conf. Artif. Intell.} {Proc. aaai conf. artif.
  intell.}\ (\BVOL~35, \BPGS\ 3065--3073).
\PrintBackRefs{\CurrentBib}

\bibitem [\protect \citeauthoryear {%
J.~Yang%
, Liu%
\BCBL {}\ \BBA {} Zhang%
}{%
J.~Yang%
\ \protect \BOthers {.}}{%
{\protect \APACyear {2017}}%
}]{%
SHN}
\APACinsertmetastar {%
SHN}%
\begin{APACrefauthors}%
Yang, J.%
, Liu, Q.%
\BCBL {} Zhang, K.%
\end{APACrefauthors}%
\unskip\
\newblock
\APACrefYearMonthDay{2017}{}{}.
\newblock
{\BBOQ}\APACrefatitle {Stacked Hourglass Network for Robust Facial Landmark
  Localisation} {Stacked hourglass network for robust facial landmark
  localisation}.{\BBCQ}
\newblock
 \APACrefbtitle {Proc. IEEE Conf. Comput. Vis. Pattern Recognit. Workshops}
  {Proc. ieee conf. comput. vis. pattern recognit. workshops}\
  (\BPG~2025-2033).
\PrintBackRefs{\CurrentBib}

\bibitem [\protect \citeauthoryear {%
S.~Yang%
, Luo%
, Loy%
\BCBL {}\ \BBA {} Tang%
}{%
S.~Yang%
\ \protect \BOthers {.}}{%
{\protect \APACyear {2016}}%
}]{%
WIDER}
\APACinsertmetastar {%
WIDER}%
\begin{APACrefauthors}%
Yang, S.%
, Luo, P.%
, Loy, C\BHBI C.%
\BCBL {} Tang, X.%
\end{APACrefauthors}%
\unskip\
\newblock
\APACrefYearMonthDay{2016}{June}{}.
\newblock
{\BBOQ}\APACrefatitle {WIDER FACE: A Face Detection Benchmark} {Wider face: A
  face detection benchmark}.{\BBCQ}
\newblock
 \APACrefbtitle {Proc. IEEE Conf. Comput. Vis. Pattern Recognit.} {Proc. ieee
  conf. comput. vis. pattern recognit.}
\PrintBackRefs{\CurrentBib}

\bibitem [\protect \citeauthoryear {%
H.~Zhang%
\ \protect \BOthers {.}}{%
H.~Zhang%
\ \protect \BOthers {.}}{%
{\protect \APACyear {2024}}%
}]{%
OpenAnimal}
\APACinsertmetastar {%
OpenAnimal}%
\begin{APACrefauthors}%
Zhang, H.%
, Xu, L.%
, Lai, S.%
, Shao, W.%
, Zheng, N.%
, Luo, P.%
\BDBL {}Zhang, K.%
\end{APACrefauthors}%
\unskip\
\newblock
\APACrefYearMonthDay{2024}{}{}.
\newblock
{\BBOQ}\APACrefatitle {Open-vocabulary animal keypoint detection with
  semantic-feature matching} {Open-vocabulary animal keypoint detection with
  semantic-feature matching}.{\BBCQ}
\newblock
\APACjournalVolNumPages{Int. J. Comput. Vis.}{}{}{1--18,}
\newblock
\begin{APACrefDOI} \doi{10.1007/s11263-024-02126-3} \end{APACrefDOI}
\newblock

\newblock

\PrintBackRefs{\CurrentBib}

\bibitem [\protect \citeauthoryear {%
J.~Zhang%
, Hu%
\BCBL {}\ \BBA {} Feng%
}{%
J.~Zhang%
\ \protect \BOthers {.}}{%
{\protect \APACyear {2020}}%
}]{%
SNH}
\APACinsertmetastar {%
SNH}%
\begin{APACrefauthors}%
Zhang, J.%
, Hu, H.%
\BCBL {} Feng, S.%
\end{APACrefauthors}%
\unskip\
\newblock
\APACrefYearMonthDay{2020}{}{}.
\newblock
{\BBOQ}\APACrefatitle {Robust Facial Landmark Detection via Heatmap-Offset
  Regression} {Robust facial landmark detection via heatmap-offset
  regression}.{\BBCQ}
\newblock
\APACjournalVolNumPages{IEEE Trans. Image Process.}{29}{}{5050-5064,}
\newblock
\begin{APACrefDOI} \doi{10.1109/TIP.2020.2976765} \end{APACrefDOI}
\newblock

\newblock

\PrintBackRefs{\CurrentBib}

\bibitem [\protect \citeauthoryear {%
Y.~Zhang%
\ \protect \BOthers {.}}{%
Y.~Zhang%
\ \protect \BOthers {.}}{%
{\protect \APACyear {2018}}%
}]{%
Unsuper_Zhang}
\APACinsertmetastar {%
Unsuper_Zhang}%
\begin{APACrefauthors}%
Zhang, Y.%
, Guo, Y.%
, Jin, Y.%
, Luo, Y.%
, He, Z.%
\BCBL {} Lee, H.%
\end{APACrefauthors}%
\unskip\
\newblock
\APACrefYearMonthDay{2018}{}{}.
\newblock
{\BBOQ}\APACrefatitle {Unsupervised Discovery of Object Landmarks as Structural
  Representations} {Unsupervised discovery of object landmarks as structural
  representations}.{\BBCQ}
\newblock
 \APACrefbtitle {Proc. IEEE Conf. Comput. Vis. Pattern Recognit.} {Proc. ieee
  conf. comput. vis. pattern recognit.}\ (\BPG~2694-2703).
\PrintBackRefs{\CurrentBib}

\bibitem [\protect \citeauthoryear {%
Zheng%
\ \protect \BOthers {.}}{%
Zheng%
\ \protect \BOthers {.}}{%
{\protect \APACyear {2022}}%
}]{%
FaRL}
\APACinsertmetastar {%
FaRL}%
\begin{APACrefauthors}%
Zheng, Y.%
, Yang, H.%
, Zhang, T.%
, Bao, J.%
, Chen, D.%
, Huang, Y.%
\BDBL {}Wen, F.%
\end{APACrefauthors}%
\unskip\
\newblock
\APACrefYearMonthDay{2022}{June}{}.
\newblock
{\BBOQ}\APACrefatitle {General Facial Representation Learning in a
  Visual-Linguistic Manner} {General facial representation learning in a
  visual-linguistic manner}.{\BBCQ}
\newblock
 \APACrefbtitle {Proc. IEEE Conf. Comput. Vis. Pattern Recognit.} {Proc. ieee
  conf. comput. vis. pattern recognit.}\ (\BPG~18697-18709).
\PrintBackRefs{\CurrentBib}

\bibitem [\protect \citeauthoryear {%
Zhou%
\ \protect \BOthers {.}}{%
Zhou%
\ \protect \BOthers {.}}{%
{\protect \APACyear {2023}}%
}]{%
StarLoss}
\APACinsertmetastar {%
StarLoss}%
\begin{APACrefauthors}%
Zhou, Z.%
, Li, H.%
, Liu, H.%
, Wang, N.%
, Yu, G.%
\BCBL {} Ji, R.%
\end{APACrefauthors}%
\unskip\
\newblock
\APACrefYearMonthDay{2023}{}{}.
\newblock
{\BBOQ}\APACrefatitle {STAR Loss: Reducing Semantic Ambiguity in Facial
  Landmark Detection} {Star loss: Reducing semantic ambiguity in facial
  landmark detection}.{\BBCQ}
\newblock
 \APACrefbtitle {Proc. IEEE Conf. Comput. Vis. Pattern Recognit.} {Proc. ieee
  conf. comput. vis. pattern recognit.}\ (\BPG~15475-15484).
\PrintBackRefs{\CurrentBib}

\bibitem [\protect \citeauthoryear {%
C.~Zhu%
, Li%
, Li%
\BCBL {}\ \BBA {} Dai%
}{%
C.~Zhu%
\ \protect \BOthers {.}}{%
{\protect \APACyear {2021}}%
}]{%
SAAT}
\APACinsertmetastar {%
SAAT}%
\begin{APACrefauthors}%
Zhu, C.%
, Li, X.%
, Li, J.%
\BCBL {} Dai, S.%
\end{APACrefauthors}%
\unskip\
\newblock
\APACrefYearMonthDay{2021}{}{}.
\newblock
{\BBOQ}\APACrefatitle {Improving Robustness of Facial Landmark Detection by
  Defending against Adversarial Attacks} {Improving robustness of facial
  landmark detection by defending against adversarial attacks}.{\BBCQ}
\newblock
 \APACrefbtitle {Proc. IEEE Int. Conf. Comput. Vis.} {Proc. ieee int. conf.
  comput. vis.}\ (\BPG~11731-11740).
\PrintBackRefs{\CurrentBib}

\bibitem [\protect \citeauthoryear {%
C.~Zhu%
, Wan%
, Xie%
, Li%
\BCBL {}\ \BBA {} Gu%
}{%
C.~Zhu%
\ \protect \BOthers {.}}{%
{\protect \APACyear {2022}}%
}]{%
GlomFace}
\APACinsertmetastar {%
GlomFace}%
\begin{APACrefauthors}%
Zhu, C.%
, Wan, X.%
, Xie, S.%
, Li, X.%
\BCBL {} Gu, Y.%
\end{APACrefauthors}%
\unskip\
\newblock
\APACrefYearMonthDay{2022}{}{}.
\newblock
{\BBOQ}\APACrefatitle {Occlusion-robust Face Alignment using A
  Viewpoint-invariant Hierarchical Network Architecture} {Occlusion-robust face
  alignment using a viewpoint-invariant hierarchical network
  architecture}.{\BBCQ}
\newblock
 \APACrefbtitle {Proc. IEEE Conf. Comput. Vis. Pattern Recognit.} {Proc. ieee
  conf. comput. vis. pattern recognit.}\ (\BPG~11102-11111).
\PrintBackRefs{\CurrentBib}

\bibitem [\protect \citeauthoryear {%
S.~Zhu%
, Li%
, Loy%
\BCBL {}\ \BBA {} Tang%
}{%
S.~Zhu%
\ \protect \BOthers {.}}{%
{\protect \APACyear {2015}}%
}]{%
CFSS}
\APACinsertmetastar {%
CFSS}%
\begin{APACrefauthors}%
Zhu, S.%
, Li, C.%
, Loy, C.C.%
\BCBL {} Tang, X.%
\end{APACrefauthors}%
\unskip\
\newblock
\APACrefYearMonthDay{2015}{}{}.
\newblock
{\BBOQ}\APACrefatitle {Face alignment by coarse-to-fine shape searching} {Face
  alignment by coarse-to-fine shape searching}.{\BBCQ}
\newblock
 \APACrefbtitle {Proc. IEEE Conf. Comput. Vis. Pattern Recognit.} {Proc. ieee
  conf. comput. vis. pattern recognit.}\ (\BPG~4998-5006).
\PrintBackRefs{\CurrentBib}

\bibitem [\protect \citeauthoryear {%
X.~Zhu%
\ \BBA {} Ramanan%
}{%
X.~Zhu%
\ \BBA {} Ramanan%
}{%
{\protect \APACyear {2012}}%
}]{%
AFW}
\APACinsertmetastar {%
AFW}%
\begin{APACrefauthors}%
Zhu, X.%
\BCBT {}\ \BBA {} Ramanan, D.%
\end{APACrefauthors}%
\unskip\
\newblock
\APACrefYearMonthDay{2012}{}{}.
\newblock
{\BBOQ}\APACrefatitle {Face detection, pose estimation, and landmark
  localization in the wild} {Face detection, pose estimation, and landmark
  localization in the wild}.{\BBCQ}
\newblock
 \APACrefbtitle {Proc. IEEE Conf. Comput. Vis. Pattern Recognit.} {Proc. ieee
  conf. comput. vis. pattern recognit.}\ (\BPG~2879-2886).
\PrintBackRefs{\CurrentBib}

\end{thebibliography}

\end{document}